\pdfoutput = 1

\RequirePackage{ifpdf}
\ifpdf
  \documentclass[journal,twocolumn]{IEEEtran}
\fi

\usepackage{ifpdf}
\usepackage{amsfonts}
\usepackage{latexsym}
\usepackage{amsfonts,amssymb,amsmath,amsthm}
\usepackage{graphics,graphicx,color}
\usepackage{subfigure,epic}
\usepackage[ruled,vlined]{algorithm2e}
\usepackage{multirow}
\usepackage{url}
\usepackage{ulem}
\usepackage{array}
\usepackage{cite}

\newtheorem{definition}{Definition}

\newtheorem{remark}{Remark}

\newcommand{\R}{{\mathbb R}}
\renewcommand{\baselinestretch}{1.2}

\DeclareMathOperator*{\tr}{tr~}
\DeclareMathOperator*{\st}{s.t.}
\DeclareMathOperator*{\vecop}{vec}

\makeatletter
\def\etal{\emph{et al}.}
\makeatother

\begin{document}

\title{\bf A new embedding quality assessment method for manifold learning}
\author{Peng~Zhang~\IEEEmembership{Member,~IEEE,}
    Yuanyuan~Ren,
    and Bo~Zhang
\thanks{P. Zhang is with the Data Center, National Disaster Reduction Center of China, Beijing, P.R. China
(e-mail: zhangpeng@ndrcc.gov.cn).}
\thanks{Y. Ren is with the Career Center, Tsinghua University, Beijing, P.R. China.}
\thanks{B. Zhang is with the LSEC and the Institute of Applied Mathematics, AMSS,
Chinese Academy of Sciences, Beijing 100190, China.}}

\maketitle


\begin{abstract}
Manifold learning is a hot research topic in the field of computer science. A
crucial issue with current manifold learning methods is that they lack a
natural quantitative measure to assess the quality of learned embeddings, which
greatly limits their applications to real-world problems. In this paper, a new
embedding quality assessment method for manifold learning, named as
Normalization Independent Embedding Quality Assessment (NIEQA), is proposed.
Compared with current assessment methods which are limited to isometric
embeddings, the NIEQA method has a much larger application range due to two
features. First, it is based on a new measure which can effectively evaluate
how well local neighborhood geometry is preserved under normalization, hence it
can be applied to both isometric and normalized embeddings. Second, it can
provide both local and global evaluations to output an overall assessment.
Therefore, NIEQA can serve as a natural tool in model selection and evaluation
tasks for manifold learning. Experimental results on benchmark data sets
validate the effectiveness of the proposed method.

\end{abstract}


\IEEEpeerreviewmaketitle

\section{Introduction}\label{sec_intro}

\IEEEPARstart{A}{long} with the advance of techniques to collect and store
large sets of high-dimensional data, how to efficiently process such data
issues a challenge for many fields in computer science, such as pattern
recognition, visual understanding and data mining. The key problem is caused by
``the curse of dimensionality" \cite{donoho00}, that is, in handling with such
data the computational complexities of algorithms often go up exponentially
with the dimension.

The main approach to address this issue is to perform dimensionality reduction.
Classical linear methods, such as Principal Component Analysis (PCA)
\cite{pca1,pca2} and Multidimensional Scaling (MDS) \cite{cox:00}, achieve
their success under the assumption that data lie in a linear subspace. However,
such assumption may not usually hold and a more realistic assumption is that
data lie on or close to a low-dimensional manifold embedded in the
high-dimensional ambient space. Recently, many methods have been proposed to
efficiently find meaningful low-dimensional embeddings from manifold-modeled
data, and they form a family of dimensionality reduction methods called
\emph{manifold learning}. Representative methods include Locally Linear
Embedding (LLE) \cite{lle1,lle2}, ISOMAP \cite{isomap1,isomap2}, Laplacian
Eigenmap (LE) \cite{le1,le2}, Hessian LLE (HLLE) \cite{hlle_pnas}, Diffusion
Maps (DM) \cite{dm_pnas,dm_pami}, Local Tangent Space Alignment (LTSA)
\cite{ltsa_siam}, Maximum Variance Unfolding (MVU) \cite{mvu_ijcv}, and
Riemannian Manifold Learning (RML) \cite{rml_pami}.

Manifold learning methods have drawn great research interests due to their
nonlinear nature, simple intuition, and computational simplicity. They also
have many successful applications, such as motion detection \cite{wang07_tip},
sample preprocessing \cite{chen07_tsmca}, gait analysis \cite{cheng08_pr},
facial expression recognition \cite{cheon09_pr}, hyperspectral imagery
processing \cite{bachmann05_tgrs}, and visual tracking \cite{hong10_tsmcb}.

Despite the above success, a crucial issue with current manifold learning
methods is that they lack a natural measure to assess the quality of learned
embeddings. In supervised learning tasks such as classification, the
classification rate can be directly obtained through label information and used
as a natural tool to evaluate the performance of the classifier. However,
manifold learning methods are fully unsupervised and the intrinsic degrees of
freedom underlying high-dimensional data are unknown. Therefore, after training
process, we can not directly assess the quality of the learned embedding. As a
consequence, model selection and model evaluation are infeasible. Although
visual inspection on the embedding may be an intuitive and qualitative
assessment, it can not provide a quantitative evaluation. Moreover, it can not
be used for embeddings whose dimensions are larger than three.

Recently, several approaches have been proposed to address the issue of
embedding quality assessment for manifold learning, which can be cast into tow
categories by their motivations.
\begin{itemize}
  \item Methods based on evaluating how well the rank of neighbor samples,
      according to pairwise Euclidean distances, is preserved within each
      local neighborhood.
  \item Methods based on evaluating how well each local neighborhood
      matches its corresponding embedding under rigid motion.
\end{itemize}

These methods are proved to be useful to isometric manifold learning methods,
such as ISOMAP and RML. However, a large variety of manifold learning methods
output normalized embeddings, such as LLE, HLLE, LE, LTSA and MVU, just to name
a few. In these method, embeddings have unit variance up to a global scale
factor. Then the distance rank of neighbor samples is disturbed in the
embedding as pairwise Euclidean distances are no longer preserved. Meanwhile,
anisotropic coordinate scaling caused by normalization can not be recovered by
rigid motion. As a consequence, existent methods would report false quality
assessments for normalized embeddings.

In this paper, we first propose a new measure, named Anisotropic Scaling
Independent Measure (ASIM), which can efficiently compare the similarity
between two configurations under rigid motion and anisotropic coordinate
scaling. Then based on ASIM, we propose a novel embedding quality assessment
method, named Normalization Independent Embedding Quality Assessment (NIEQA),
which can efficiently assess the quality of normalized embeddings
quantitatively. The NIEQA method owns three characteristics.
\begin{enumerate}
\item NIEQA can be applied to both isometric and normalized embeddings.
    Since NIEQA uses ASIM to assess the similarity between patches in
    high-dimensional input space and their corresponding low-dimensional
    embeddings, the distortion caused by normalization can be eliminated.
    Then even if the aspect ratio of a learned embedding is scaled, NIEQA
    can still give faithful evaluation of how well the geometric structure
    of data manifold is preserved.
\item NIEQA can provide both local and global assessments. NIEQA consists
    of two components for embedding quality assessment, a global one and a
    local one. The global assessment evaluates how well the skeleton of a
    data manifold, represented by a set of landmark points, is preserved,
    while the local assessment evaluates how well local neighborhoods are
    preserved. Therefore, NIEQA can provide an overall evaluation.
\item NIEQA can serve as a natural tool for model selection and evaluation
    tasks. Using NIEQA to provide quantitative evaluations on learned
    embeddings, we can select optimal parameters for a specific method and
    compare the performance among different methods.
\end{enumerate}

In order to evaluate the performance of NIEQA, we conduct a series of
experiments on benchmark data sets, including both synthetic and real-world
data. Experimental results on these data sets validate the effectiveness of the
proposed method.

The rest of the paper is organized as follows. A literature review on related
works is presented in Section \ref{sec:review}. The Anisotropic Scaling
Independent Measure (ASIM) is described in Section \ref{sec:asim}. Then the
Normalization Independent Embedding Quality Assessment (NIEQA) method is
depicted in Section \ref{sec:nieqa}. Experimental results are reported in
Section \ref{sec:expt}. Some concluding remarks as well as outlooks for future
research are given in Section \ref{sec:conclusion}.


\section{Literature review on related works}\label{sec:review}

In this section, the current state-of-the-art on embedding quality assessment
methods are reviewed. For convenience and clarity of presentation, main
notations used in this paper are summarized in Table \ref{tab:notations}.
Throughout the whole paper, all data samples are in the form of column vectors.
The superscript of a data vector is the index of its component.

\begin{table}[t]
\caption{Main notations.} \label{tab:notations}
\begin{center}
\begin{tabular}{|c|l|}
\hline $\R^n$ & $n$-dimensional Euclidean space where\\
    & high-dimensional data samples lie\\
$\R^m$ & $m$-dimensional Euclidean space, $m<n$, where \\
       & low-dimensional embeddings lie \\
$x_i$ & The $i$-th data sample in $\R^n$, $i=1,2,\ldots,N$\\
$\mathcal{X}$ & $\mathcal{X}=\{x_1,x_2,\ldots,x_N\}$\\
$X$ & $X=[x_1\ x_2\ \cdots\ x_N]$, $n\times N$ data matrix\\
$\mathcal{X}_i$ &
    $\mathcal{X}_i=\{x_{i_1},x_{i_2},\ldots,x_{i_k}\}$, local
    neighborhood of $x_i$\\
$X_i$ & $X_i=[x_{i_1}\ x_{i_2}\ \cdots\ x_{i_k}]$, $n\times k$ data matrix\\
$\mathcal{N}_k(x_i)$ & The index set of the $k$ nearest
neighbors of $x_i$ in $\mathcal{X}$\\
$y_i$ & low-dimensional embedding of $x_i$, $i=1,2,\ldots,N$\\
$\mathcal{Y}$ & $\mathcal{Y}=\{y_1,y_2,\ldots,y_N\}$\\
$Y$ & $Y=[y_1\ y_2\ \cdots\ y_N]$, $m\times N$ data matrix\\
$\mathcal{Y}_i$ &
    $\mathcal{Y}_i=\{y_{i_1},y_{i_2},\ldots,y_{i_k}\}$, low-dimensional
    embedding \\
    & of $\mathcal{X}_i$\\
$Y_i$ & $Y_i=[y_{i_1}\ y_{i_2}\ \cdots\ y_{i_k}]$, $m\times k$ data matrix\\
$\mathcal{N}_k(y_i)$ & The index set of the $k$ nearest
neighbors of $y_i$ in $\mathcal{Y}$\\
$e_k$ & $e=[1~1~\cdots~1]^T$, $k$ dimensional column vector \\
    & of all ones \\
$I_k$ & Identity matrix of size $k$\\
$\|\cdot\|_2$ & $L_2$ norm for a vector\\
$\|\cdot\|_F$ & Frobenius norm for a matrix\\
\hline
\end{tabular}
\end{center}
\end{table}


According to motivation and application range, existent embedding quality
assessment methods can be categorized into two groups: local approaches and
global approaches. Related works in the two categories are reviewed
respectively as follows.

\subsection{Local approaches}
\label{sec:review-eqa-local}

Goldberg and Ritov \cite{goldberg09_machlearn} proposed the Procrustes Measure
(PM) that enables quantitative comparison of outputs of isometric manifold
learning methods. For each $\mathcal{X}_i$ and $\mathcal{Y}_i$, their method
first uses Procrustes analysis \cite{sibson78_jrss,sibson79_jrss,seber} to find
an optimal rigid motion transformation, consisting of a rotation and a
translation, after which $\mathcal{Y}_i$ best matches $\mathcal{X}_i$. Then the
local similarity is computed as
\begin{equation*}
    \label{eq_rev_1}
    L(X_i,Y_i) = \sum_{j=1}^k\|x_{i_j}-Ry_{i_j}-b\|_2^2~,
\end{equation*}
where $R$ and $t$ are the optimal rotation matrix and translation vector,
respectively. Finally, the assessment is is given by
\begin{equation}
    \label{eq:Mp}
    M_{P} = \frac{1}{N}\sum_{i=1}^NL(X_i,Y_i)/\|H_kX_i\|_F^2~,
\end{equation}
where $H_k=I_k-e_ke_k^T$.

An $M_P$ close to zero suggests a faithful embedding. Reported experimental
results show that the PM method provides good estimation of embedding quality
for isometric methods such as ISOMAP. However, as pointed out by the authors,
PM is not suitable for normalized embedding since the geometric structure of
every local neighborhood is distorted by normalization. Although a modified
version of PM is proposed in \cite{goldberg09_machlearn}, which eliminates
global scaling of each neighborhood, it still can not address the issue of
sperate scaling of coordinates in the low-dimensional embedding.

Besides the PM method, a series of works follow the line that a faithful
embedding would yield a high degree of overlap between the neighbor sets of a
data sample and of its corresponding embedding. Several works are proposed by
using different ways to define the overlap degree. A representative one is the
LC meta-criteria (LCMC) proposed by Chen and Buja
\cite{chen06_phdthesis,chen09_jasa},  which can serve as a diagnostic tool for
measuring local adequacy of learned embedding. The LCMC assessment is defined
as the sum of local overlap degree and given by
\begin{equation}
    \label{eq:Mlc}
    M_{LC} = \frac{1}{kN}\sum_{i=1}^N|\mathcal{N}_k(x_i)\cap\mathcal{N}_k(y_i)|~.
\end{equation}

Venna and Kaski \cite{venna06_nn} proposed an assessment method which consists
of two measures, one for trustworthiness and one for continuity, based on the
change of indices of neighbor samples in $\R^n$ and $R^m$ according to pairwise
Euclidean distances, respectively. Aguirre \etal  proposed an alternative
approach for quantifying the embedding quality, by evaluating the possible
overlaps in the low-dimensional embedding. Their assessment is used for
automatic choice of the number of nearest neighbors for LLE
\cite{valencia09_lncs} and also exploited in \cite{daza10_neurocompt} to
evaluate the embedding quality of LLE with optimal regularization parameter.
Akkucuk and Carroll \cite{akkucuk06_jc} independently developed the Agreement
Rate (AR) metric which shares the same form to $M_{LC}$. Based on AR, they
suggested another useful assessment method called corrected agreement rate, by
randomly reorganize the indices of data in $\mathcal{Y}$. Also with AR, France
and Carroll \cite{france07_lncs} proposed a method using the RAND index to
evaluate dimensionality reduction methods.

Lee and Verleysen \cite{lee09_neurocompt,lee08_jmlr} proposed a general
framework, named co-ranking matrix, for rank-based criteria. The aforementioned
methods, which are based on distance ranking of local neighborhoods, can all be
cast into this unified framework. The block structure of the co-ranking matrix
also provides an intuitive way to visualize the differences between distinct
methods. In \cite{lee:2010}, they further extended their work to circumvent the
global scale dependency.

The above assessments based on overlap degrees of neighborhoods are implemented
in the same way: an embedding with good quality corresponds to a high value of
the assessment. They work well for isometric embeddings since pairwise
distances within each neighborhood are preserved. However, when the embedding
is normalized, the neighborhood structure is distorted since pairwise distances
are no longer kept. The overlap degree would be much lower than expected even
if the embedding is of high quality under visual inspection.

\subsection{Global approaches}
\label{sec:review-eqa-global}

Tenenbaum \etal \cite{isomap1} suggested to use the residual variance as a
diagnostic measure to evaluate the embedding quality. Given $\mathcal{X}$ and
$\mathcal{Y}$, the residual variance is computed by
\begin{equation}\label{eq:Mrv}
    M_{RV} = 1-\rho^2(G_X,D_Y))~,
\end{equation}
where $\rho(G_X,D_Y)$ is the standard linear correlation coefficients taken
over all entries of $G_X$ and $D_Y$. Here $G_X(i,j)$ is the approximated
geodesic distance between $x_i$ and $x_j$ \cite{isomap1} and
$D_Y(i,j)=\|y_i-y_j\|_2$. A low value of $M_{RV}$ close to zero indicates a
good equality of the embedding.

The $M_{RV}$ measure was applied to choose the dimension of learned embedding
for ISOMAP \cite{isomap1} and the optimal parameter for LLE
\cite{kouropteva02_icfskd}. Nevertheless, for a normalized embedding the
geodesic distances are no longer preserved and the reliability of $M_{RV}$ may
decrease in such case.

Doll\'ar \emph{et al}. \cite{dollar07_icml} proposed a supervised method for
model evaluation problem of manifold learning. They assume that there is a very
large ground truth data set containing the training data. Pairwise geodesic
distances are approximated within this set using ISOMAP, and the assessment is
defined as the average error between pairwise Euclidean distances in the
embedding and corresponding geodesic distances. However, in real situations we
do not usually have such ground truth set and their assessment can not be used
in general cases.

Recently, Meng \etal\ proposed a new quality assessment criterion to encode
both local-neighborhood-preserving and global-structure-holding performances
for manifold learning. In their method, a shortest path tree is first
constructed from the $k$-NN neighborhood graph of training data. Then the
global assessment is computed by using Spearman's rank order correlation
coefficient defined on the rankings of branch lengths. Finally, the overall
assessment is defined to be a linear combination of the global assessment and
$M_{LC}$. In their work, normalization is treated as a negative aspect in
quality assessment, while our work is to define a new assessment which is
independent of normalization.


\section{ASIM: Anisotropic Scaling Independent Measure}
\label{sec:asim}

In this section, we introduce a novel measure, named Anisotropic Scaling
Independent Measure (ASIM), which can effectively evaluate the similarity
between two configurations under rigid motion and anisotropic coordinate
scaling. A synthetic example is first given in Subsection \ref{sec:asim-eg} to
demonstrate why existent assessments fail under normalization. Then the
motivation and overall description of ASIM are presented in Subsection
\ref{sec:asim-description}. Finally, the computational details are stated in
Subsection \ref{sec:asim-computation}.

\subsection{A synthetic example}
\label{sec:asim-eg}

\begin{figure}[!t]
  \centering
  \subfigure[]{\includegraphics[scale=.45]{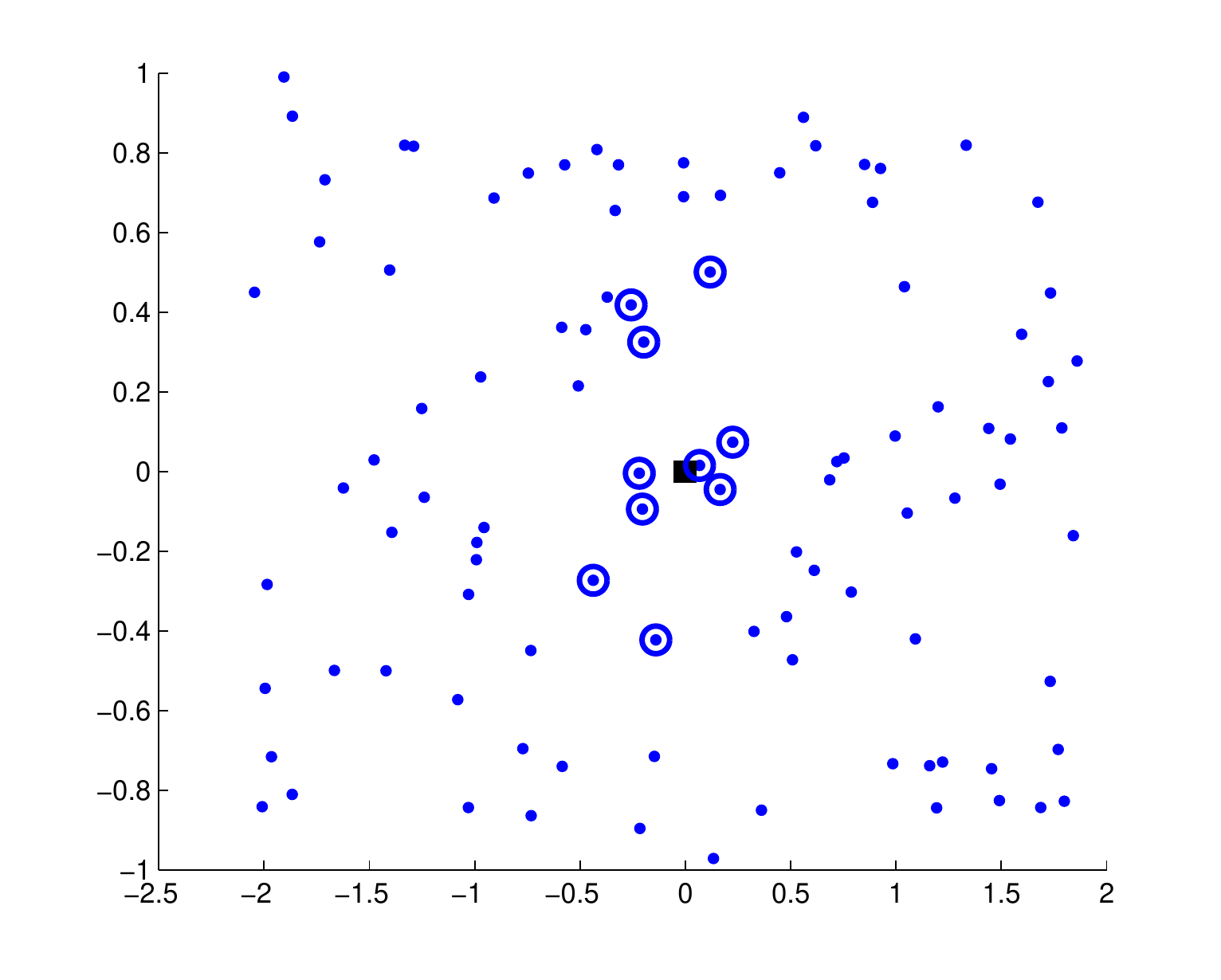}}
  \subfigure[]{\includegraphics[scale=.45]{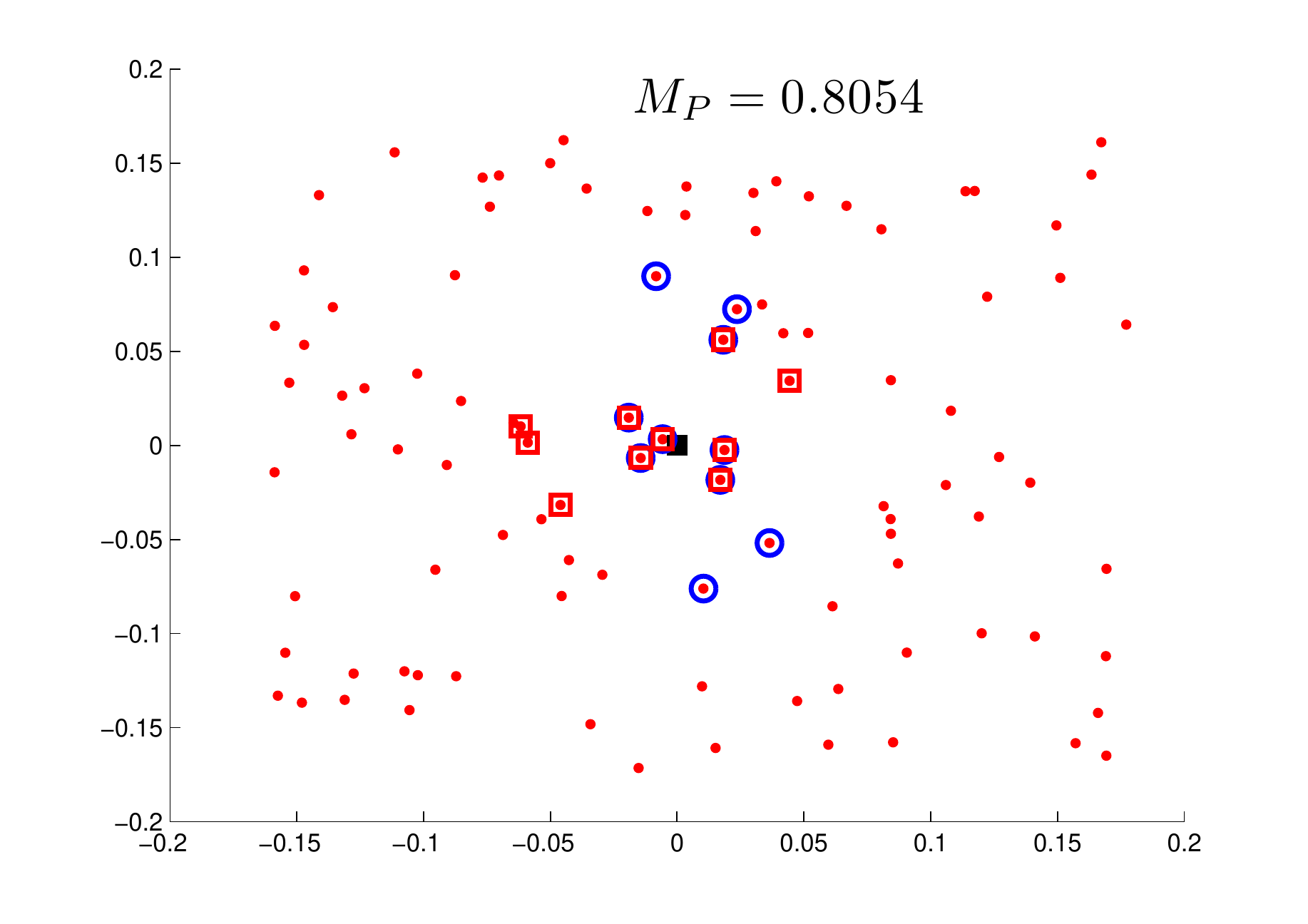}}
  \caption{A synthetic example where existent assessments fail.
  (a) Input data $\mathcal{X}$, marked by blue dots. (b) Normalized embedding $\mathcal{Y}$, marked by red dots. Black filled square: the origin $(0,0)^T$. Blue circles: the $k$ nearest neighbors of the origin in $\mathcal{X}$ and their corresponding embeddings. Red squares: the
  $k$ nearest neighbors of the origin in $\mathcal{Y}$. }
  \label{fig:asim-eg}
\end{figure}

We randomly generate 100 points within the area $[-2,2]\times[-1,1]$ in $\R^2$,
which form the input data set $\mathcal{X}=\{x_1,x_2,\ldots,x_{100}\}$. Next we
normalize $\mathcal{X}$ to get output data $\mathcal{Y}$ such that $YY^T=I_2$,
which are taken as the embedding of $\mathcal{X}$. In fact, $\mathcal{X}$ can
be obtained from $\mathcal{Y}$ through a rotation and anisotropic coordinate
scaling, that is, $X=RSY$ where
\begin{equation*}
  R=\begin{pmatrix}
      -0.9991 & 0.0434 \\
      0.0434 & 0.0991 \\
    \end{pmatrix}~,\
  S=\begin{pmatrix}
      11.6414 & 0 \\
      0 & 5.6236 \\
    \end{pmatrix}~.
\end{equation*}

In Fig. \ref{fig:asim-eg}(a), $x_i,i=1,2,\ldots,100$ are marked with blue dots
and the 10 nearest neighbors of the origin in $\mathcal{X}$ are marked with
blue circles. In Fig. \ref{fig:asim-eg}(b), $y_i,i=1,2,\ldots,100$ are marked
with red dots and the 10 nearest neighbors of the origin in $\mathcal{Y}$ are
marked with red squares. Meanwhile, the corresponding embeddings of the 10
nearest neighbors of the origin in $\mathcal{X}$ are marked with blue circles.
From Fig. \ref{fig:asim-eg}(b) we can see that the neighborhood of the origin
change a lot after normalization. Only 6 nearest neighbors are still in the
neighborhood after normalization and the overlap degree is only 60\%.
Meanwhile, we also compute the Procrustes measure $M_{P}$ between $\mathcal{X}$
and $\mathcal{Y}$ and show it in Fig. \ref{fig:asim-eg}(b). After
normalization, $M_{P}$ is as high as 0.8054.

Through this synthetic example, we can clearly observe the distortion on
$M_{P}$ and local neighborhood overlap degree caused by normalization.

\subsection{Motivation and description of ASIM}
\label{sec:asim-description}

Since a manifold is a topological space which is locally equivalent to a
Euclidean subspace, an embedding would be \emph{faithful} if it preserves the
structure of local neighborhoods. Then we face a question that how to define
the ``preservation" of local neighborhood structure.

Under the assumption that the data manifold is \emph{dense}, each local
neighborhood $\mathcal{X}_i$ can be roughly viewed as a linear subspace
embedded in the ambient space. Considering possible normalization on
$\mathcal{Y}$, a rational and reasonable choice is to define a new measure
which can efficiently assess the similarity between $\mathcal{X}_i$ and
$\mathcal{Y}_i$ under rigid motion and anisotropic coordinate scaling.

Formally, for each index $i$, we assume that there exists a rigid motion and
anisotropic coordinate scaling between $\mathcal{X}_i$ and $\mathcal{Y}_i$.
Since a rigid motion can be decomposed into a rotation and a translation, then
for any $x_{i_j}\in\mathcal{X}_i$ we assume that
\begin{equation}
  \label{eq:NIEQA-ma-17}
  x_{i_j} = P_iD_iy_{i_j}+t_i~,
\end{equation}
where $P_i\in\mathbb{R}^{n\times m}$ is orthogonal, that is, $P_i^TP_i=I_m$.
$D_i$ is a diagonal matrix of rank $m$ and $t_i\in\mathbb{R}^n$ stands for an
arbitrary translation.

To evaluate how similar $\mathcal{X}_i$ and $\mathcal{Y}_i$ are, our goal is to
find optimal $P_i^*$, $D_i^*$ and $t_i^*$ such that $\mathcal{Y}_i$ best
matches $\mathcal{X}_i$ under Eq. (\ref{eq:NIEQA-ma-17}). Equivalently, we need
to solve the following constrained optimization problem
\begin{equation}
  \label{eq:asim_opt}
  \begin{array}{ll}
  \min & \sum_{j=1}^k\|x_{i_j}-P_i^*D_i^*y_{i_j}-t_i^*\|_2^2\\
  \st  & P_i^TP_i=I_m\\
        & D_i \in \mathcal{D}(m)
  \end{array}~,
\end{equation}
where $\mathcal{D}(m)$ is the set of all diagonal matrices of rank $m$.

Then the neighborhood ``preservation" degree can be defined as the sum of
squared distances between corresponding samples in $\mathcal{X}_i$ and
$\mathcal{Y}_i$ under the above transformation. Formally, the anisotropic
scaling independent measure (ASIM) is defined as follows
\begin{equation}
  \label{eq:asim_def_vec}
  M_{asim}(X_i,Y_i) = \sum_{j=1}^k\|x_{i_j}-P_i^*D_i^*y_{i_j}-t_i^*\|_2^2/\sum_{j=1}^k\|x_{i_j}\|_2^2~,
\end{equation}
or in matrix form
\begin{equation}
  \label{eq:asim_def_mat}
  M_{asim}(X_i,Y_i) = \|X_i-P_i^*D_i^*Y_i-t_i^*e_k^T\|_F^2/\|X_i\|_F^2~,
\end{equation}
where the normalization item in denominator is introduced to eliminate
arbitrary scaling.

\subsection{Computation of ASIM}
\label{sec:asim-computation}

The optimization problem Eq. (\ref{eq:asim_opt}) does not admit a closed-form
solution. Alternatively, we use gradient descent method to solve Eq.
(\ref{eq:asim_opt}). Note that all $n\times m$ orthogonal matrices form the
so-called Stiefel manifold, which is a Riemannian submanifold embedded in
$\R^{nm}$. We denote this manifold by $St(n,m)$. Also note that
$\mathcal{D}(m)$ is closed for matrix addition, multiplication and scalar
multiplication, hence $\mathcal{D}(m)$ is homeomorphic to $\R^m$. Then Eq.
(\ref{eq:asim_opt}) can be resolved by using gradient descent method over
matrix manifolds.

For convenience of presentation, we first introduce the $\delta$ operator
\cite{dattorro:05}, which is defined as follows
\begin{definition}
\label{def:delta_operator} When the $\delta$ operator is defined on a
$n$-dimensional vector $v=(v_1,v_2,\cdots,v_n)^T$, $\delta(v)$ ia a $n\times n$
diagonal matrix whose diagonal entries are just components of $v$, that is
\begin{equation*}
  \delta(v)=\begin{pmatrix}
              v_1 &  &  &  \\
               & v_2 &  &  \\
               &  & \ddots &  \\
               &  &  & v_n \\
            \end{pmatrix}
            ~.
\end{equation*}
When the $\delta$ operator is defined on a $n\times n$ square matrix
$A=(a_{ij})$, $\delta(A)$ is a $n$-dimensional vector formed by the diagonal
entries of $v$, that is,
\begin{equation*}
  \delta(A)=(a_{11},a_{22},\cdots,a_{nn})^T~.
\end{equation*}
The $\delta$ operator can be compounded, which yields
\begin{eqnarray*}
  \delta^2(v)&=&v\\
  \delta^2(A)&=&\begin{pmatrix}
                  a_{11} &  &  &  \\
               & a_{22} &  &  \\
               &  & \ddots &  \\
               &  &  & a_{nn} \\
                \end{pmatrix}~.
\end{eqnarray*}
\end{definition}

With the above notations, Eq. (\ref{eq:asim_opt}) now can be rewritten in
matrix form as
\begin{eqnarray}
\label{eq:asim_opt_2}
    &\min\limits_{P_i,D_i,t_i}&\|X_i-P_iD_iY_i-t_ie_k^T\|_F^2\nonumber\\
    &\st& P_i\in St(n,m),\ D_i\in\mathcal{D}_m~.
\end{eqnarray}

Next, we solve Eq. (\ref{eq:asim_opt_2}) in three steps, which are described
respectively as follows.


\subsubsection{Computation $t_i^*$}

Let $L_i=P_iD_i$ and note that for any matrix $A$, $\|A\|_F^2=\tr(A^TA)$. Then
the objective function can be written as
\begin{equation}\label{eq:NIEQA-ma-7}
    f(L_i,t_i) = \tr\left((X_i-L_iY_i-t_ie_k^T)^T(X_i-L_iY_i-t_ie_k^T)\right)~.
\end{equation}
By using the propositions of matrix trace
Eq. (\ref{eq:NIEQA-ma-7}) can be expanded as
\begin{eqnarray}
  \label{eq:NIEQA-ma-8}
  f(L_i,t_i) & = &
    \tr(X_i^TX_i)+\tr(Y_i^TL_i^TL_iY_i)-\nonumber\\
  & & 2\tr(Y_iX_i^TL_i)+\tr(t_ie_k^Te_kt_i^T)-\nonumber\\
  & & 2\tr(e_k^TX_i^Tt_i)+\tr(e_k^TY_i^TL_i^Tt_i)~\mbox{.}
\end{eqnarray}

Taking derivative with resect to $t_i$ yields
\begin{equation*}\label{eq:NIEQA-ma-9}
    \frac{\partial f(L_i,t_i)}{\partial
    t_i}=2kt_i-2X_ie_k+2L_iY_it_i~.
\end{equation*}
Since $f(L_i,t_i)$ is a strict convex function of $t_i$, then by making both
sides of the above equation to be zero, we can get the optimal solution to
$t_i$ as follows
\begin{equation}\label{eq:NIEQA-ma-10}
    t_i^*=\frac{1}{k}(X_i-P_iD_iY_i)e_k~.
\end{equation}

Substitute $t_i^*$ into Eq. (\ref{eq:asim_opt_2}), and the latter one is
rewritten as
\begin{eqnarray}\label{eq:NIEQA-ma-20}
    &\min\limits_{P_i,D_i}&\|\bar{X}_i-P_iD_i\bar{Y}_i\|_F^2\nonumber\\
    &\st& P_i\in St(n,m),\ D_i\in\mathcal{D}_m~,
\end{eqnarray}
where $\bar{X}_i=X_i(I_k-\frac{1}{k}e_ke_k^T)$ and
$\bar{Y}_i=Y_i(I_k-\frac{1}{k}e_ke_k^T)$.


\subsubsection{Computation of $D_i^*$}

In the second step, we compute the optimal solution $D_i^*$ to $D_i$ with
respect to $P_i$. Let $A_i=\bar{Y_i}\bar{Y_i}^T$ and
$B_i=P^T\bar{X_i}\bar{Y_i}^T$, and denote the objective function in Eq.
(\ref{eq:NIEQA-ma-20}) by $f(P_i,D_i)$. Then we have
\begin{eqnarray*}
    f(P_i,D_i)
    &=&\tr(D^2A_i)-2\tr(DB_i)+\tr(\bar{X}_i\bar{X}_i^T)\\
    &=&\sum_{j=1}^ma_{jj}^{(i)}(d_j^{(i)})^2-2\sum_{j=1}^mb_{jj}^{(i)}d_j^{(i)}+\tr(\bar{X}_i\bar{X}_i^T)~,
\end{eqnarray*}
where $a_{jj}^{(i)}$, $b_{jj}^{(i)}$ and $d_j^{(i)}$ are the $j$-th diagonal
entries of $A_i$, $B_i$ and $D_i$, respectively.

Since $a_{jj}^{(i)}\geq0,\ j=1,2,\ldots,m$, $f$ is a convex function of vector
$\delta(D_i)$. Taking partial derivative with respect to $d_j^{(i)}$
($j=1,2,\ldots,m$) and by making them to be zero, we can get the global optimal
solutions to $d_j^{(i)}$ ($j=1,2,\ldots,m$) as follows
\begin{equation*}
    d_j^{(i)} = \frac{b_{jj}^{(i)}}{a_{ii}^{(i)}},\ j=1,2,\ldots,m~.
\end{equation*}
Then $D_i^*$ is given by
\begin{equation}
    \label{eq:NIEQA-Dstar}
    D_i^*=(\delta^2(A_i))^{-1}\delta^2(B_i)~.
\end{equation}

Substituting Eq. (\ref{eq:NIEQA-Dstar}) into $f$ yields
\begin{eqnarray*}
    f(P_i) 
        & = & \tr(A_i((\delta^2(A_i))^{-1}\delta^2(B_i))^2)-\\
        & & 2\tr((\delta^2(A_i))^{-1}\delta^2(B_i)B_i)+\tr(\bar{X_i}\bar{X_i}^T)\\
        & = & \sum_{j=1}^ma_{jj}^{(i)}\frac{(b_{jj}^{(i)})^2}{(a_{jj}^{(i)})^2}-2\sum_{j=1}^mb_{jj}^{(i)}\frac{b_{jj}^{(i)}}{a_{jj}^{(i)}}
            +\tr(\bar{X_i}\bar{X_i}^T)\\
        & = & -\sum_{j=1}^m\frac{(b_{jj}^{(i)})^2}{a_{jj}^{(i)}}+\tr(\bar{X_i}\bar{X_i}^T)~.
\end{eqnarray*}

Let $M_i=\bar{X}_i\bar{Y}_i^T(\delta^2(A_i))^{-1/2}$, then $f(P_i)$ can be
rewritten as
\begin{eqnarray*}
    \label{eq:NIEQA-fPi1}
    f(P_i) & = & -\sum_{j=1}^m(P_{i_j}^TM_{i_j})^2+\tr(\bar{X_i}\bar{X_i}^T)\nonumber\\
        & = & -\tr((P_i^TM_i)\odot(P_i^TM_i))+\tr(\bar{X_i}\bar{X_i}^T)~,
\end{eqnarray*}
where $P_{i_j}$ and $M_{i_j}$ are the $j$-th columns of matrices $P_i$ and
$M_i$, respectively. $\odot$ stands for the Hadamard product over matrices. The
optimization problem Eq. (\ref{eq:NIEQA-ma-20}) can be transformed into
\begin{eqnarray}\label{eq:NIEQA-piopt}
    &\max\limits_{P_i}&\phi(P_i)=\tr((P_i^TM_i)\odot(P_i^TM_i))\nonumber\\
    &\st& P_i\in St(n,m)~.
\end{eqnarray}

\subsubsection{Computation of $P_i^*$}

In the third step, we use gradient descent method over matrix manifold to solve
Eq. (\ref{eq:NIEQA-piopt}), which is an optimization problem for matrix
function over the Stiefel manifold $St(n,m)$.

Denote the gradient of $\phi$ in $\R^{nm}$ by $\nabla\bar{\phi}(P_i)$ and the
gradient of $\phi$ on $St(n,m)$ by $\nabla\phi(P_i)$, then by the proposition
of Stiefel manifold \cite{absil:07}, $\nabla\phi(P_i)$ is the projection of
$\nabla\bar{\phi}(P_i)$ onto the tangential space at $P_i$ and can be computed
by the following formula
\begin{equation}
    \label{eq:NIEQA-gradP}
    \nabla\phi(P_i)=\nabla\bar{\phi}(P_i)-
        P_i\frac{P_i^Z\nabla\bar{\phi}(P_i)+(\nabla\bar{\phi}(P_i))^TP_i}{2}~.
\end{equation}

Now all we need is to compute $\nabla\bar{\phi}(P_i)$. Let
$F(P_i)=(P_i^TM_i)\odot(P_i^TM_i)$. From matrix calculus, the differentiation
of $\phi$ with respect to $P_i$ is
\begin{equation}
    \label{eq:NIEQA-Dphi}
    D\phi(P_i)=(\vecop I_m)^TDF(P_i)~,
\end{equation}
where the $\vecop$ operator reformulates a $n\times m$ matrix into a
$nm$-dimensional vector by stacking its columns one underneath other.

Next we derive $DF(P_i)$. First, we have
\begin{equation*}
    dF(P_i)=2(M_i^TP_i)\odot(M_i^TdP_i)=2W_m^T((M_i^TP_i)\otimes(M_i^TdP_i))W_m~,
\end{equation*}
where $\otimes$ stands for the Kronecker product over matrices and $W_m=(\vecop
w_1w_1^T,\vecop w_2w_2^T,\cdots,\vecop w_mw_m^T)$ is an $m^2\times m$ matrix.
$w_i,\ i=1,2,\ldots,m$ is an $m$-dimensional vector who has 1 in its $i$-th
component and 0 elsewhere. Then we have
\begin{eqnarray*}
    \vecop dF(P_i) & = & 2\vecop(W_m^T((M_i^TP_i)\otimes(M_i^TdP_i))W_m)\nonumber\\
        & = & 2(W_m^T\otimes W_m)\vecop(M_i^TP\otimes(M_i^TdP_i))\nonumber\\
        & = & 2(W_m^T\otimes W_m)(H_i\otimes I_m)\vecop(M_i^TdP_i)\nonumber\\
        & = & 2(W_m^T\otimes W_m)(H_i\otimes I_m)(I_m\otimes M_i^T)d\vecop P_i~,
\end{eqnarray*}
where $H_i=((I_m\otimes K_{mm})((\vecop M_i^TP_i)\otimes I_m))\otimes I_m$.
Here $K_{mm}$ is a permutation matrix of order $m^2$, and for any square matrix
$M$ of order $m$, $K_{mm}\vecop M=\vecop M^T$. Then by matrix calculus
\cite{magnus:99}, we have
\begin{equation*}
    DF(P_i) = 2(W_m^T\otimes W_m)(H_i\otimes I_m)(I_m\otimes M_i^T)~.
\end{equation*}
Furthermore, through algebraic deduction and Eq. (\ref{eq:NIEQA-Dphi}), we have
\begin{equation*}
    D\phi(P_i)=(\vecop I_m)^TDF(P_i)=2\vecop(M_i\delta^2(P_i^TM_i))^T~.
\end{equation*}

Then $\nabla\bar{\phi}(P_i)$ is given by the following formula
\begin{equation*}
    \nabla\bar{\phi}(P_i)=2M_i\delta^2(P_i^TM_i)~,
\end{equation*}
and by using Eq. (\ref{eq:NIEQA-gradP}), $\nabla\phi(P_i)$ now reads
\begin{equation}
    \label{eq:NIEQA-gradPst}
    \nabla\phi(P_i)=2M_i\delta^2(P_i^TM_i)-P_iP_i^T\delta^2(P_i^TM_i)-P\delta^2(P_i^TM_i)M_i^TP_i~.
\end{equation}

Given a step length for iteration, we apply gradient descent method to find
$P_i^*$ such that $\nabla\phi(P_i)$ vanishes. In each iteration, we first
update $P_i$ as
\begin{equation*}
    \tilde{P}_i=P_i+\alpha\nabla\phi(P_i)~.
\end{equation*}
Then we retract $\tilde{P}_i$ to $St(n,m)$. From the property of $St(n,m)$,
such retraction can be obtained through the QR decomposition of $\tilde{P_i}$.
Let $\tilde{P}_i=Q_iR_i$, where $Q_i\in St(n,m)$ and $R_i$ is an
upper-triangular matrix. The retraction of $\tilde{P_i}$ to $St(n,m)$ is just
$Q_i$.

In each iteration, we use $Q_i$ to update $P_i$ until $\|\nabla\phi(P_i)\|_F$
is less than a given threshold $\epsilon$. After $P_i^*$ is computed, $D_i^*$
can be given by Eq. (\ref{eq:NIEQA-Dstar}), and the optimal value to Eq.
(\ref{eq:NIEQA-ma-20}) is $f(P_i^*,D_i^*)$.


\subsubsection{The algorithm and discussion}

Finally, we summarize the computation process of $M_{asim}$ in Algorithm
\ref{alg:NIEQA-2}.

When the dimension $n$ of input samples is very high, performing QR
decomposition of $\tilde{P}_i$ in each iteration will greatly increase of
computational complexity of Algorithm \ref{alg:NIEQA-2}. A possible solution to
this issue is first projecting $\mathcal{X}_i$ to its tangential space, denoted
as $T\mathcal{X}_i$, and then computing $M_{asim}(T\mathcal{X}_i,Y_i)$. When
data are densely distributed on the manifold, $T\mathcal{X}_i$ can optimally
recover the local linear structure of a manifold. Therefore, such strategy is
feasible. The tangential space can be approximated by using PCA, MDS or the
method proposed in \cite{iltsa}.

\begin{algorithm}[!t]
    \label{alg:NIEQA-2}
    \caption{Anisotropic Scaling Independent Measure (ASIM), $M_{asim}$.}
    \SetKwInOut{Input}{Input}
    \Input{Local neighborhood matrix $X_i$ and corresponding embedding matrix
    $Y_i$, number of nearest neighbors $k$, step length for iteration $\alpha$,
    and threshold $\epsilon$ for stopping criterion.}
    \SetKwInOut{Output}{Output}
    \Output{$M_{asim}(X_i,Y_i)$.}
    Step 1. Assign $\bar{X}_i=X_i(I_k-e_ke_k^T)$.\\
    Step 2. Assign $\bar{Y}_i=Y_i(I_k-e_ke_k^T)$.\\
    Step 3. Set initial value $P_i^{(0)}$ for $P_i$.\\
    Step 4. Use Eq. (\ref{eq:NIEQA-gradPst}) to compute $\nabla\phi(P_i^{(0)})$.\\
    Step 5. If $\|\nabla\phi(P_i^{(0)})\|_F<\epsilon$, goto Step 6; otherwise, do
    \begin{equation*}
        P_i^{(0)}\leftarrow P_i^{(0)}+\alpha\nabla\phi(P_i^{(0)})~.
    \end{equation*}
    Compute the QR decomposition of $P_i^{(0)}$, $P_i^{(0)}=Q_iR_i$. Let $P_i^{(0)}\leftarrow
    Q_i$ and goto Step 5.\\
    Step 6. Let $P_i^*=P_i^{(0)}$ and use Eq. (\ref{eq:NIEQA-Dstar}) to compute $D_i^*$.\\
    Step 7. Use Eq. (\ref{eq:NIEQA-ma-10}) to compute $t_i^*$.
    Step 8. Compute $M_{asim}(X_i,Y_i)$ through Eq. (\ref{eq:asim_def_mat}).
\end{algorithm}


\section{Normalization independent embedding quality assessment}
\label{sec:nieqa}

When assessing the quality of embeddings, we need to consider both local and
global evaluations. This leads to two issues.
\begin{itemize}
\item Does the embedding preserve the global topology of the manifold?
\item Does the embedding preserve the geometric structure of local neighbor
    neighborhoods?
\end{itemize}

In this section, we propose Normalization Independent Embedding Quality
Assessment method (NIEQA) to address these two issues, which is independent of
normalization. NIEQA is based on the ASIM measure stated in Section
\ref{sec:asim} and consists of two assessments, a local one and a global one.
In the following subsections, we introduce these two assessments respectively
as well as how NIEQA can be implemented in model selection and model
evaluation.


\subsection{Local assessment}
\label{sec:nieqa-local}

For local neighborhood $\mathcal{X}_i$ on a data manifold and its corresponding
low-dimensional embedding $\mathcal{Y}_i$, the local measure $M_{asim}$ defined
in last section characterizes how well local neighborhood structure is
preserved and is independent of normalization. Therefore, we define the local
assessment as the mean value of $M_{asim}(X_i,Y_i)$ over index $i$, that is,
\begin{equation}
    \label{eq:NIEQA-ML}
    M_L(X,Y) = \frac{1}{N}\sum_{i=1}^NM_{asim}(X_i,Y_i)~.
\end{equation}

\subsection{Global assessment}
\label{sec:nieqa-global}

From geometric intuition, if an embedding preserves the global topology of the
data manifold well, then such embedding should preserve relative positions
among ``representative" samples on the manifold. In other words, if we treat
these ``representative" samples as a local neighborhood, where pairwise
Euclidean distances among neighborhood samples are replaced with pairwise
geodesic distances on the manifold, then a good embedding should preserve the
geometric structure of this neighborhood.

Motivated by the above consideration, we define the global assessment as the
matching degree between the aforementioned described neighborhood and its
corresponding embedding under rigid motion and anisotropic coordinate scaling.

The computation of the global assessment consists of three steps, which are
depicted below, respectively.
\begin{enumerate}
\item \textbf{Selecting landmark points.} First, for each training sample
    $x_i$, find its $k_l$ nearest neighbors. Treat $x_i$ as a node in a
    graph and add edges among neighboring samples with edge length being
    pairwise Euclidean distance. Through such construction we get a
    connected graph. Then we use the shortest path length between $x_i$ and
    $x_j$ to approximate the geodesic distance between them for all $i$ and
    $j$. Next, we count how many shortest paths going through each $x_i$
    and record this number as its importance degree. Finally, the top 10\%
    most important data samples are selected as landmark points on the
    manifold and the set they formed is denoted by $\mathcal{X}_l$.
\item \textbf{Computing $\tilde{\mathcal{Y}}_l$}. Once $\mathcal{X}_l$ is
    fixed in the first step, the distance between any two landmark points
    is defined to be the approximated geodesic distance. Then we implement
    MDS \cite{cox:00} to $\mathcal{X}_l$ to obtain its isometric embedding
    $\tilde{\mathcal{Y}}_l$, which optimally preserve relative positions of
    landmark points on the manifold. Note that the dimensions of
    $\tilde{\mathcal{Y}}_l$ and $\mathcal{Y}_l$ are equal, and the latter
    one is the subset in $\mathcal{Y}$ corresponding to $\mathcal{X}_l$.
\item \textbf{Computing the global assessment}. We define the global
    assessment $M_G$ to be the ASIM measure between $\tilde{\mathcal{Y}}_l$
    and $\mathcal{Y}_l$
    \begin{equation}
        \label{eq:NIEQA-MG}
        M_G(X,Y) = M_{asim}(\tilde{Y}_l,Y_l)~,
    \end{equation}
    where $\tilde{Y}_l$ and $Y_l$ are the $m\times l$ data matrices
    corresponding to $\tilde{\mathcal{Y}}_l$ and $\mathcal{Y}_l$,
    respectively.
\end{enumerate}

\begin{remark}
During landmark points selection, the parameter $k_l$ needs to be set manually.
Based on experimental experience, setting $k_l=0.1N$ can yield a connected
graph that approximates the manifold structure well. However, if the graph is
disconnected under current $k_l$, $k_l$ should be set to be the smallest
integer which makes the graph fully connected.

The landmark points selection method stated above has intuitive geometric
motivation and is easy to implement. It can also be replaced with other more
accurate yet more complicated approaches, for example, the methods proposed in
\cite{silva06_nips} and \cite{li09_pr}.
\end{remark}

\subsection{Implementation in model evaluation and model selection}
\label{sec:NIEQA-4.3}

In this subsection, we state how to implement the NIEQA method to model
evaluation and model selection for manifold learning.

\begin{itemize}
    \item \textbf{Model evaluation}. Given $X$, suppose that we have two
        embeddings, namely $Y_1$ and $Y_2$, obtained by different manifold
        learning methods. Then we say that $Y_1$ owns better locality
        preservation than $Y_2$ if $M_L(X,Y_1)<M_L(X,Y_2)$ and vice versa.
        We say $Y_1$ owns better global topology preservation than $Y_2$ if
        $M_G(X,Y_1)<M_G(X,Y_2)$ and vice versa.
    \item \textbf{Model selection}. Given $X$ and a set of parameters
        $\mathcal{P}=\{p_1,p_2,\ldots,p_l\}$, for each parameter $p_i$ we
        compute its corresponding embedding $Y^{(i)}$ using specific
        manifold learning method. Then we use $M_G$ or $M_L$ or their
        combination, which depends on the user's demand, to evaluate the
        quality of $Y^{(i)}$. Finally, the $p_i$ corresponding to the
        lowest assessment score is chosen to be the optimal parameter.
\end{itemize}


\section{Experiments}
\label{sec:expt}

In this section, the effectiveness of the NIEQA method is validated through a
series of experimental tests on benchmark data sets. In Subsection
\ref{sec:expt-me}, NIEQA is applied to model evaluation for manifold learning.
In Subsection \ref{sec:expt-me}, NIEQA is used to select optimal parameters for
the LTSA method which outputs normalized embeddings. In experiments, NIEQA is
compared with three commonly used assessment methods. We compute $1-M_{LC}$
instead $M_{LC}$ to obtain a unified criterion, that is, a small assessment
value close to zero indicates good quality of the embedding. The benchmark data
sets used in experiments are briefly depicted in Table \ref{tab:expt-data} and
notations for methods are summarized in Table \ref{tab:expt-assessments}.

\begin{table}[!t]
\caption{Description of experimental data sets.} \label{tab:expt-data}
\begin{center}
\begin{tabular}{llllll}
\hline
Data manifold & $N$ & $n$ & $m$ & Description\\
\hline
\texttt{Swissroll} & 1000 & 3 & 2 & Surface isometrically \\
  & & & & embedded in $\mathbb{R}^3$\\
\texttt{Swisshole} & 1000 & 3 & 2 & Surface embedded \\
  & & & & in $\mathbb{R}^3$\\
\texttt{Gaussian} & 1000 & 3 & 2 & Surface isometrically \\
  & & & & embedded in $\mathbb{R}^3$\\
\hline
\texttt{lleface} & 1493 & 560 & 2 & Face manifold with \\
  & & & & resolution $28\times20$\\
\hline
\end{tabular}
\end{center}
\end{table}

\begin{table}[!t]
\caption{Notations used in experiments.} \label{tab:expt-assessments}
\begin{center}
\begin{tabular}{l|l}
\hline
Notation & Description\\
\hline
$M_P$ & Procrustes measure (Eq. (\ref{eq:Mp})) \cite{goldberg09_machlearn}\\
$M_P^c$ & $M_P$ with global scaling removed \cite{goldberg09_machlearn}\\
$M_{LC}$ & LCMC measure (Eq. (\ref{eq:Mlc})) \cite{chen06_phdthesis,chen09_jasa}\\
$M_{RV}$ & Residual Variance measure (Eq. (\ref{eq:Mrv})) \cite{isomap1}\\
\hline
$M_L$ & Local assessment of NIEQA (Eq. (\ref{eq:NIEQA-ML}))\\
$M_G$ & Global assessment of NIEQA (Eq. (\ref{eq:NIEQA-MG}))\\
\hline
$M_t$ & Matching degree between $\mathcal{Y}$ and \\
  & ground truth $\mathcal{U}$, $M_{asim}(Y,U)$\\
\hline
\end{tabular}
\end{center}
\end{table}

\subsection{Model evaluation}
\label{sec:expt-me}

\begin{figure*}[!t]
    \centering
    \subfigure[$\mathcal{X}$]{\includegraphics[scale=.3]{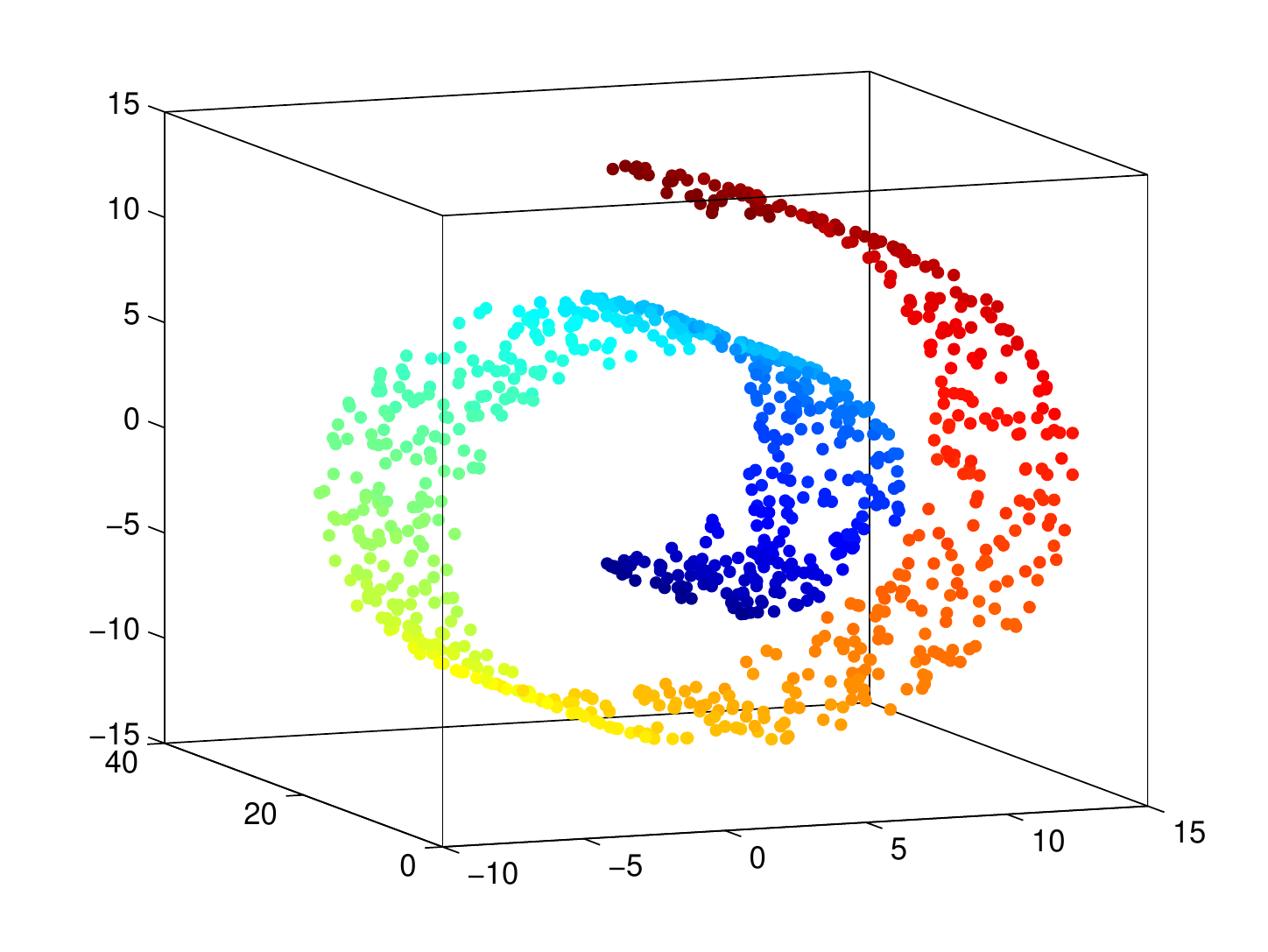}}
    \subfigure[$\mathcal{U}$]{\includegraphics[scale=.3]{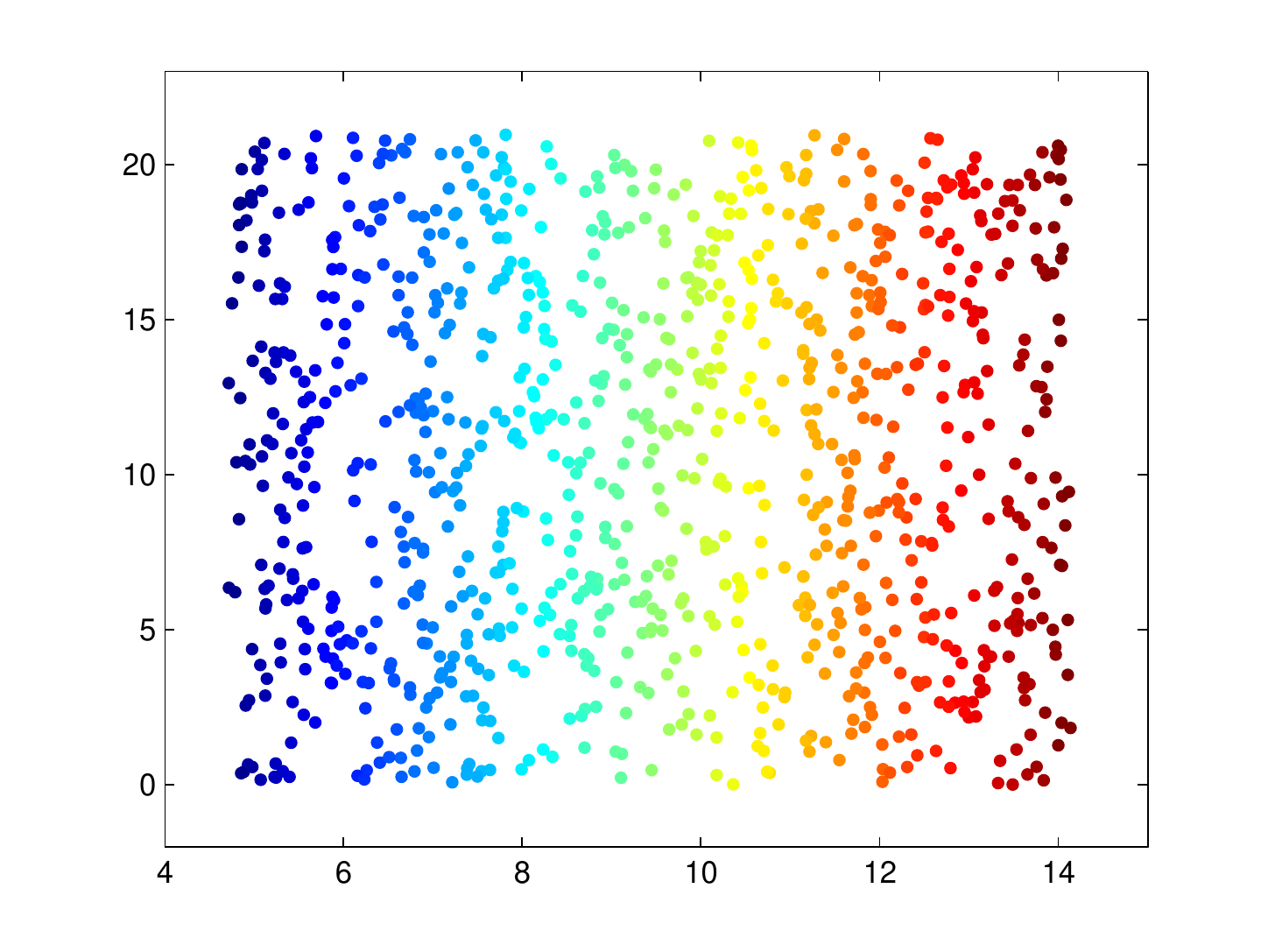}}
    \subfigure[LLE]{\includegraphics[scale=.3]{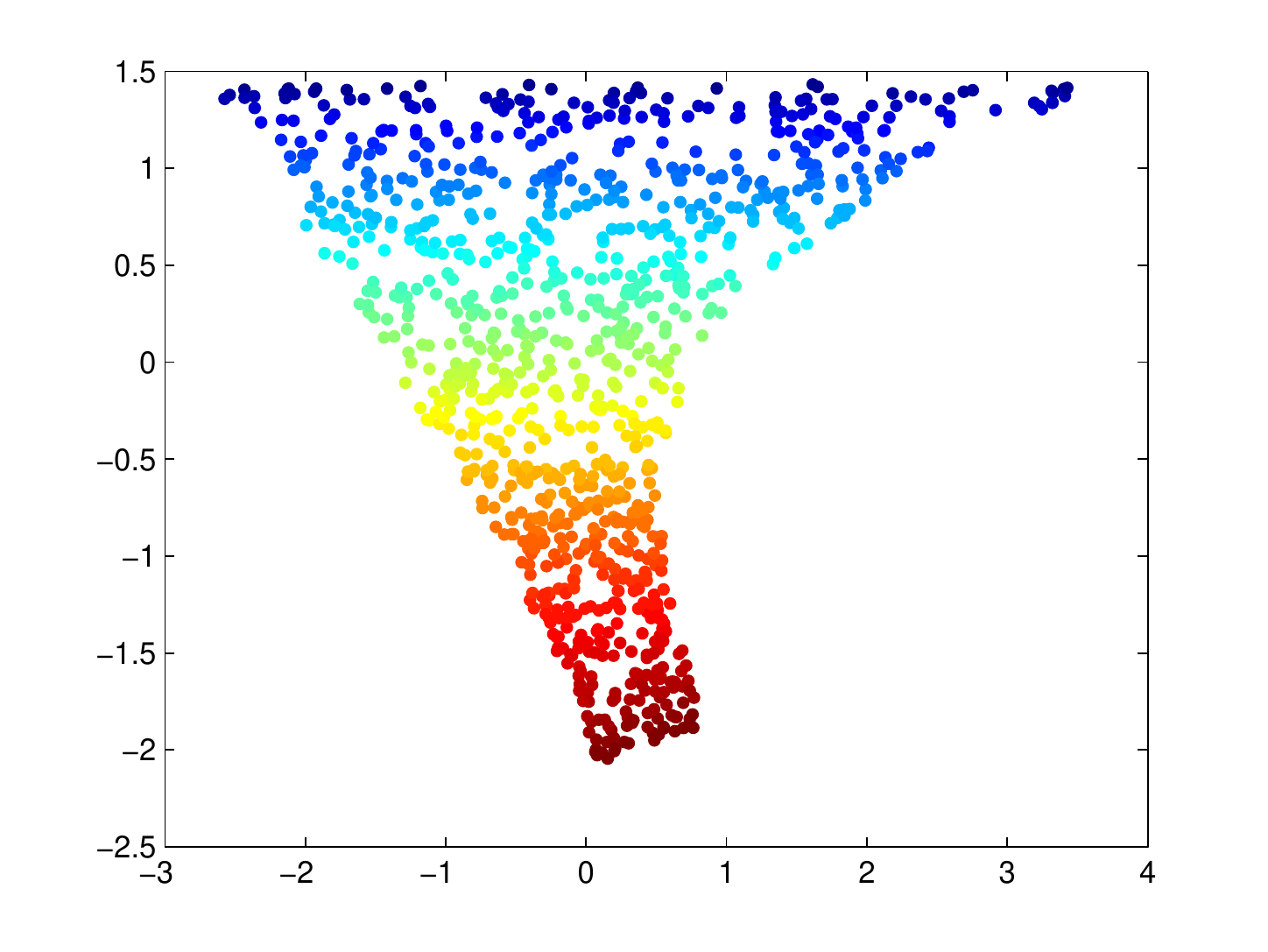}}
    \subfigure[LE]{\includegraphics[scale=.3]{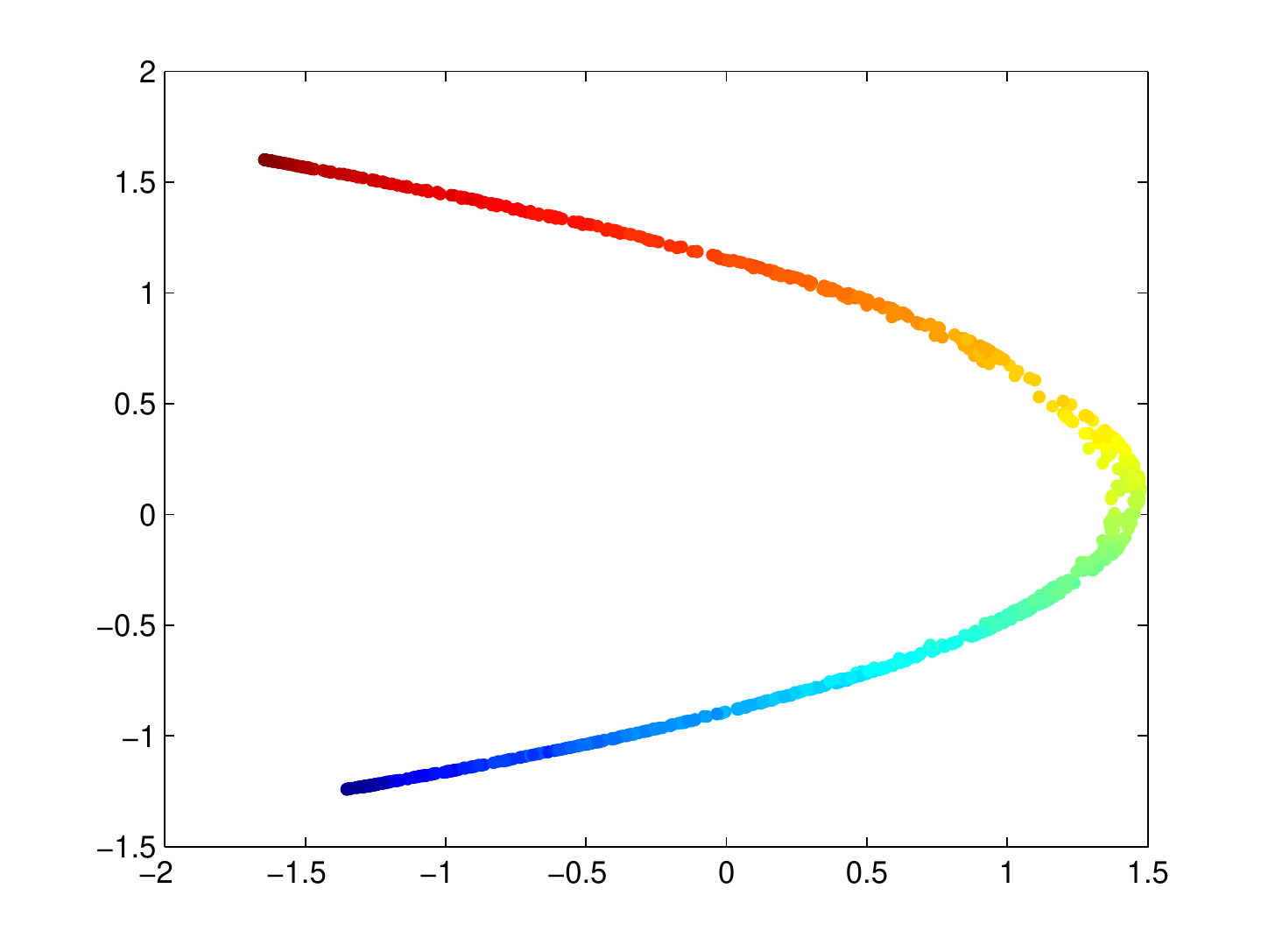}}
    \subfigure[LTSA]{\includegraphics[scale=.3]{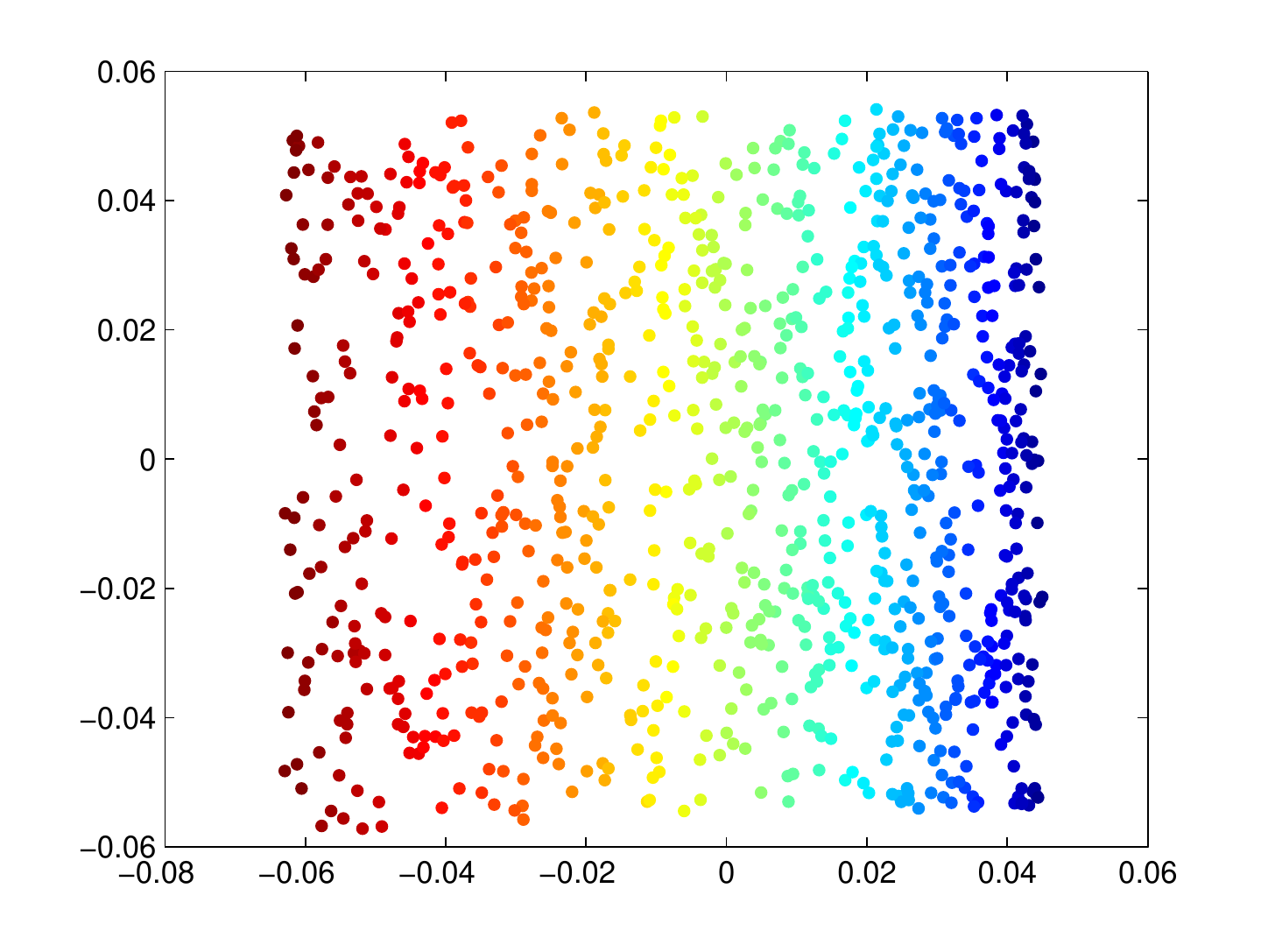}}
    \subfigure[ISOMAP]{\includegraphics[scale=.3]{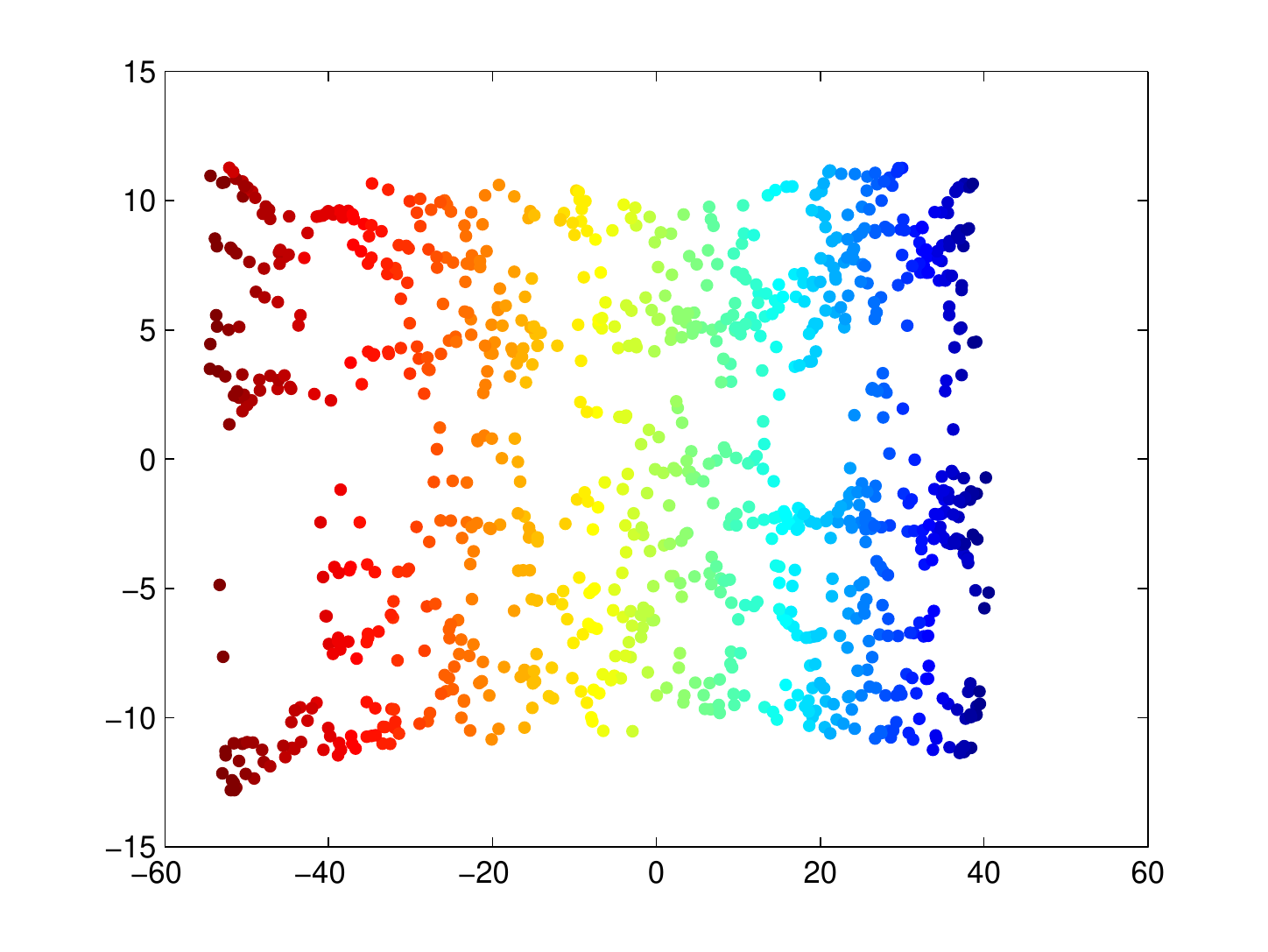}}
    \subfigure[RML]{\includegraphics[scale=.3]{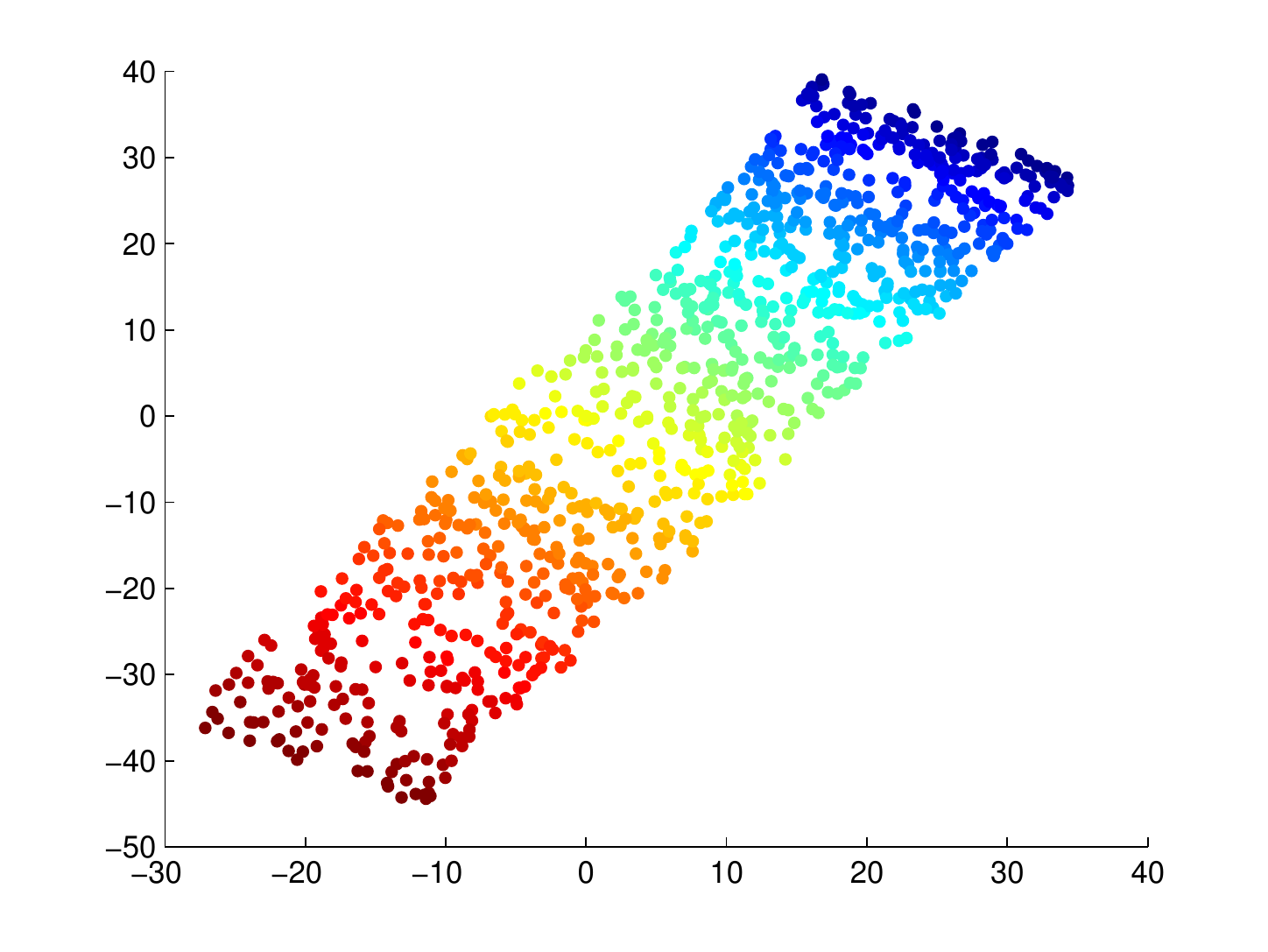}}\\
    \subfigure[$M_P$]{\includegraphics[scale=.3]{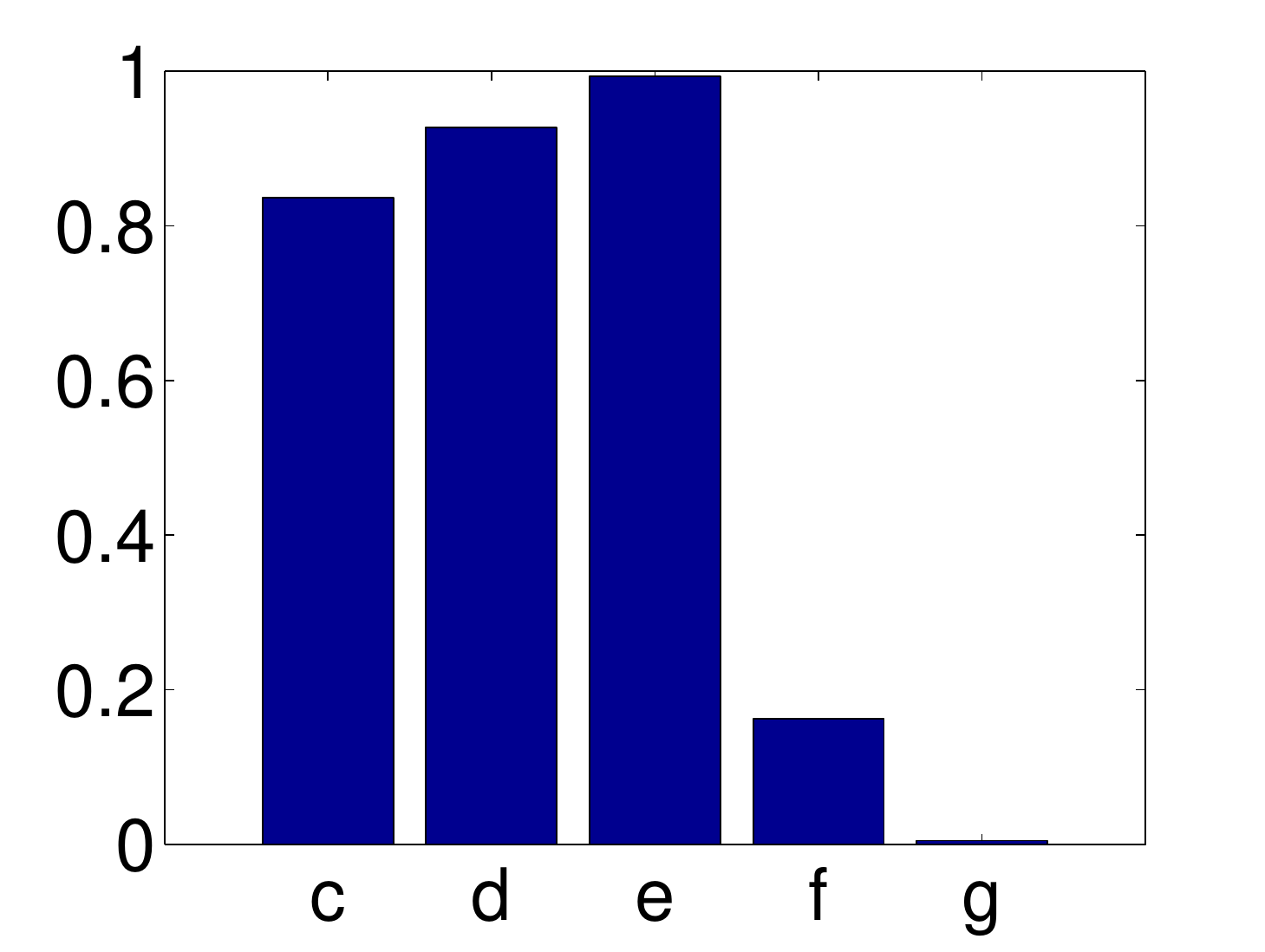}}
    \subfigure[$M_P^c$]{\includegraphics[scale=.3]{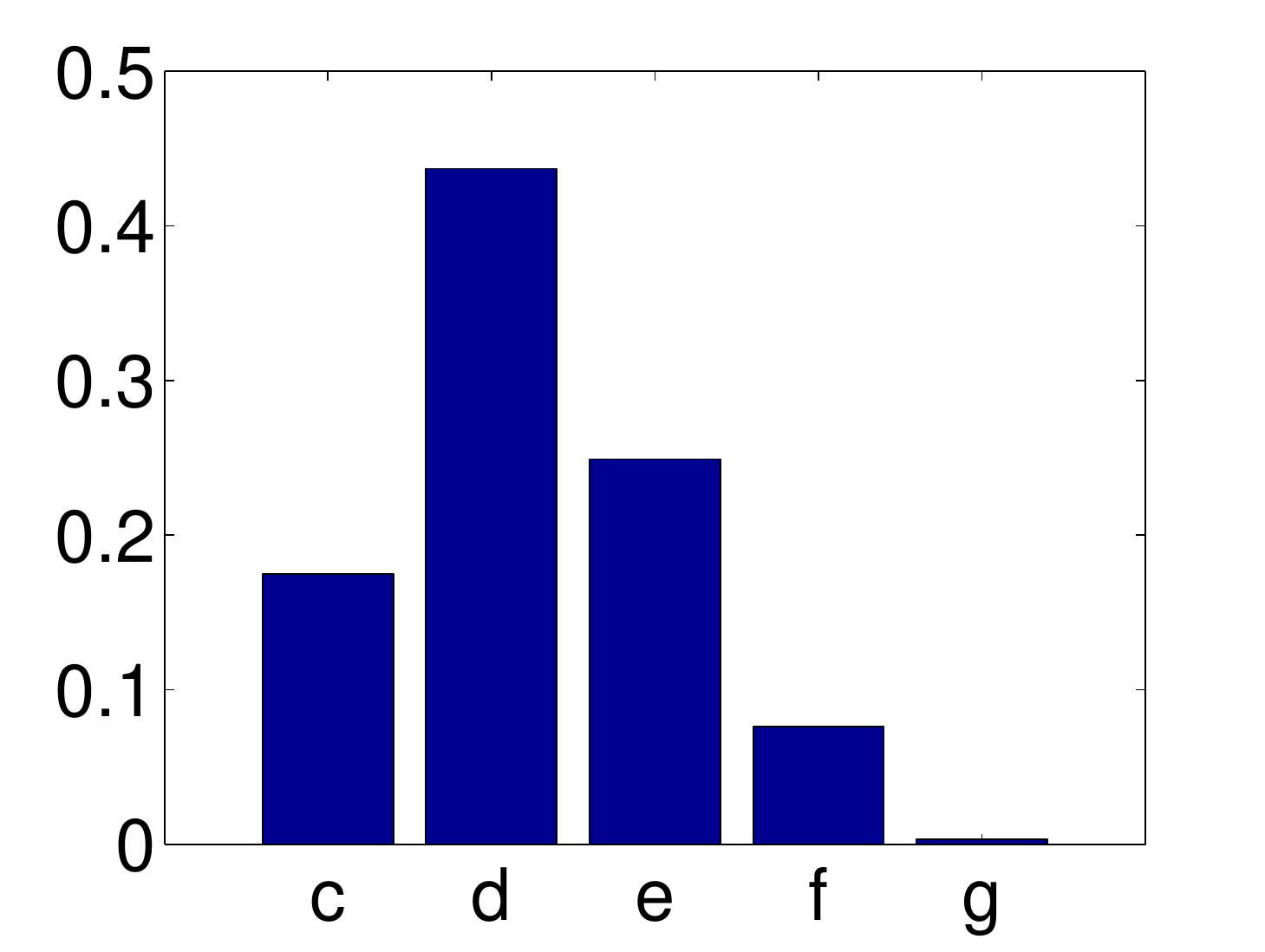}}
    \subfigure[1-$M_{LC}$]{\includegraphics[scale=.3]{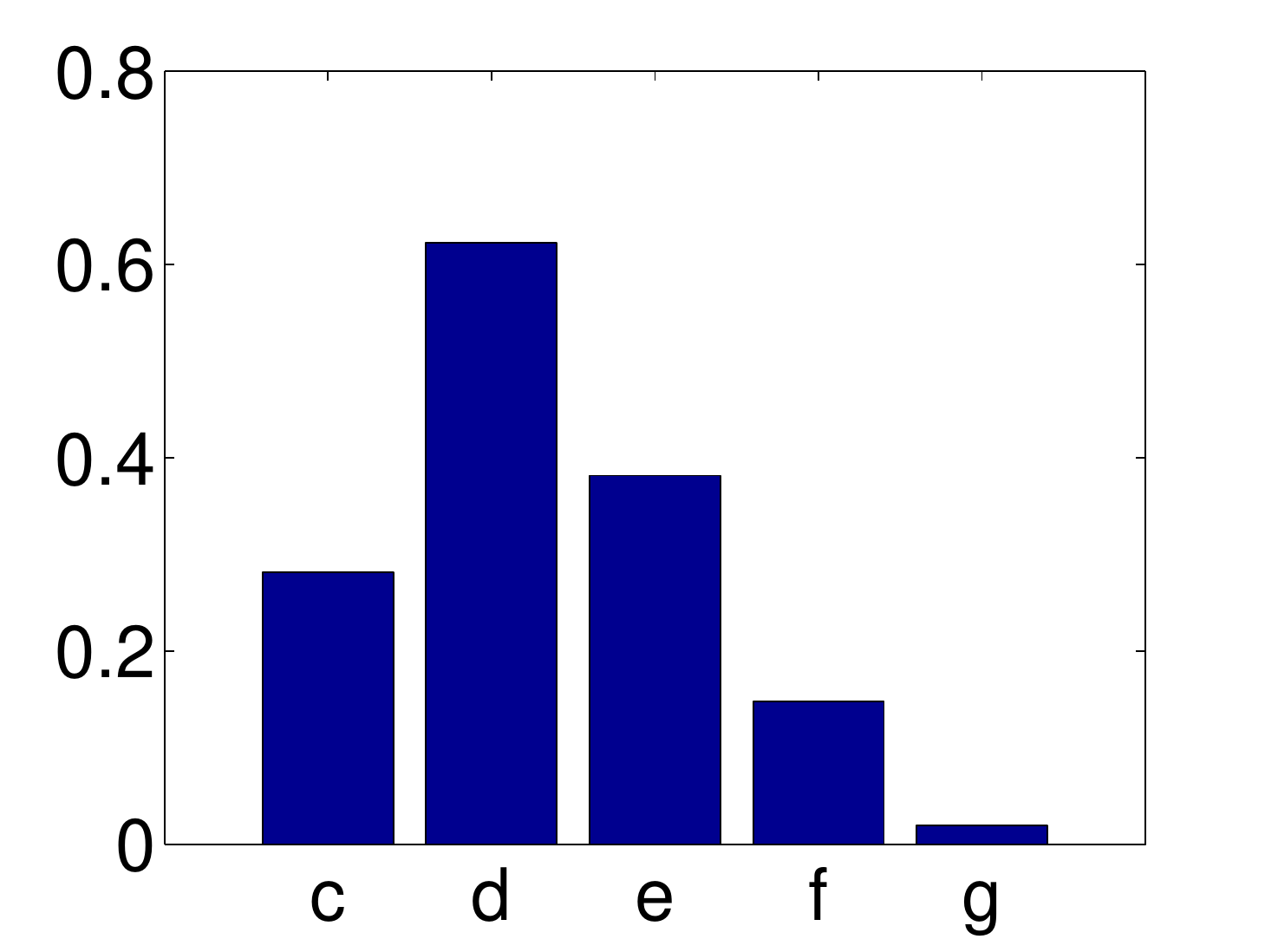}}
    \subfigure[$M_{RV}$]{\includegraphics[scale=.3]{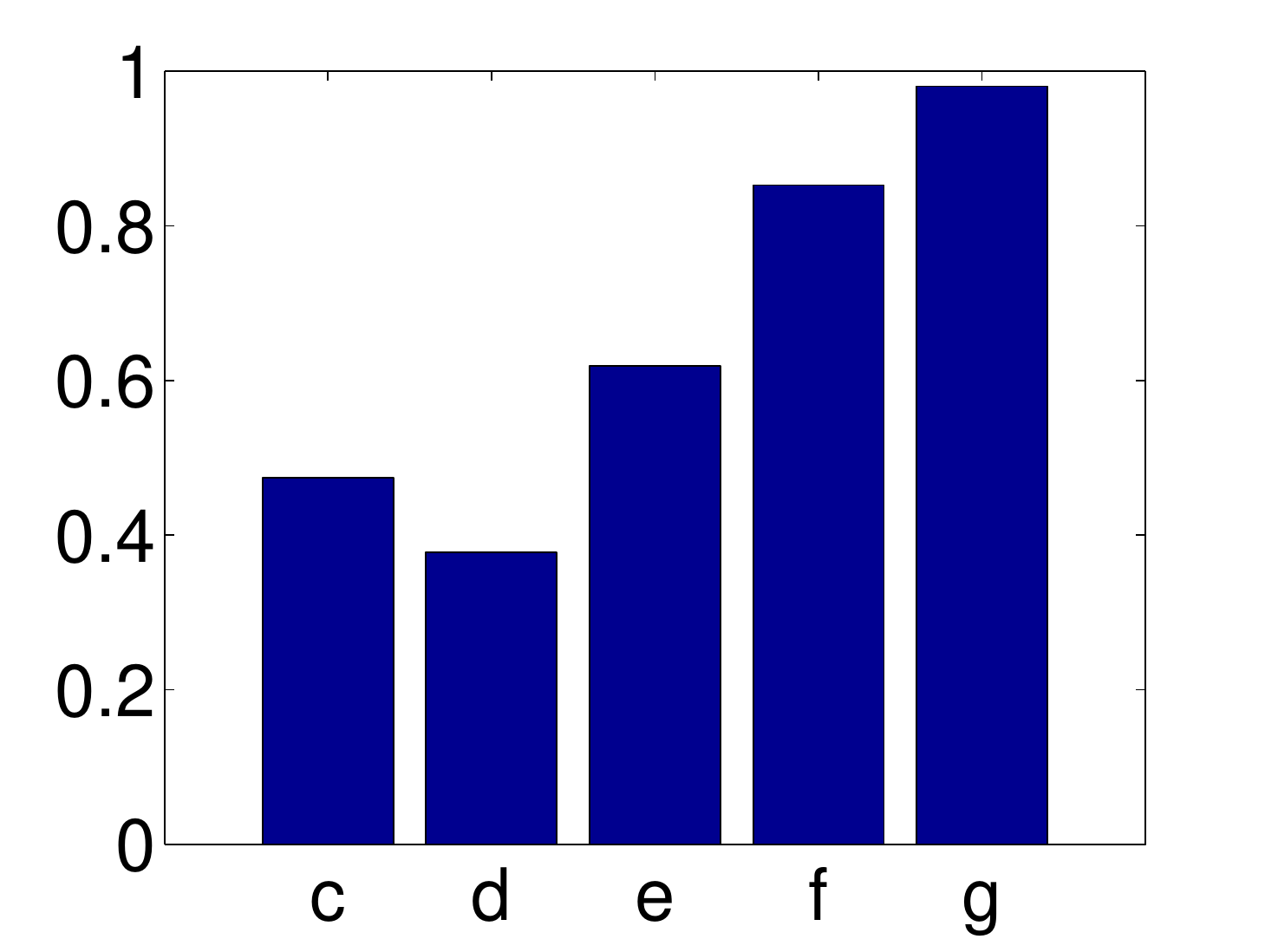}}
    \subfigure[$M_L$]{\includegraphics[scale=.3]{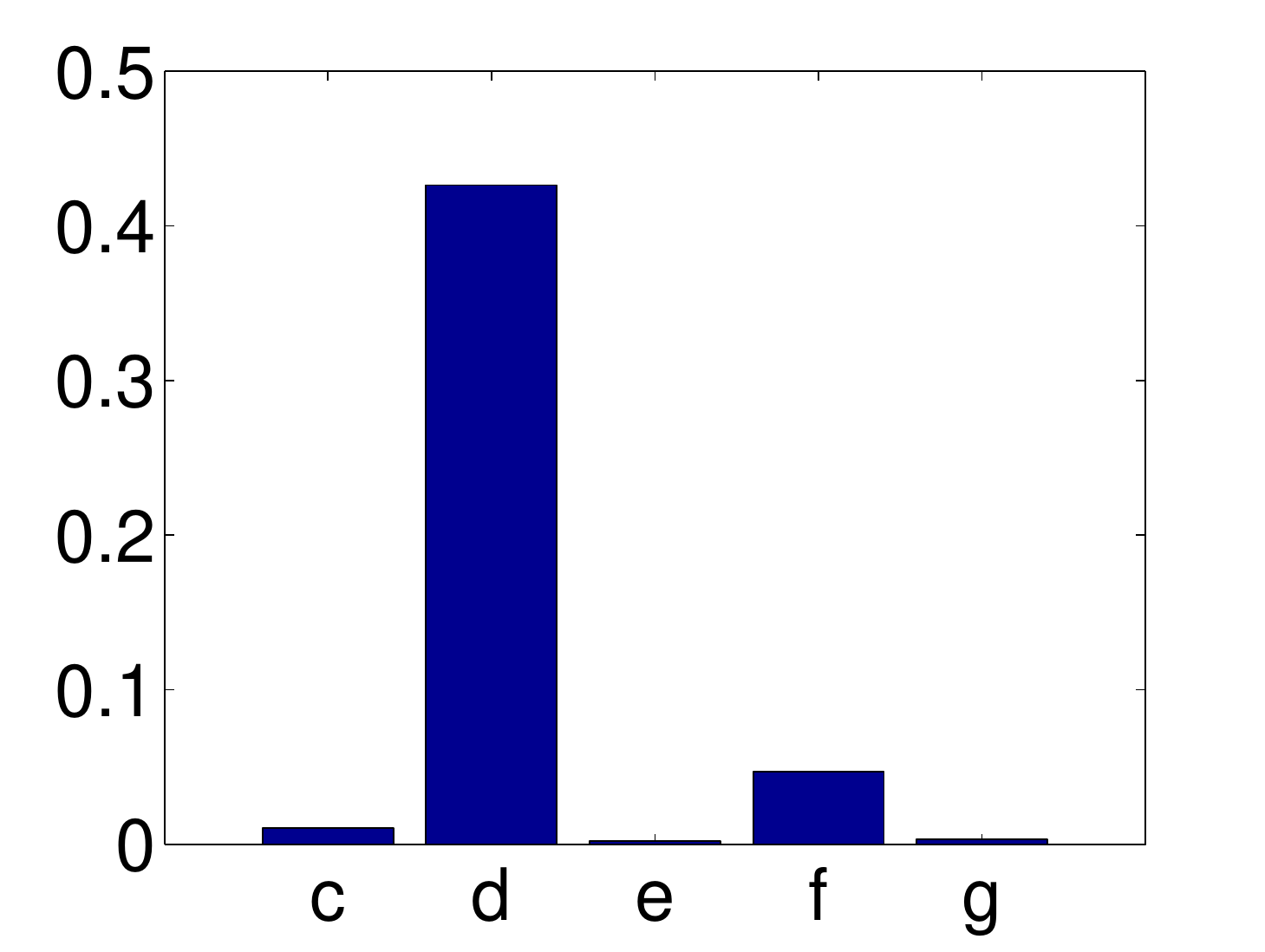}}
    \subfigure[$M_G$]{\includegraphics[scale=.3]{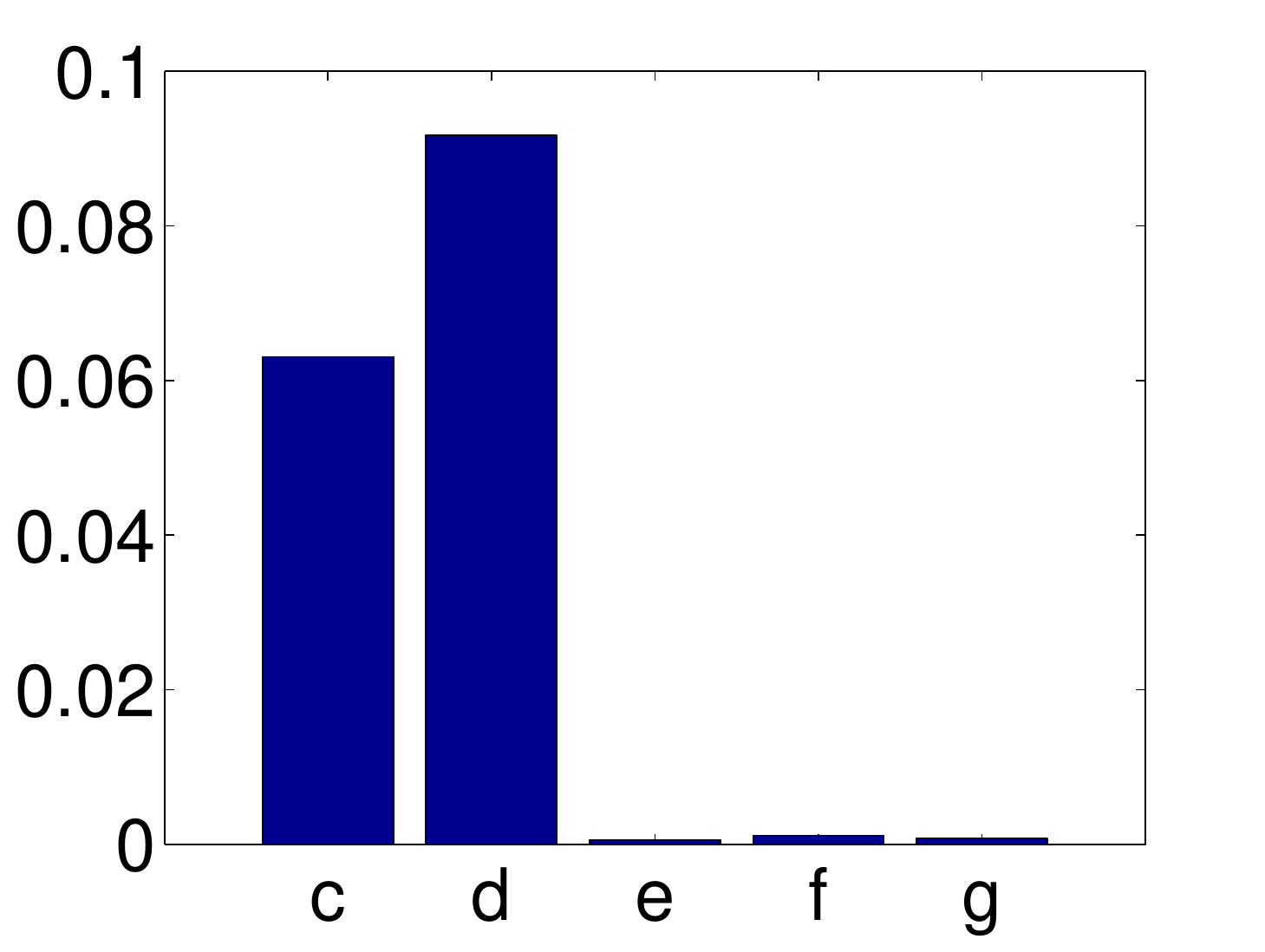}}
    \subfigure[$M_t$]{\includegraphics[scale=.3]{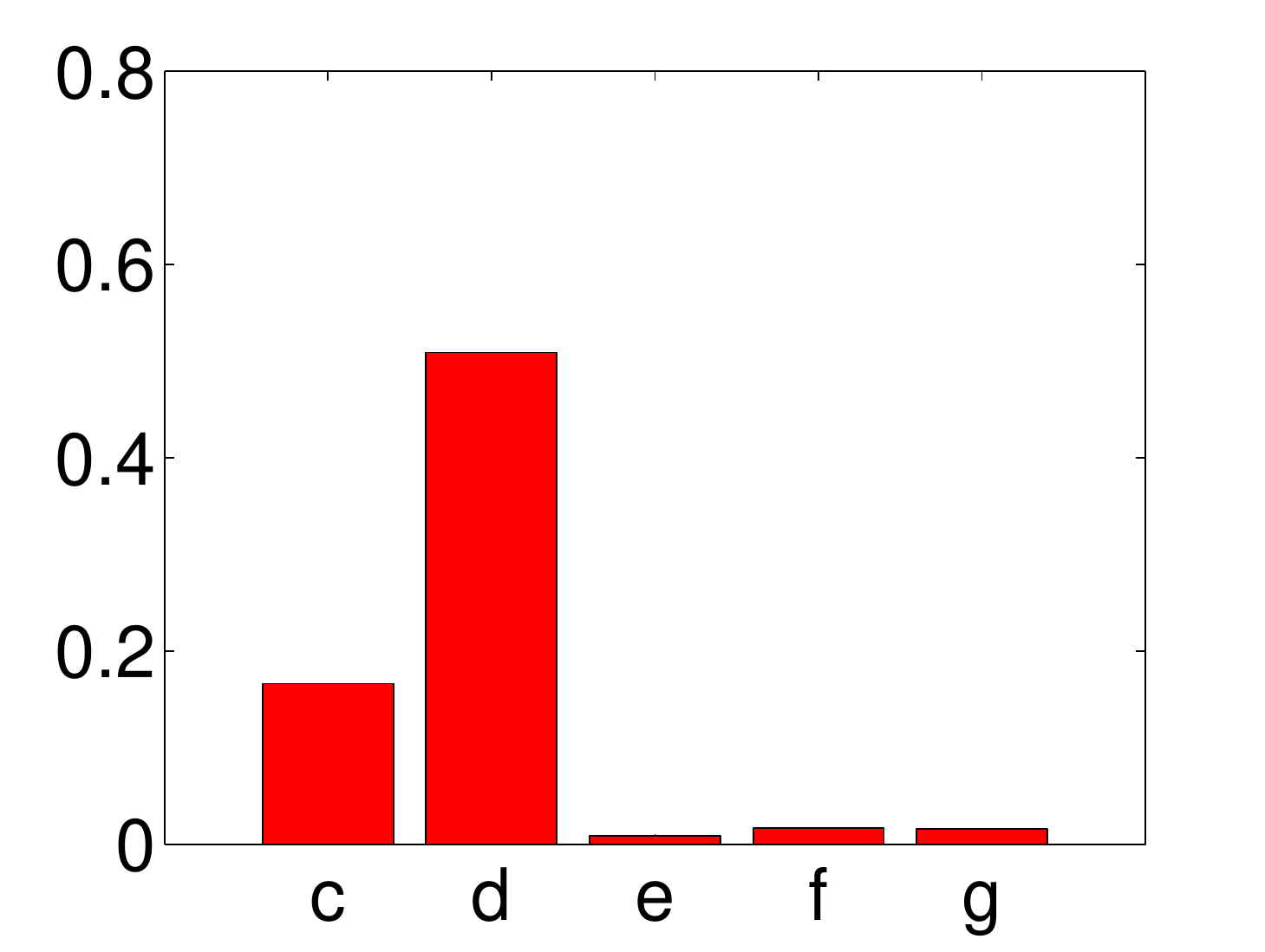}}
    \caption{Manifold learning results on \texttt{Swissroll}. (a) Training data $\mathcal{X}$. (b) Groundtruth of intrinsic degrees of freedom $\mathcal{U}$. (c)-(g) Embeddings learned by various method. The name of each method is stated below each subfigure. (h)-(n) Bar plots of different assessments on learned embeddings. The lower-case character under each bar corresponds to the index of the subfigure above.}
    \label{fig:NIEQA-me-sw}
\end{figure*}

In the first experiment, we apply NIEQA to model evaluation of the
\texttt{Swissroll} manifold with parameter equation
\begin{equation*}
    \left\{
    \begin{array}{lll}
        x^1 & = & u^1\cos u^1\\
        x^2 & = & u^2  \\
        x^3 & = & u^1\sin u^1
    \end{array}
    \right.~.
\end{equation*}
We use LLE\cite{lle1}, LE\cite{le2}, LTSA\cite{ltsa_siam}, ISOMAP\cite{isomap1}
and RML\cite{rml_pami} to learn this manifold, respectively. 1000 training
samples are randomly generated and the number of nearest neighbors is 10. Figs.
\ref{fig:NIEQA-me-sw} (c)-(g) shows the results of manifold learning, where
$\mathcal{X}$ and $\mathcal{U}$ stands for the training data and the
groundtruth of intrinsic degrees of freedom, respectively. By visual
inspection, the embeddings given by LTSA and RML are the most similar to
$\mathcal{U}$. The one given by ISOMAP is a litter worse, and the one learned
by LLE has a great change in global shape. LE fails to recover the geometric
structure of $\mathcal{U}$.

For embeddings given by the above methods, we compute the different assessments
described in Table \ref{tab:expt-assessments} and use bar plots to visualize
their values in Figs. \ref{fig:NIEQA-me-sw}(h)-(m). From the bar plots, we can
see that $M_P$ only works for isometric embeddings given by ISOMAP and RML
while reports false high values for normalized embeddings learned by LTSA and
LLE. Although $M_P^c$ eliminates the affects of global scaling, only the scale
of $M_P$ is normalized and it still reports false high values for normalized
embeddings. $M_{LC}$ and $M_{RV}$ fails to output reasonable equality
evaluations. It should be noted that $M_{RV}$ is originally designed for the
ISOMAP method and hence works well for the embedding given by ISOMAP.


The two assessments $M_L$ and $M_G$ in NIEQA provide overall and reasonable
evaluations on embedding quality for various methods. $M_L$ shows that LTSA and
RML best preserve local neighborhood. LLE and ISOMAP perform worse, and LE
performs the worst. $M_G$ further indicates that the global-shape-preservation
of the embedding given by LLE is not good. This completely matches visual
inspection, which demonstrates that NIEQA can effectively evaluate the quality
of both isometric and normalized embeddings.

Besides, the bar plot of the matching degree $M_t$ between $\mathcal{Y}$ and
$\mathcal{U}$ is shown in Fig. \ref{fig:NIEQA-me-sw}(n). We can see that only
$M_L$ and $M_G$ match $M_t$, which validates the effectiveness of NIEQA.

\begin{figure*}[!t]
    \centering
    \subfigure[$\mathcal{X}$]{\includegraphics[scale=.3]{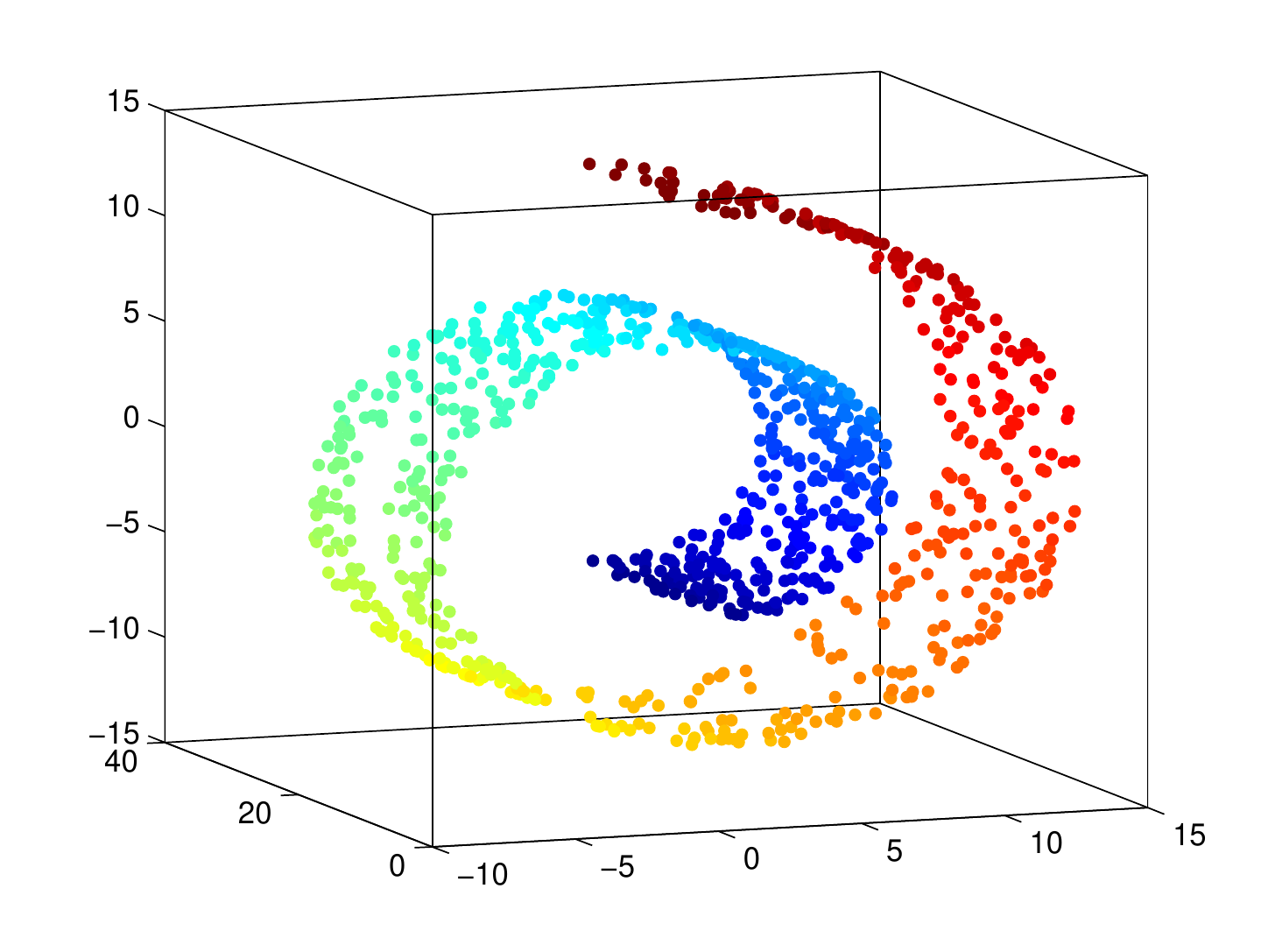}}
    \subfigure[$\mathcal{U}$]{\includegraphics[scale=.3]{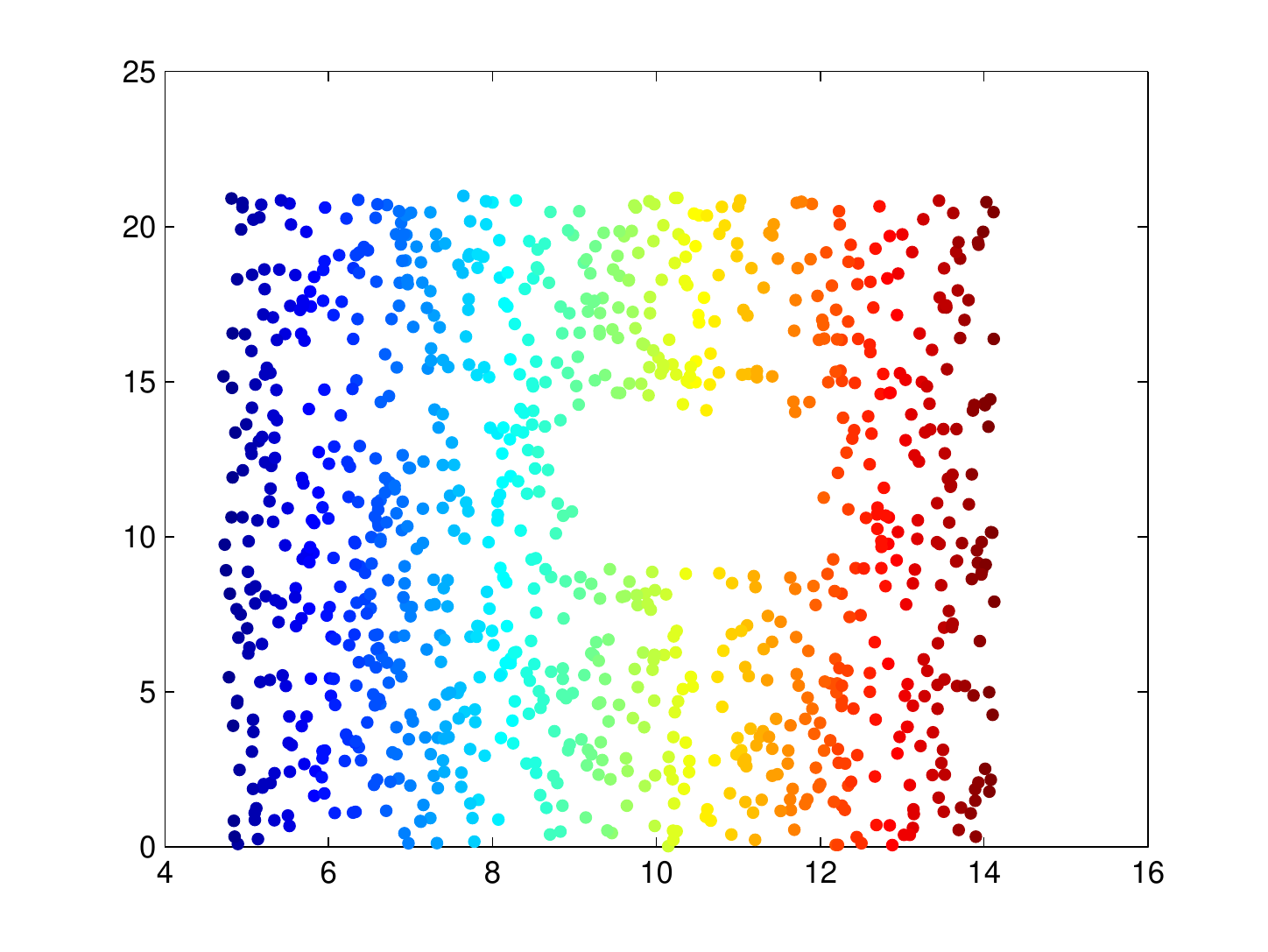}}
    \subfigure[LLE]{\includegraphics[scale=.3]{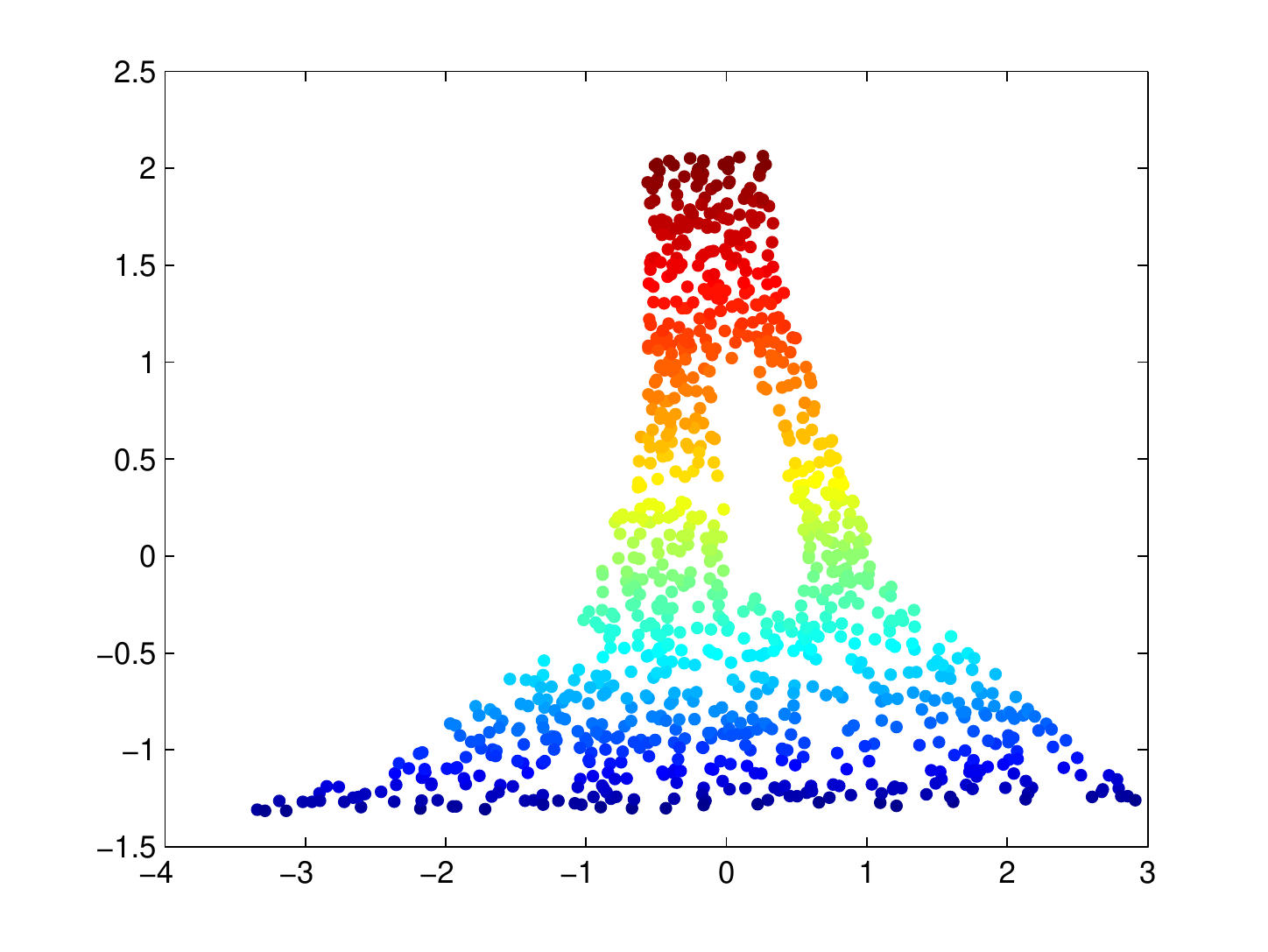}}
    \subfigure[LE]{\includegraphics[scale=.3]{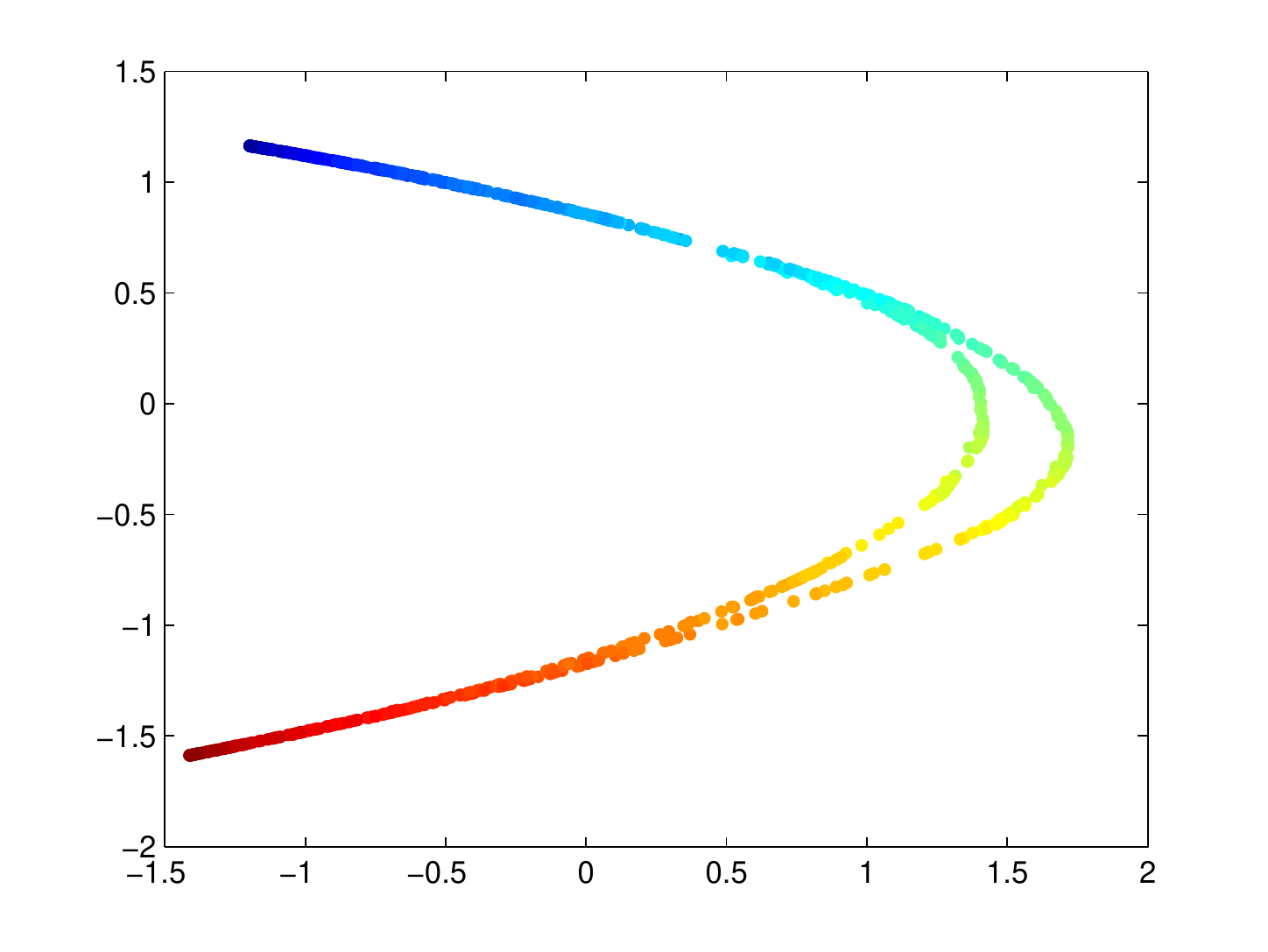}}
    \subfigure[LTSA]{\includegraphics[scale=.3]{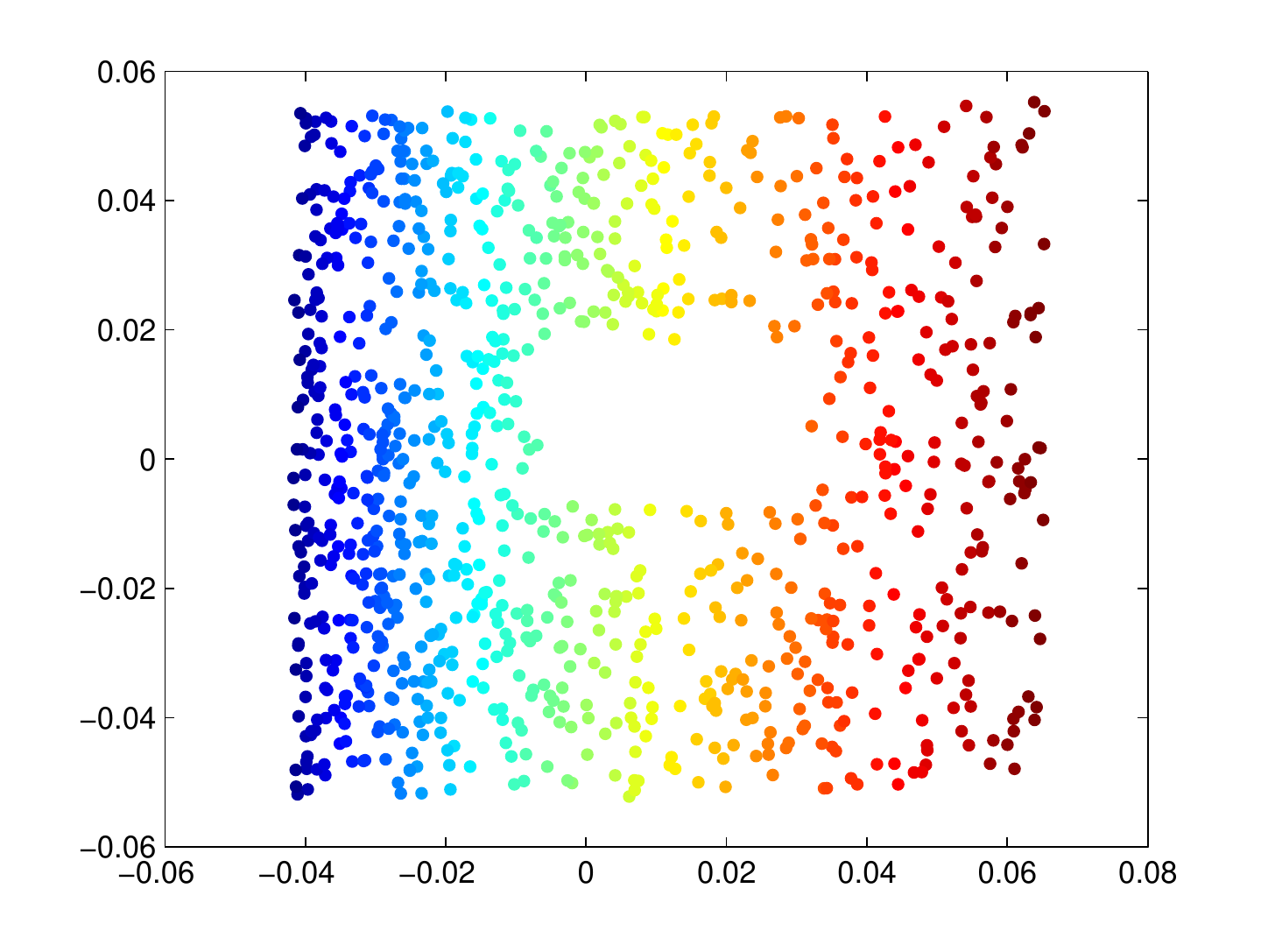}}
    \subfigure[ISOMAP]{\includegraphics[scale=.3]{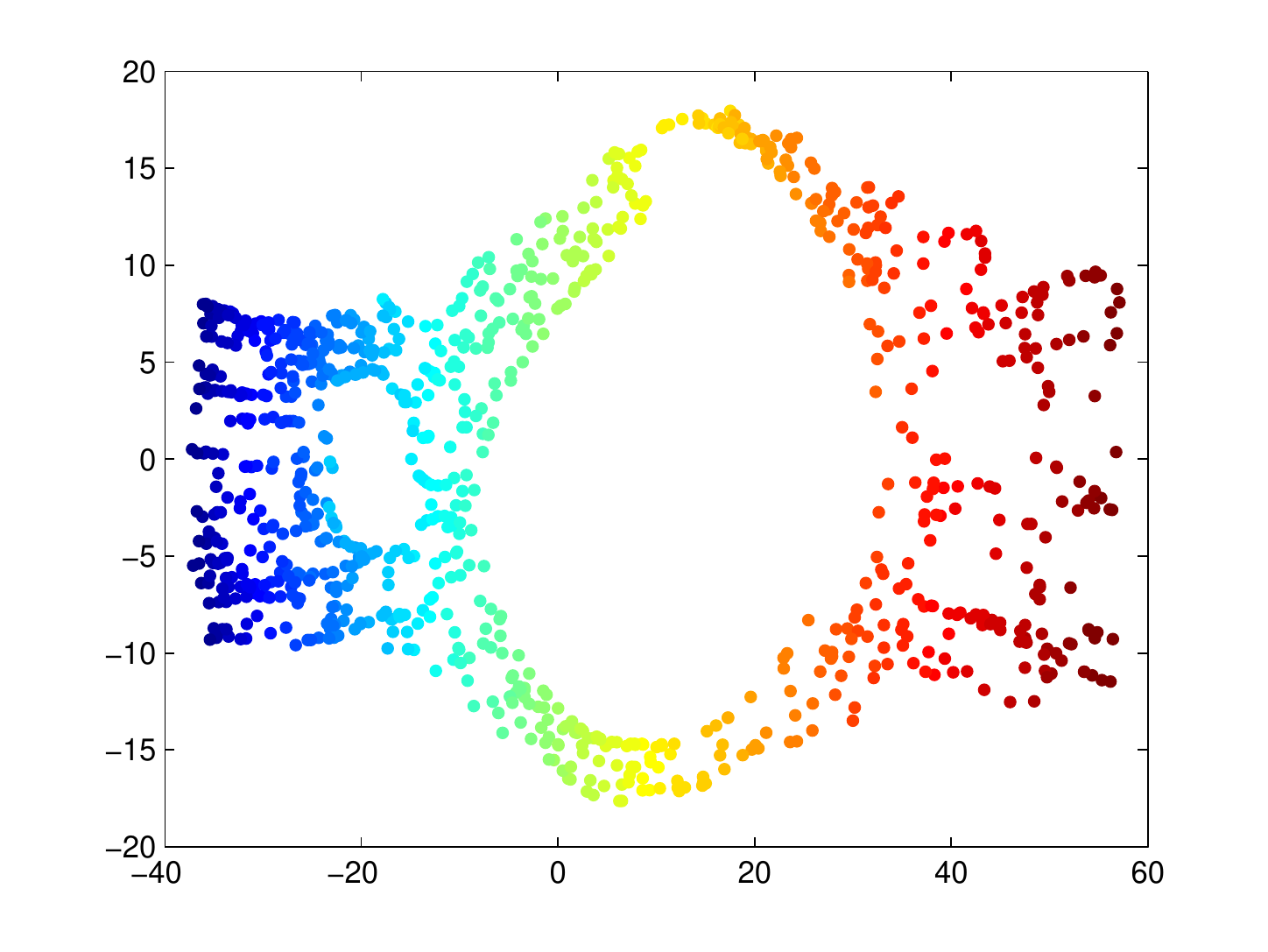}}
    \subfigure[RML]{\includegraphics[scale=.3]{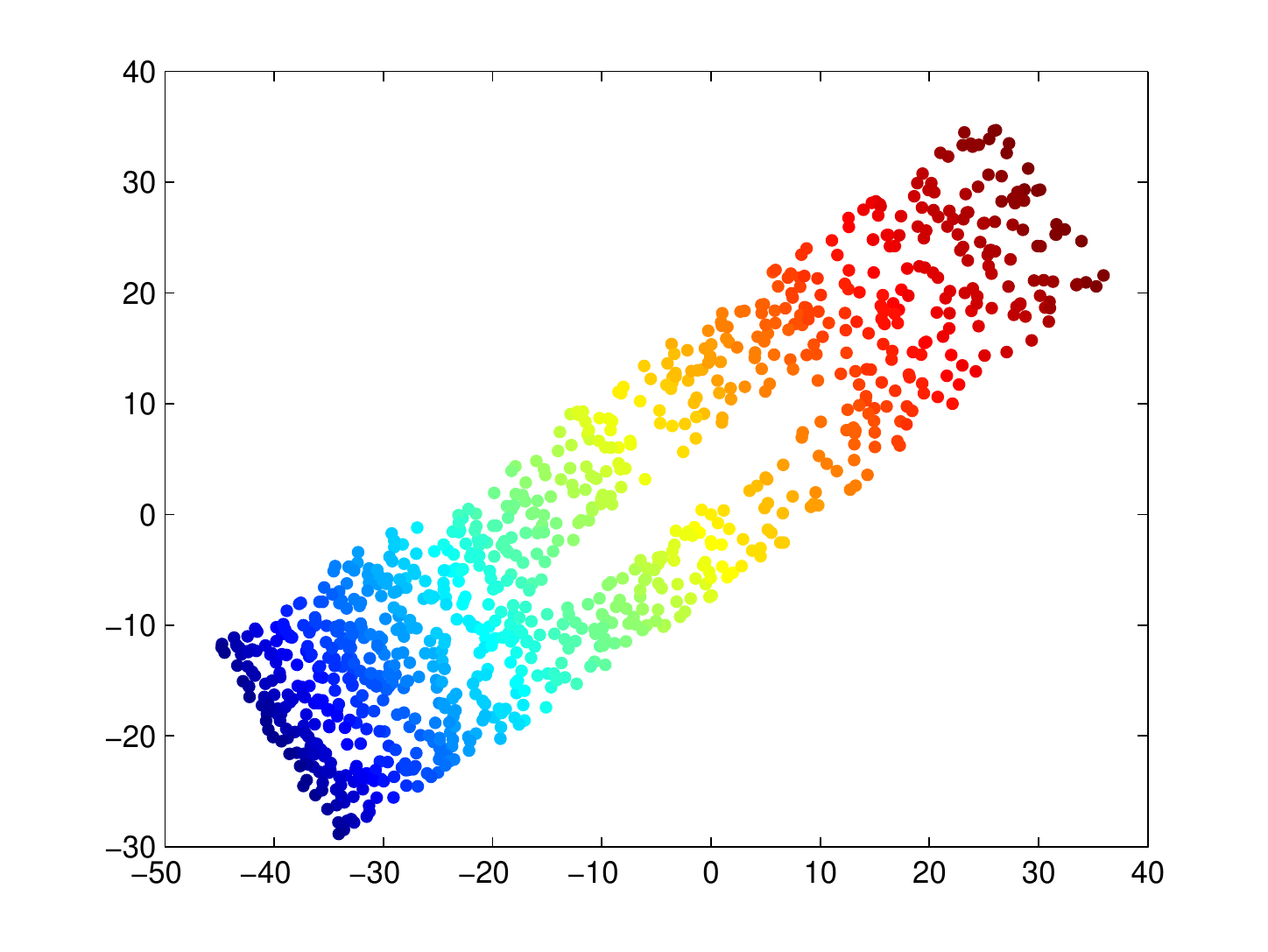}}\\
    \subfigure[$M_P$]{\includegraphics[scale=.3]{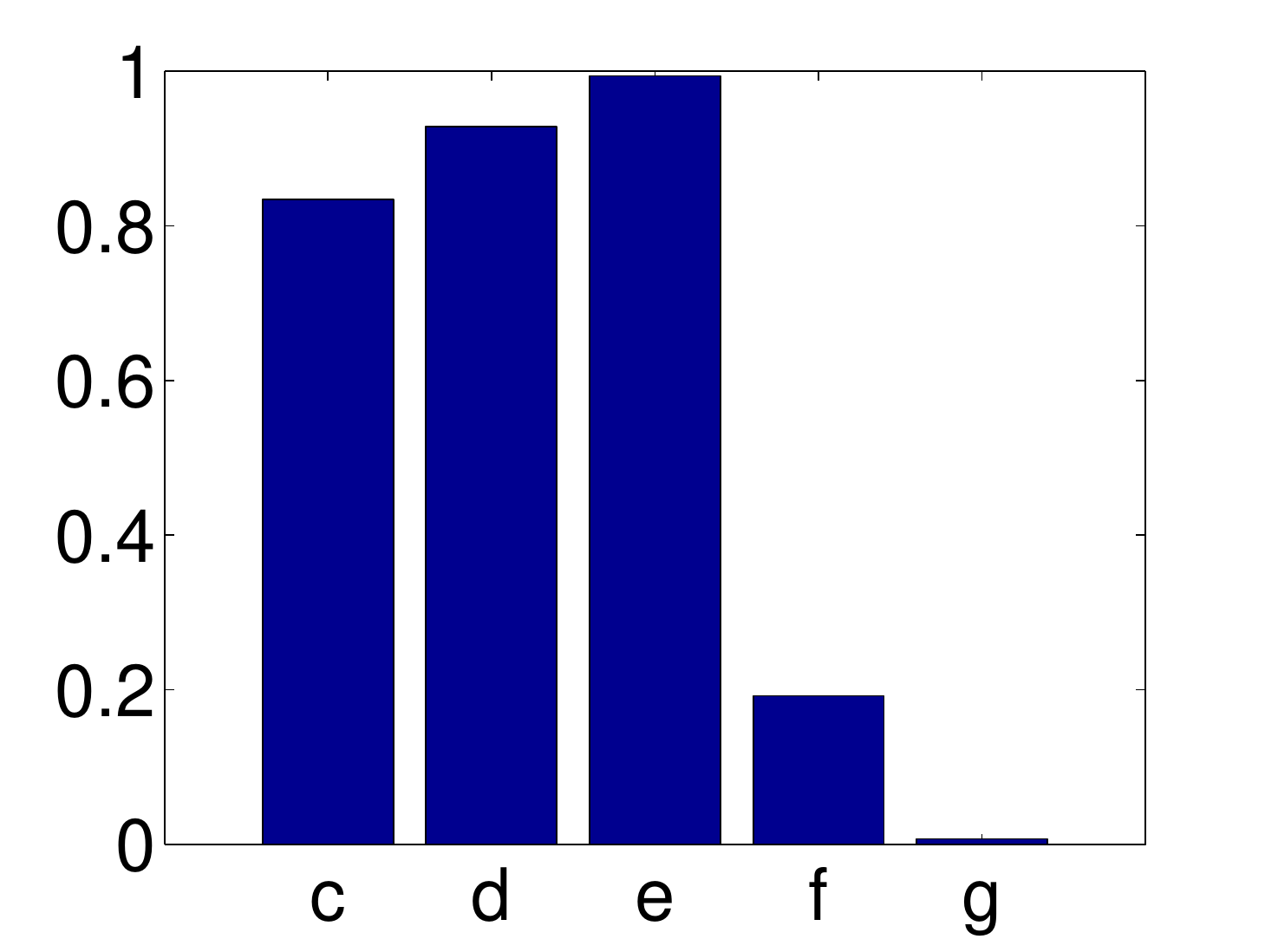}}
    \subfigure[$M_P^c$]{\includegraphics[scale=.3]{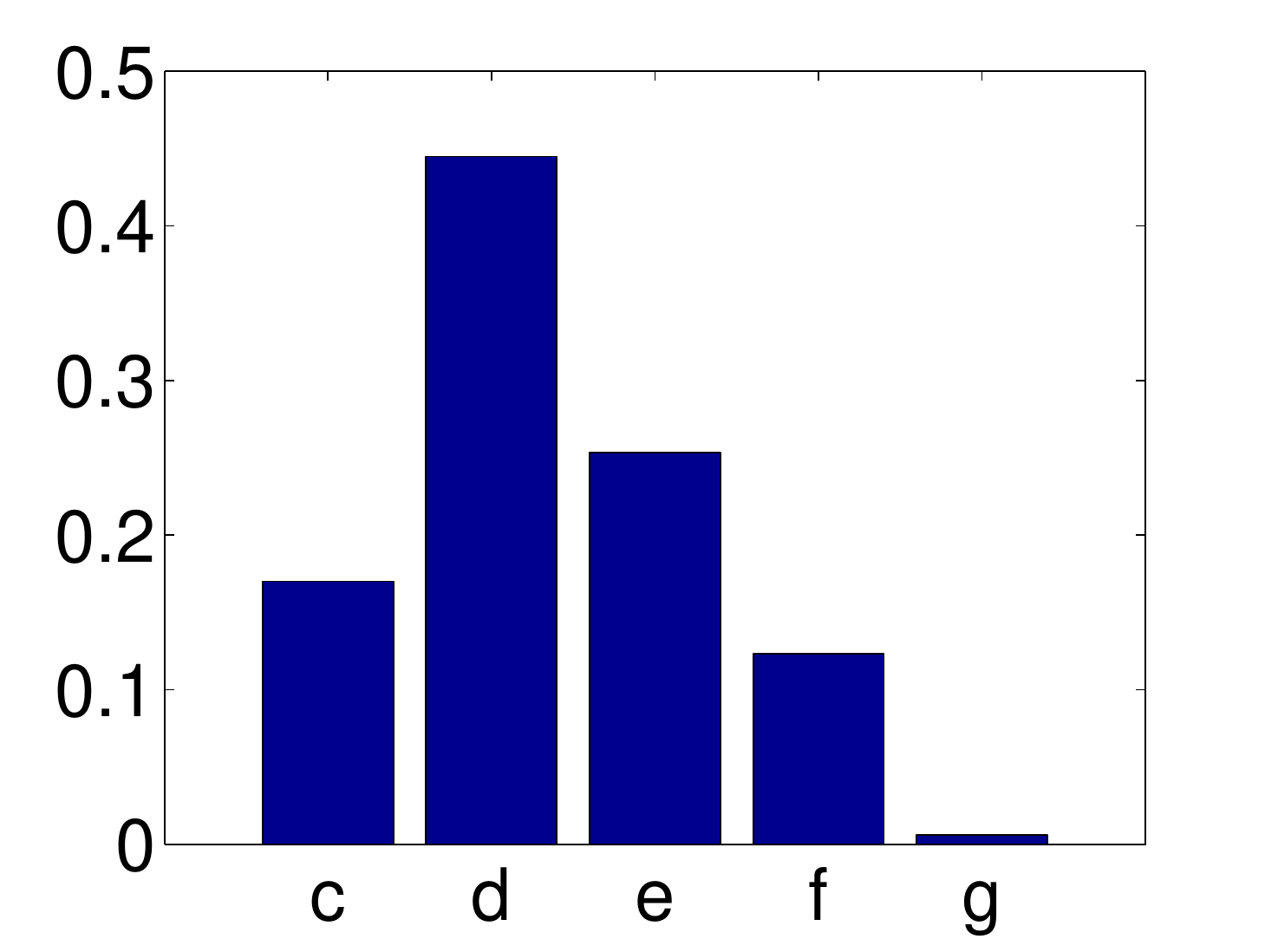}}
    \subfigure[1-$M_{LC}$]{\includegraphics[scale=.3]{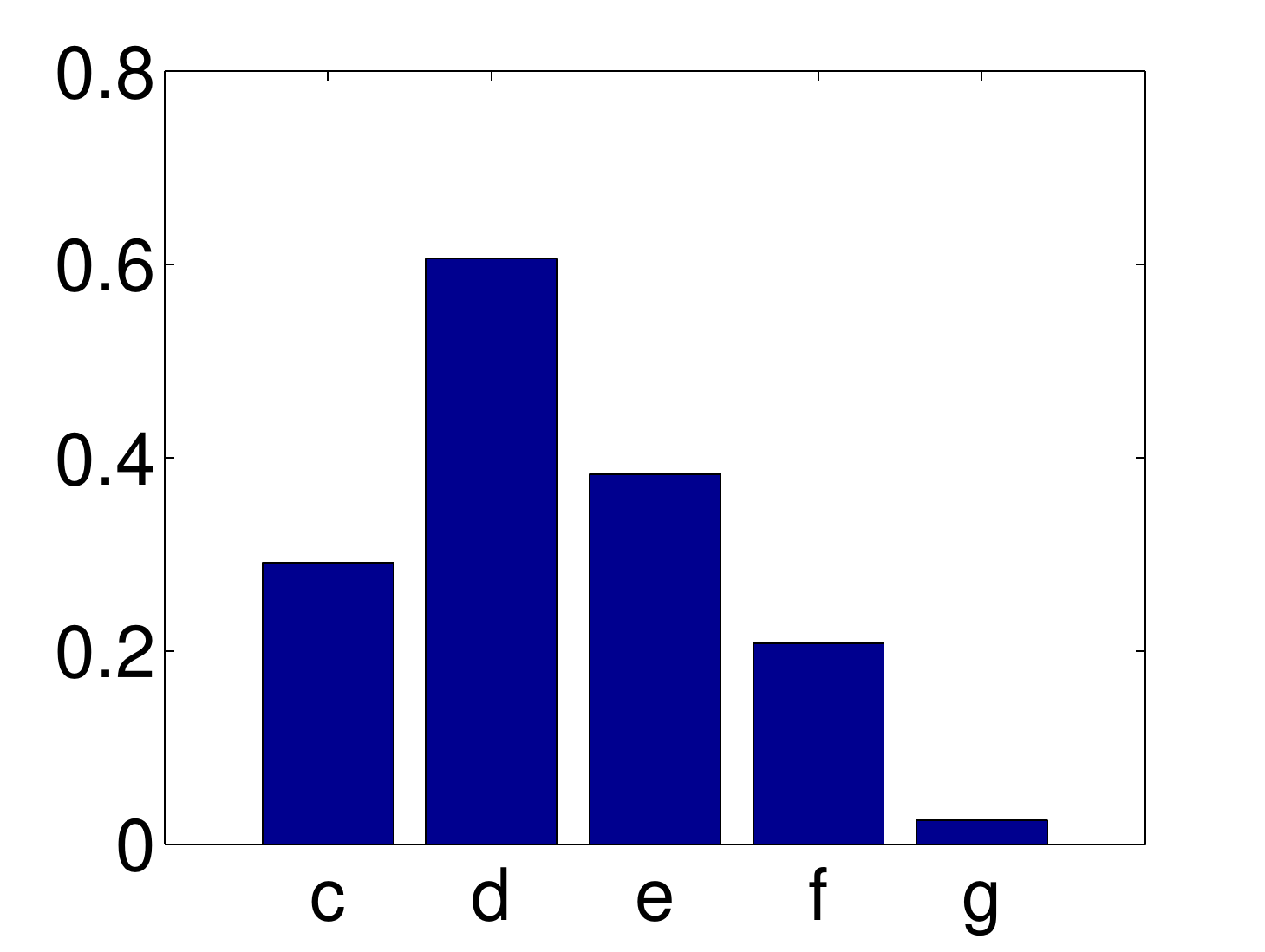}}
    \subfigure[$M_{RV}$]{\includegraphics[scale=.3]{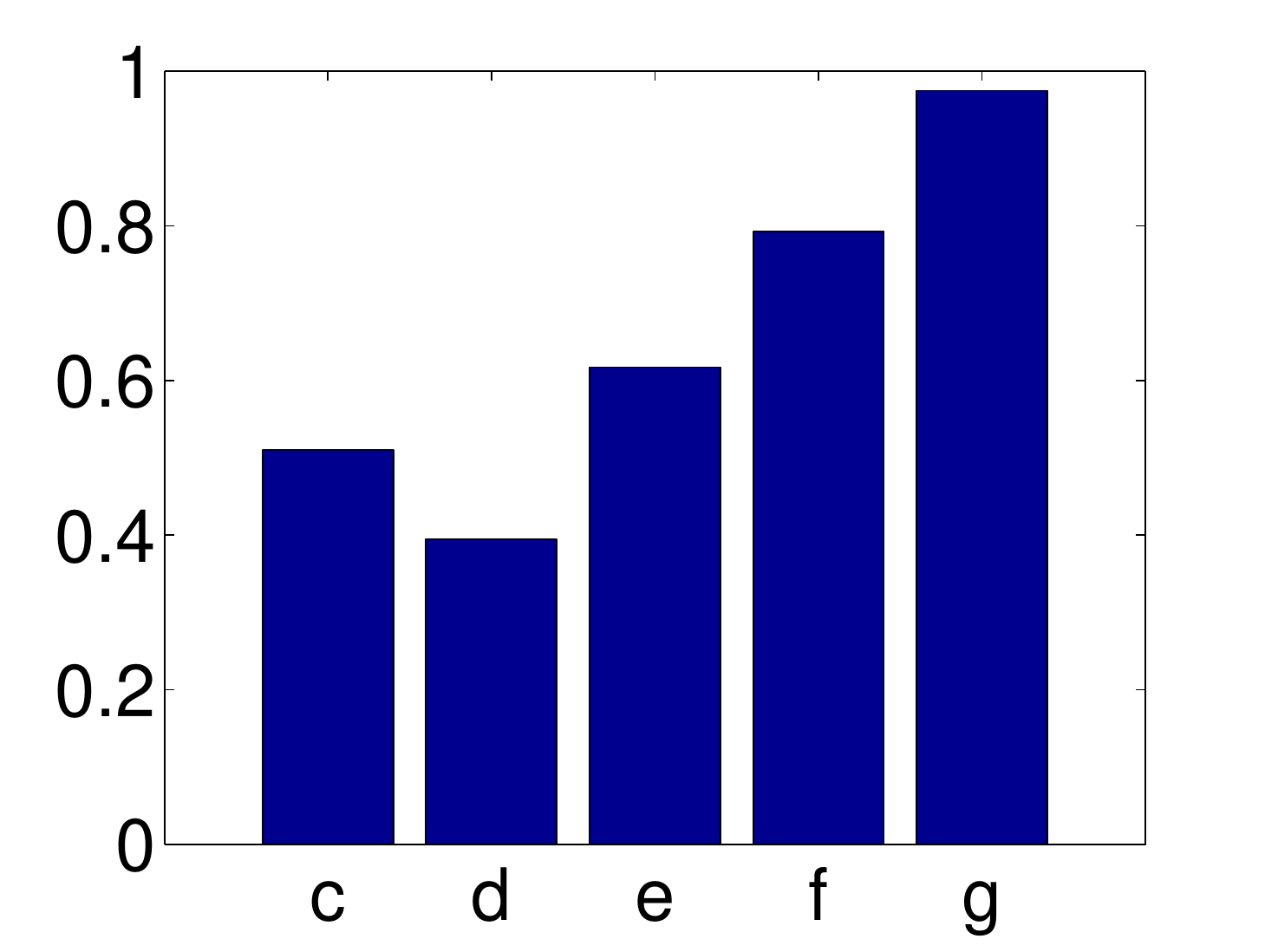}}
    \subfigure[$M_L$]{\includegraphics[scale=.3]{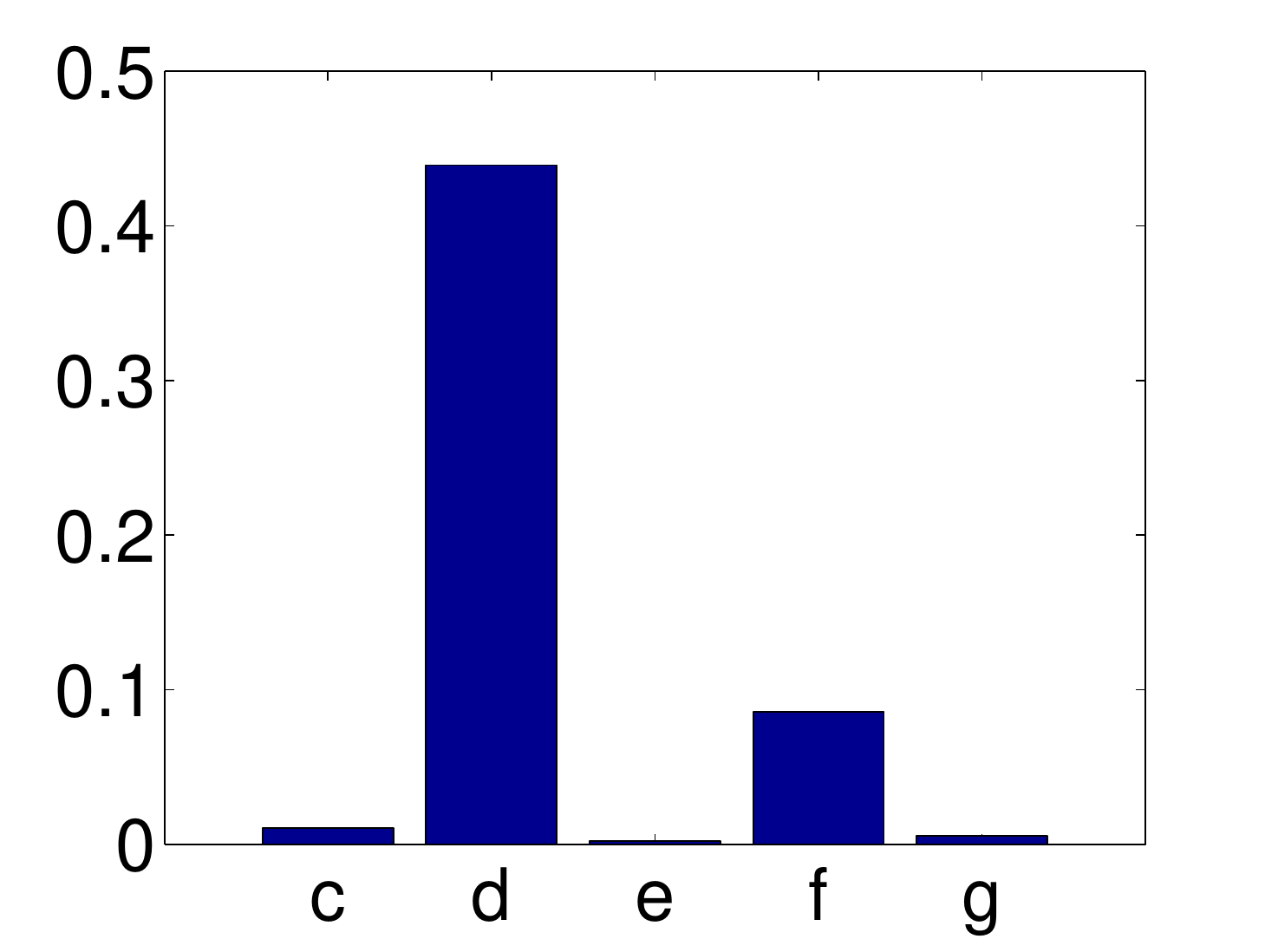}}
    \subfigure[$M_t$]{\includegraphics[scale=.3]{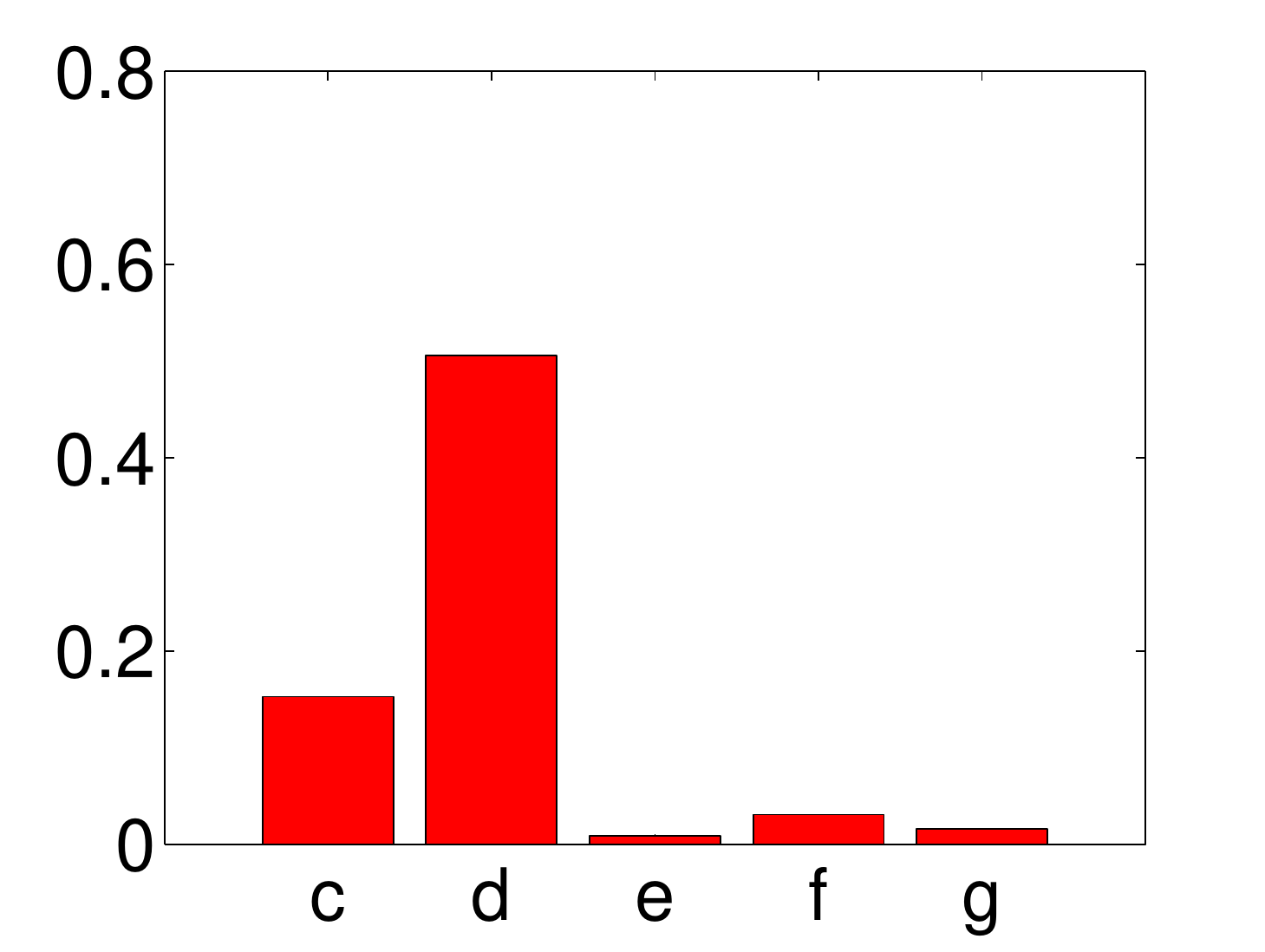}}
    \caption{Manifold learning results on \texttt{Swisshole}. (a) Training data $\mathcal{X}$. (b) Groundtruth of intrinsic degrees of freedom $\mathcal{U}$. (c)-(g) Embeddings learned by various method. The name of each method is stated below each subfigure. (h)-(m) Bar plots of
    different assessments on learned embeddings. The lower-case character under each bar corresponds to the index of the subfigure above.}
    \label{fig:NIEQA-me-sh}
\end{figure*}


Similar to the first experiment, we apply NIEQA to model evaluation of the
\texttt{Swisshole} manifold, which shares the same parameter equation to
\texttt{Swissroll}. The difference is that the set of intrinsic degree of
freedoms $\mathcal{U}$ is no longer a convex set, where a rectangular region in
$\mathcal{U}$ is digged out. Therefore, \texttt{Swisshole} manifold is geodesic
non-connected. 1000 training samples are randomly generated from the manifold
and the number of nearest neighbors $k$ is 10. The learned low-dimensional
embeddings and the bar plots of quality assessments are shown in Fig.
\ref{fig:NIEQA-me-sh}.

From Fig. \ref{fig:NIEQA-me-sh}, we can see that LTSA and RML correctly learned
the geometric structure of $\mathcal{U}$ with the highest quality over other
approaches. The embedding given by LLE has a distortion in global shape. ISOMAP
and LE fails to learn the structure of $\mathcal{U}$. From the bar plots in
Figs. \ref{fig:NIEQA-me-sh} (h)-(l), we can see that $M_L$ reports a reasonable
quality assessment and matches $M_t$ well which is illustrated in Fig.
\ref{fig:NIEQA-me-sh}(m). $M_P$ and $M_P^c$ works only for isometric embeddings
provided by ISOMAP and RML. $M_{LC}$ and $M_{RV}$ fails to report reasonable
evaluations. Since \texttt{Swisshole} manifold is geodesic non-connected, using
shortest path length would fail to approximate geodesci distance. Therefore, we
do not compute the global assessment $M_G$ in NIEQA.

\begin{figure*}[!t]
    \centering
    \subfigure[$\mathcal{X}$]{\includegraphics[scale=.3]{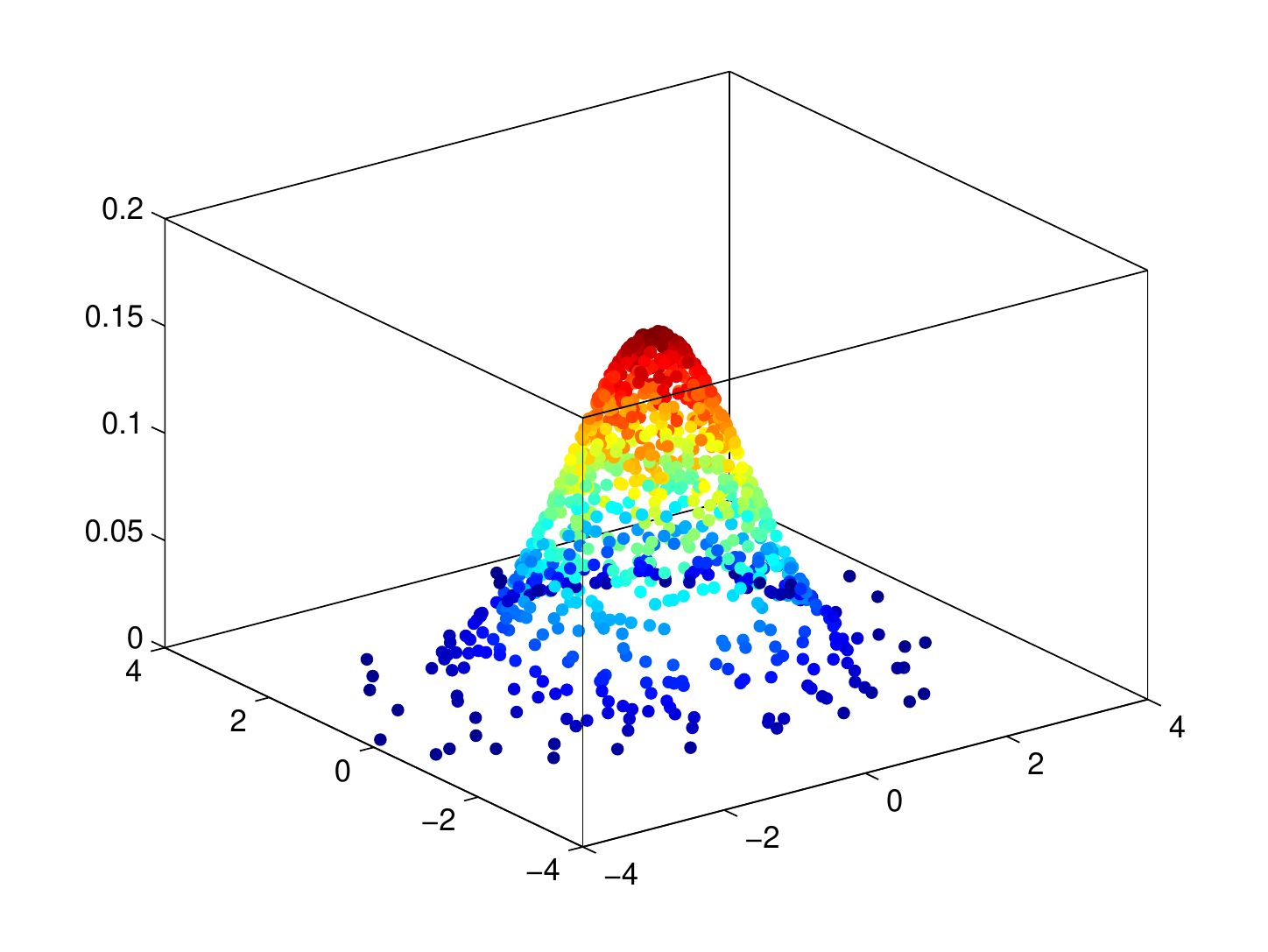}}
    \subfigure[$\mathcal{U}$]{\includegraphics[scale=.3]{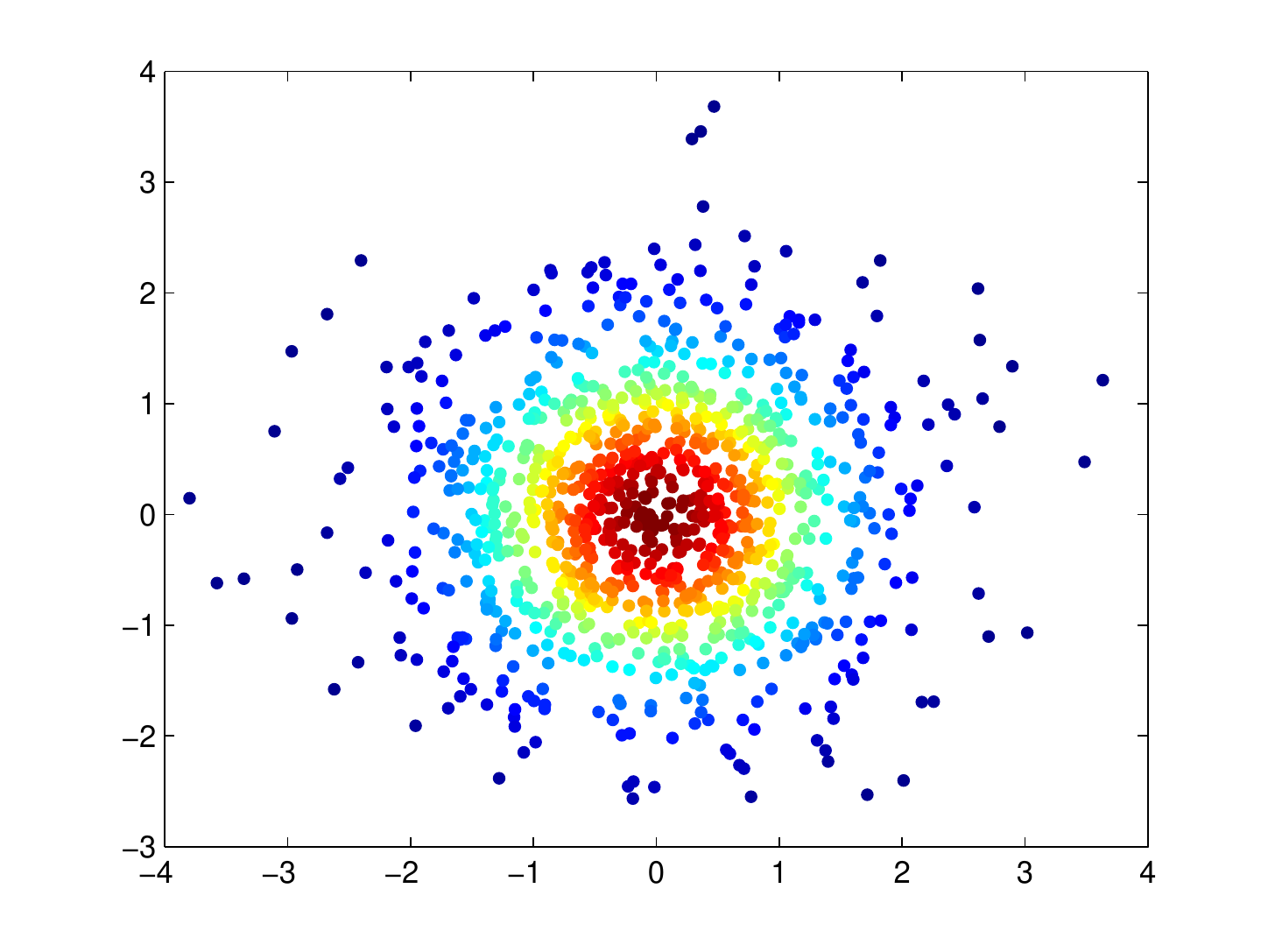}}
    \subfigure[LLE]{\includegraphics[scale=.3]{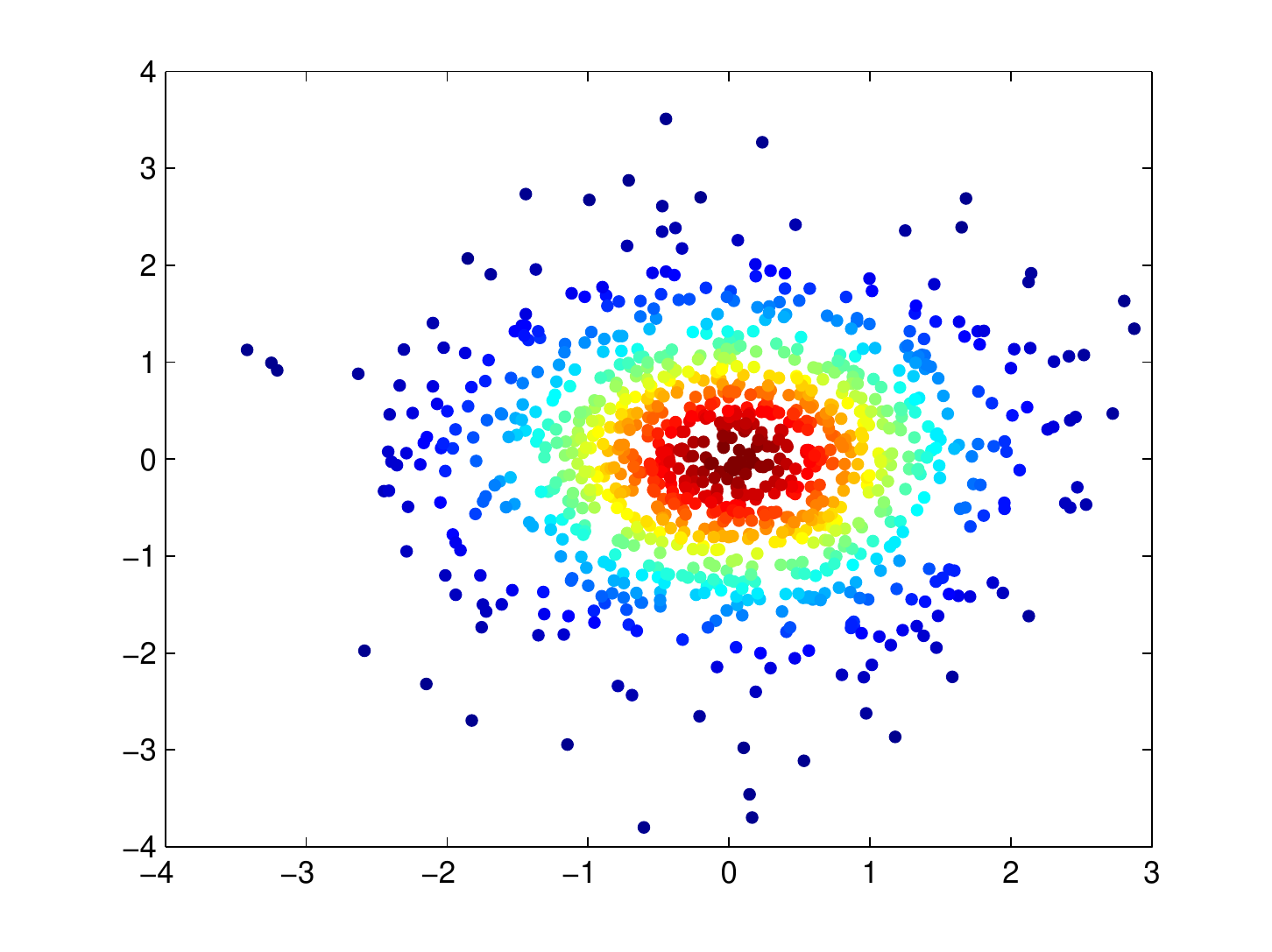}}
    \subfigure[LE]{\includegraphics[scale=.3]{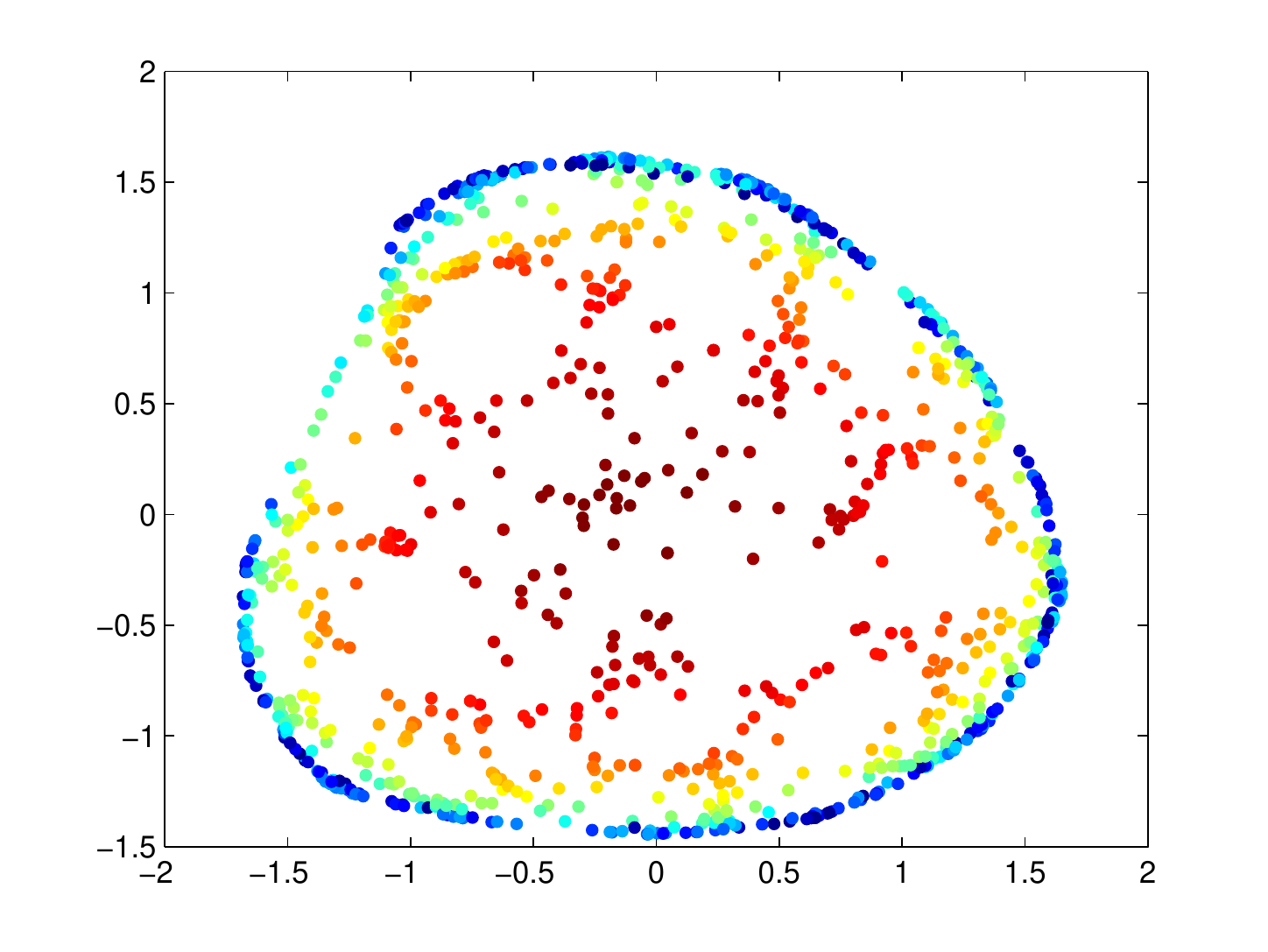}}
    \subfigure[LTSA]{\includegraphics[scale=.3]{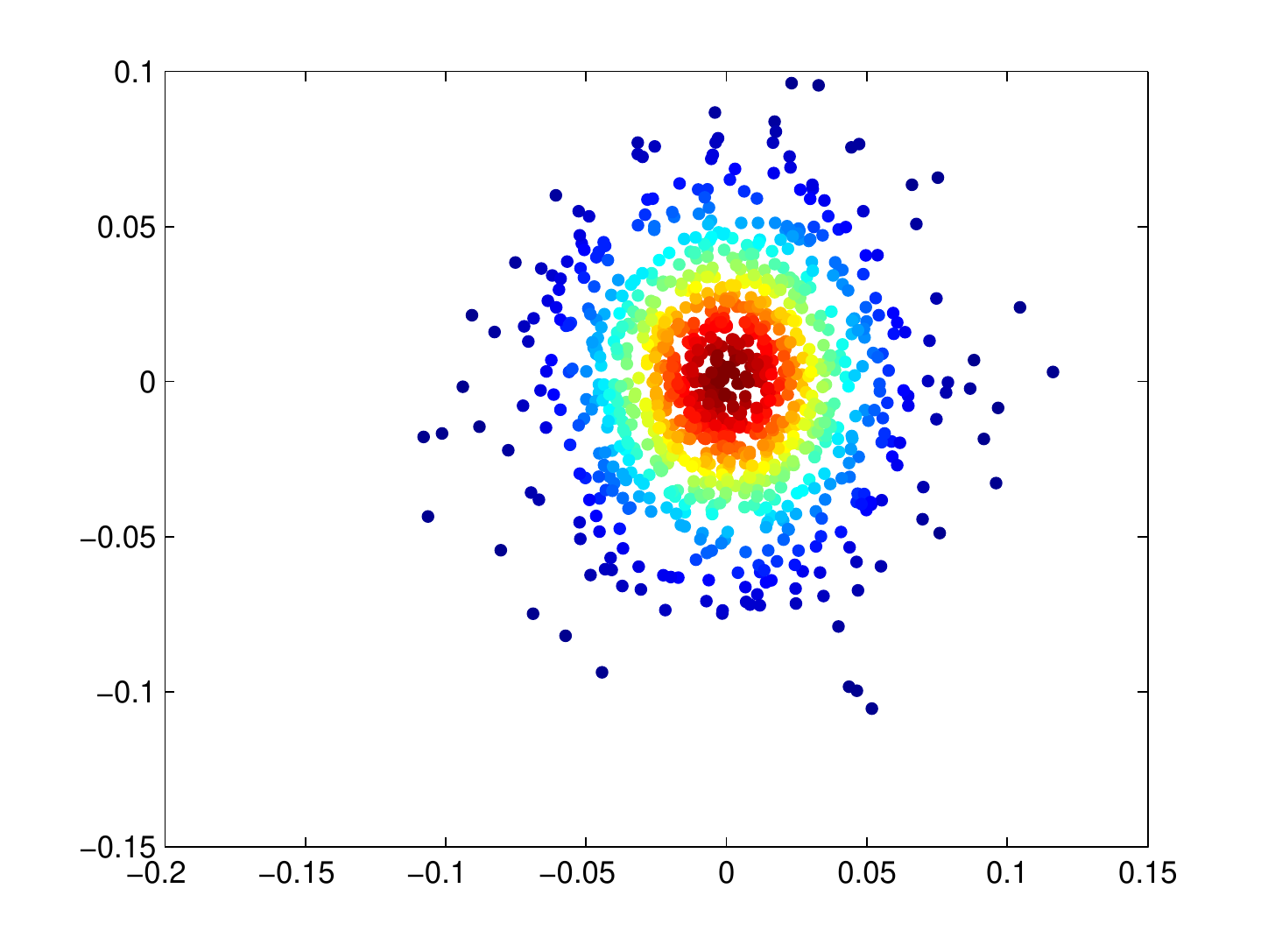}}
    \subfigure[ISOMAP]{\includegraphics[scale=.3]{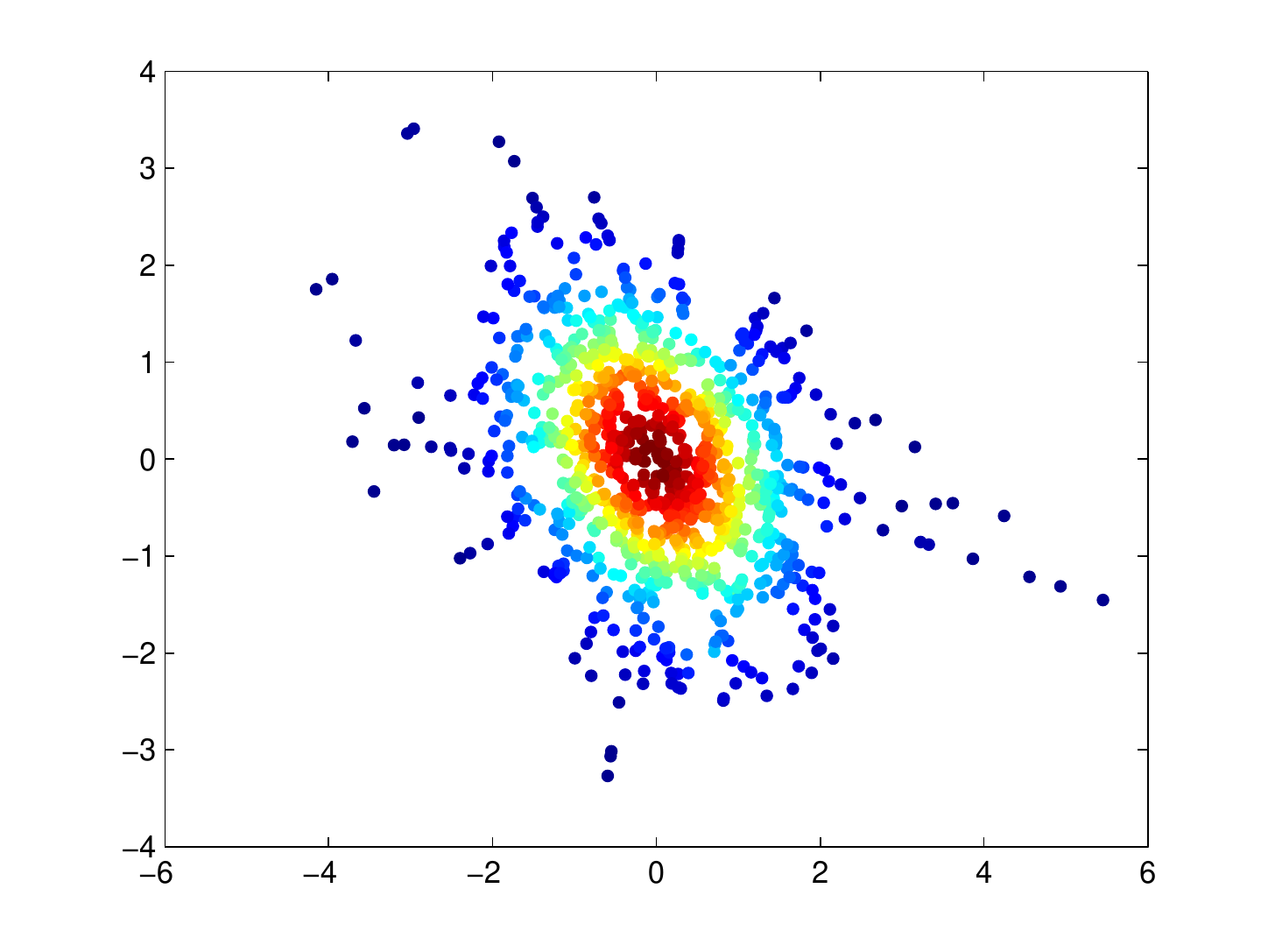}}
    \subfigure[RML]{\includegraphics[scale=.3]{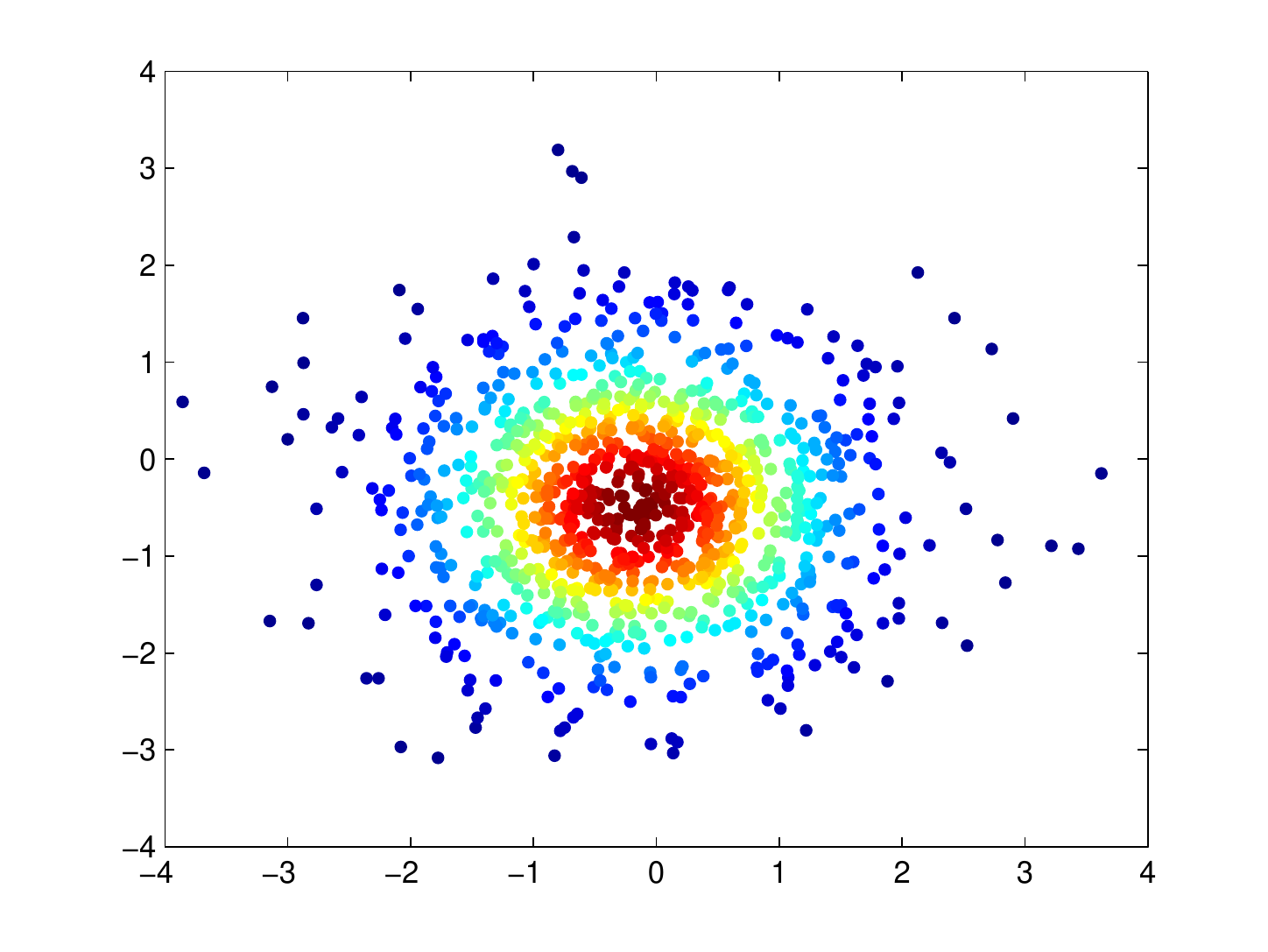}}\\
    \subfigure[$M_P$]{\includegraphics[scale=.3]{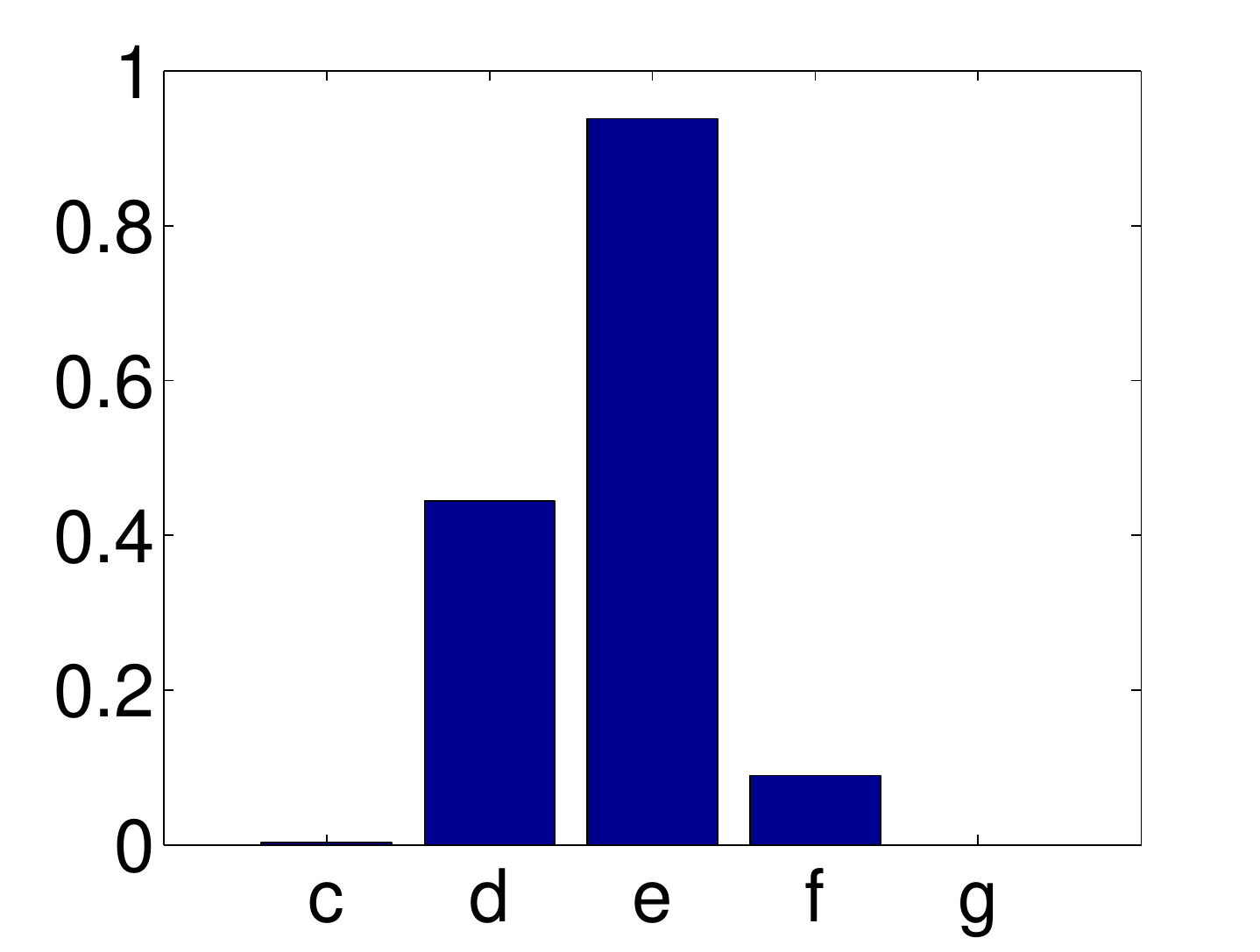}}
    \subfigure[$M_P^c$]{\includegraphics[scale=.3]{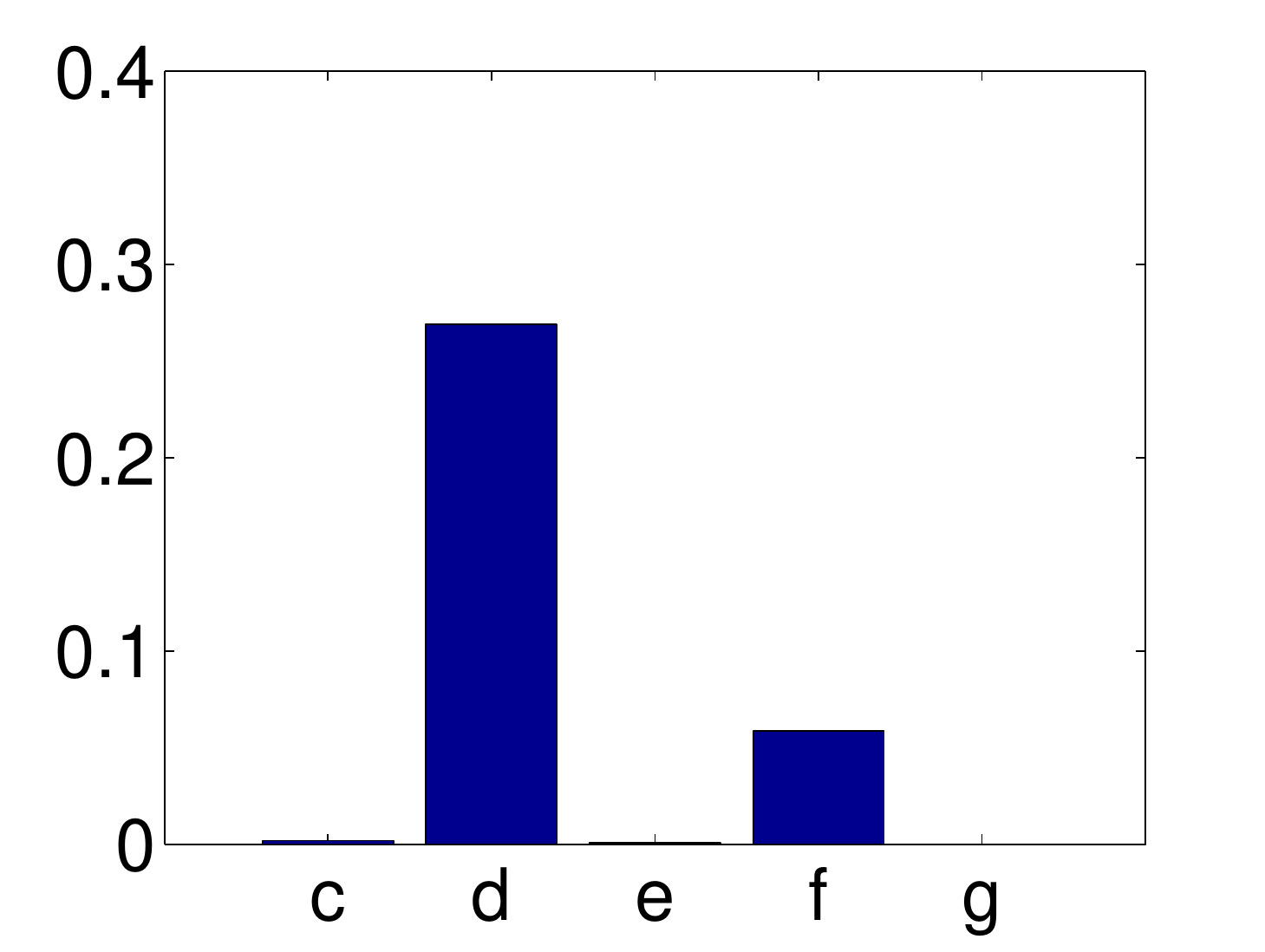}}
    \subfigure[1-$M_{LC}$]{\includegraphics[scale=.3]{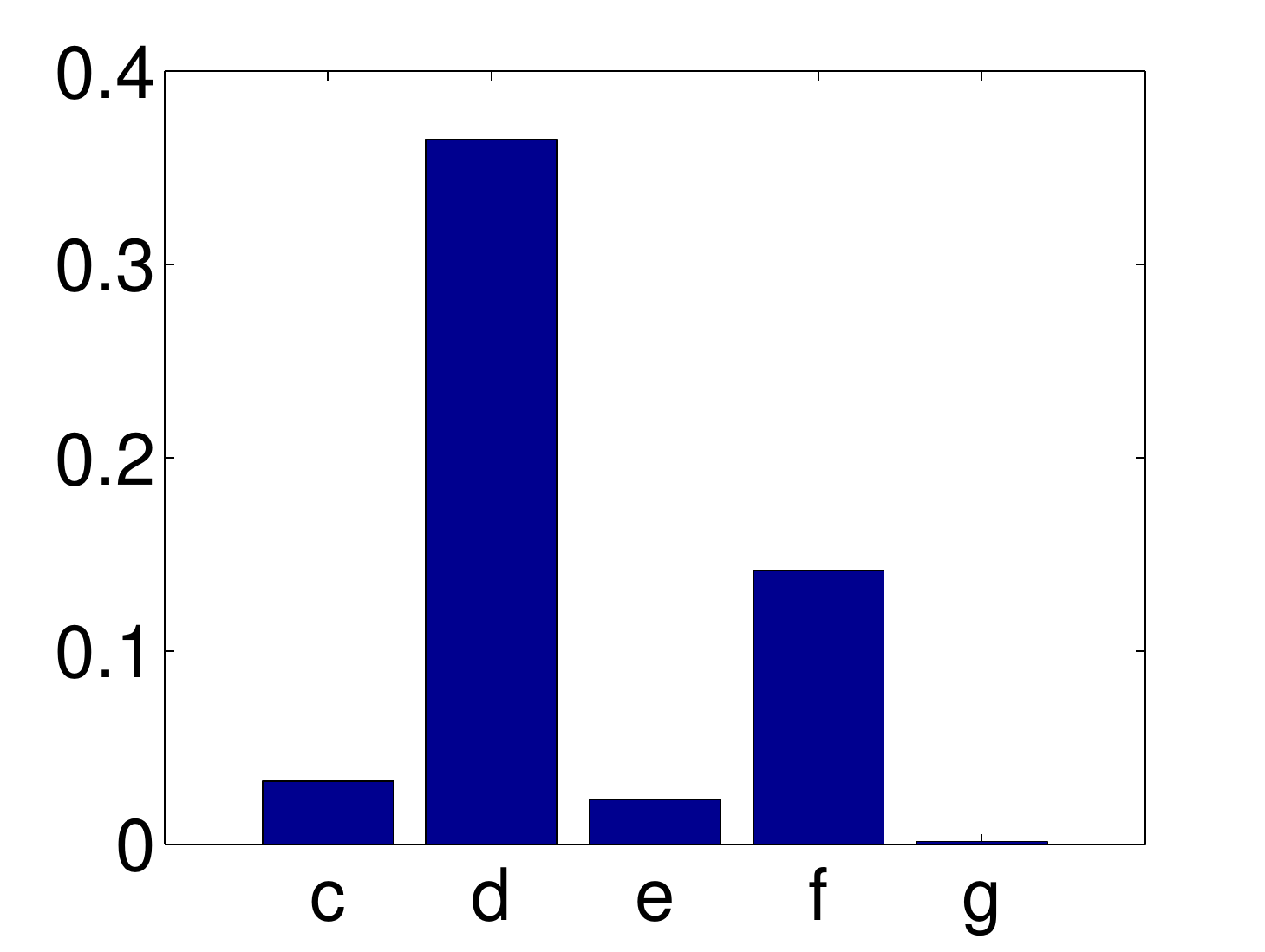}}
    \subfigure[$M_{RV}$]{\includegraphics[scale=.3]{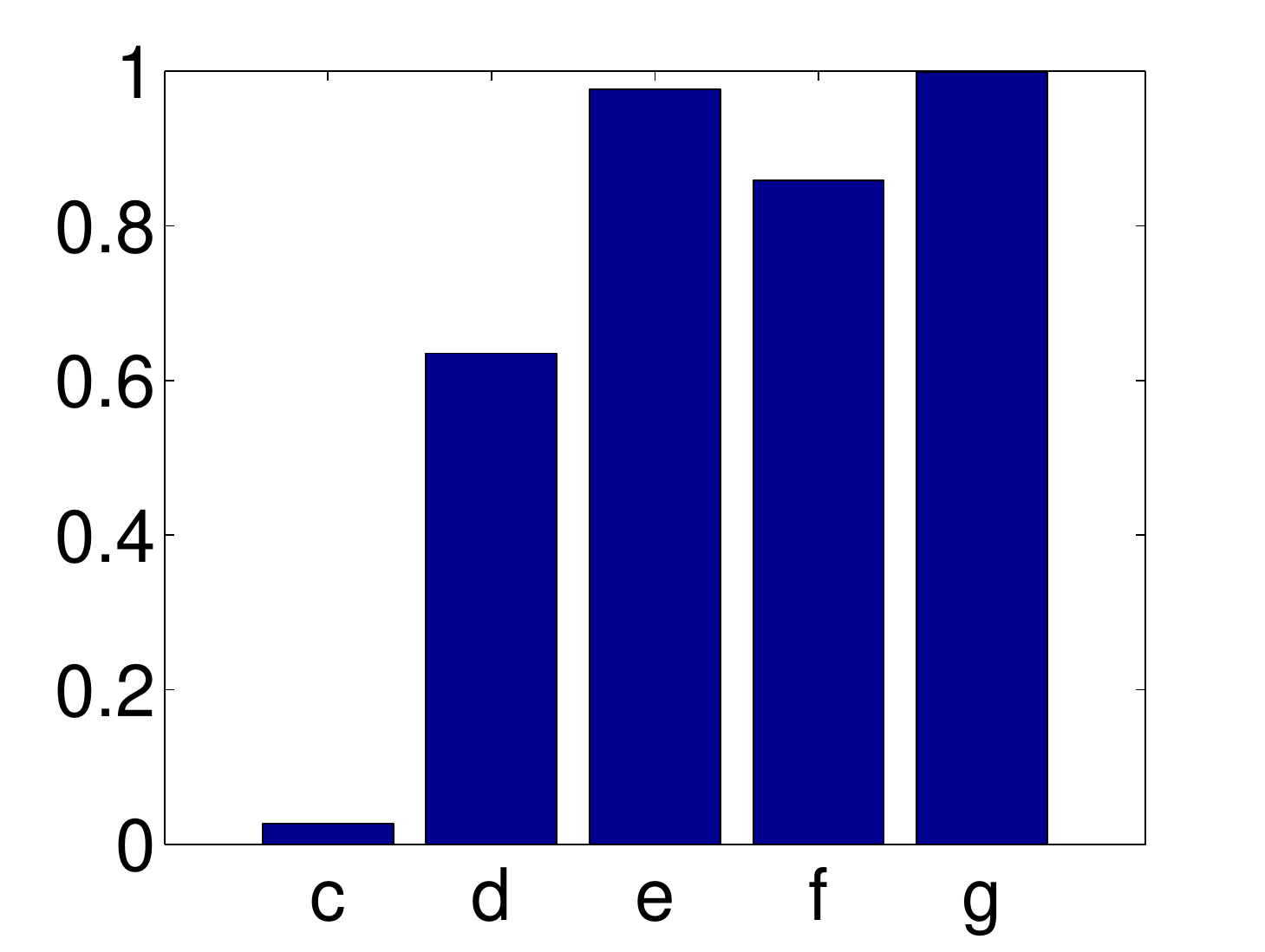}}
    \subfigure[$M_L$]{\includegraphics[scale=.3]{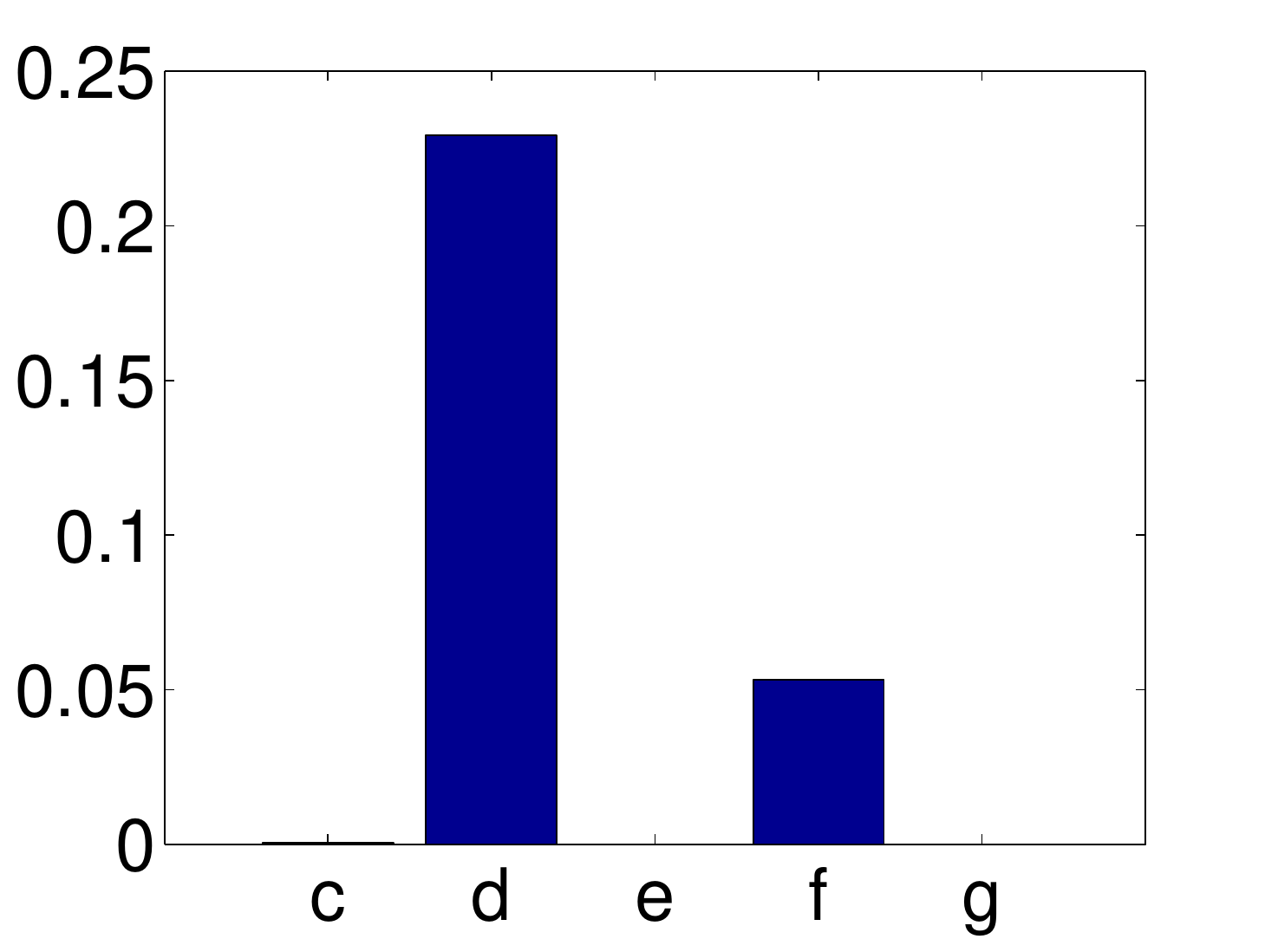}}
    \subfigure[$M_G$]{\includegraphics[scale=.3]{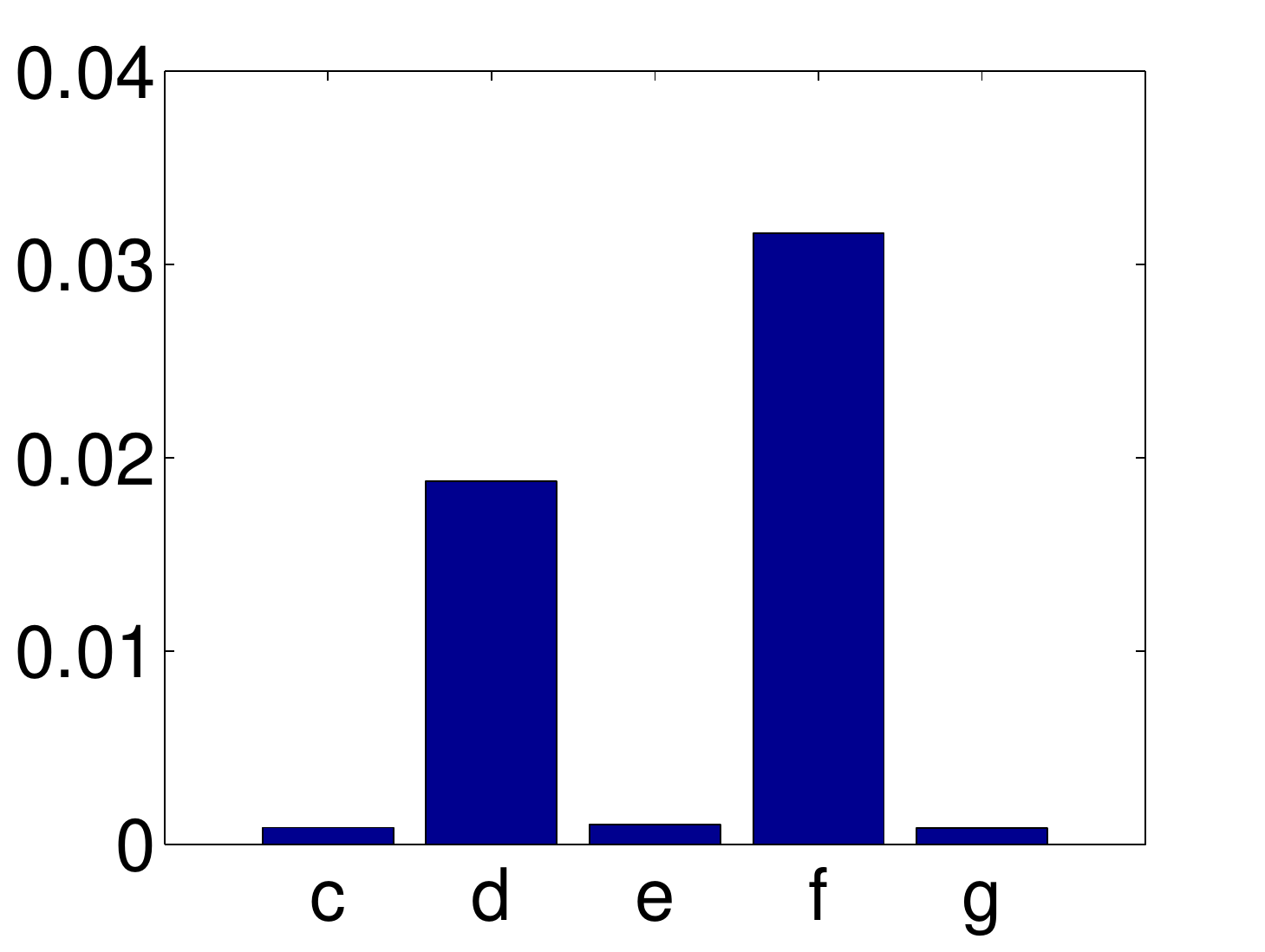}}
    \subfigure[$M_t$]{\includegraphics[scale=.3]{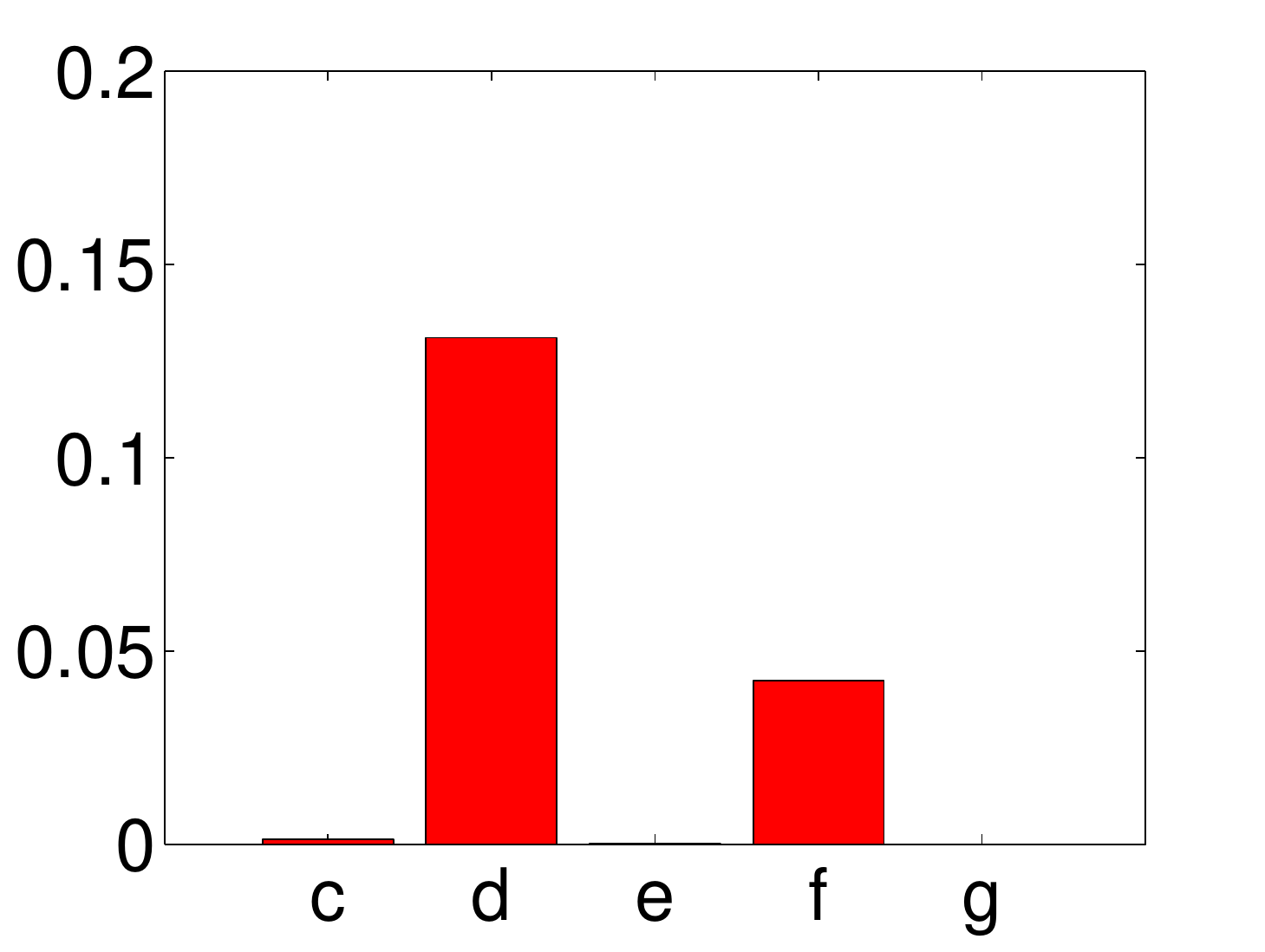}}
    \caption{Manifold learning results on \texttt{Gaussian}. (a) Training data $\mathcal{X}$. (b) Groundtruth of intrinsic degrees of freedom $\mathcal{U}$. (c)-(g) Embeddings learned by various method. The name of each method is stated below each subfigure. (h)-(n) Bar plots of
    different assessments on learned embeddings. The lower-case character under each bar corresponds to the index of the subfigure above.}
    \label{fig:NIEQA-me-gs}
\end{figure*}


In the third experiment, we apply NIEQA to model evaluation of the
\texttt{Gaussian} manifold, whose parameter equation is
\begin{equation*}
    \left\{
    \begin{array}{lll}
        x^1 & = & u^1\\
        x^2 & = & u^2  \\
        x^3 & = & (1/2\pi)\exp\{-((u^1)^2+(u^2)^2)/2\}
    \end{array}
    \right.~.
\end{equation*}
1000 training samples are randomly generated from the manifold and the number
of nearest neighbors $k$ is 10. Fig. \ref{fig:NIEQA-me-gs} shows the learned
low-dimensional embeddings as well as bar plots of different quality
assessments.

From Fig. \ref{fig:NIEQA-me-gs}, we can observe that except LE all the other
methods successfully learned the geometric structure of this manifold, whilst
the quality of the embedding given by ISOMAP is a litter worse. From Figs.
\ref{fig:NIEQA-me-gs} (h)-(m), we can see that $M_P^c$ performs well in this
case by eliminating the global scaling factor. This is due to the isotropic
property of this manifold. $M_{LC}$ reports correct evaluations but still leans
against to RML. $M_{RV}$ fails to assess the embeddings correctly. Both the two
assessments in NIEQA successfully evaluate the quality of different embeddings
and match $M_t$ well. Note that the \texttt{Gaussian} surface is isotropic,
hence the measure $M_P^c$ also works. However, for anisotropic surfaces like
$\texttt{Swissroll}$ and $\texttt{Swisshole}$, only removing global scaling
wound not yield a reasonable assessment.

\begin{figure*}
    \centering
    \subfigure[LLE]{\includegraphics[scale=.45]{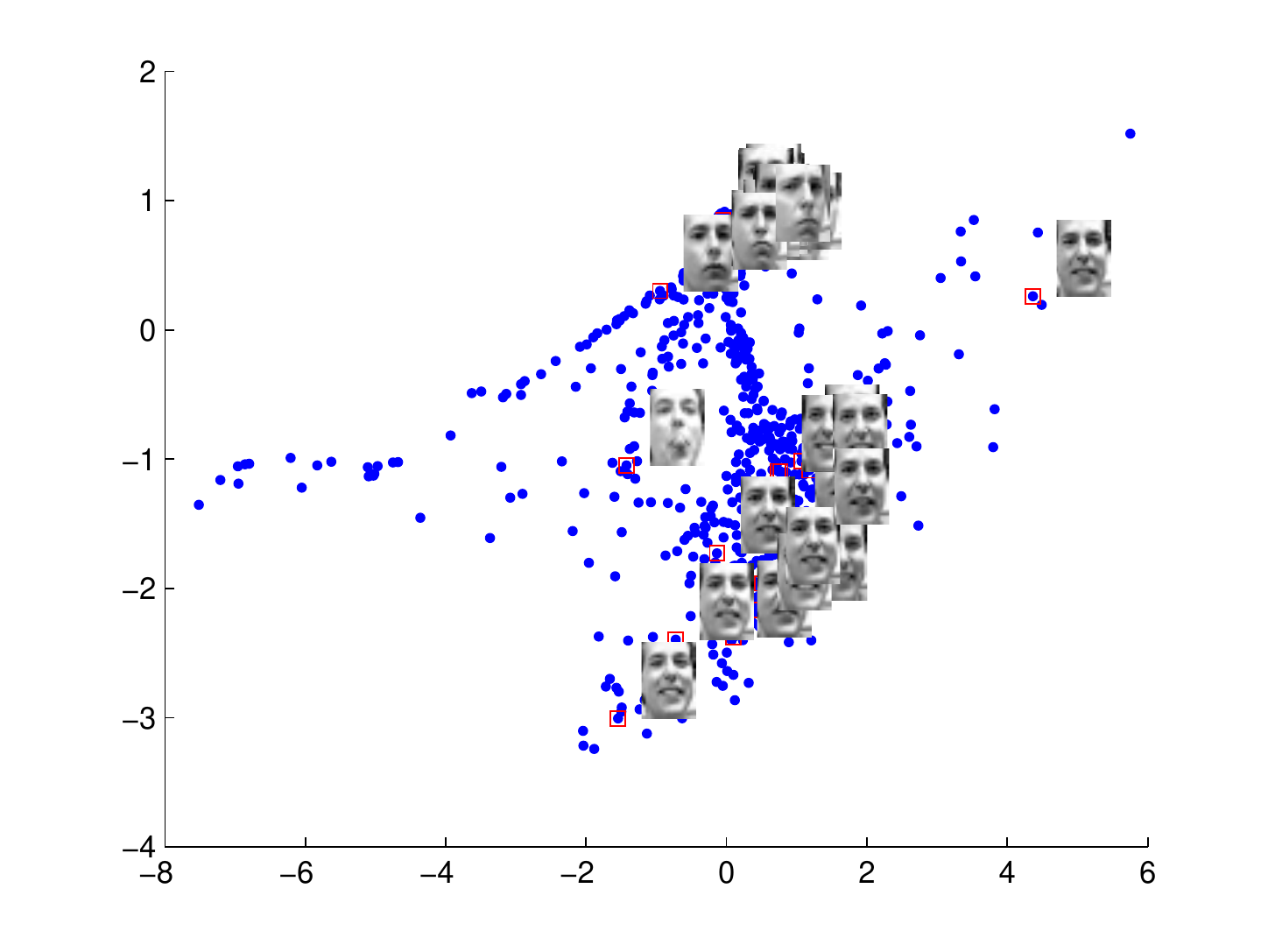}}
    \subfigure[LE]{\includegraphics[scale=.45]{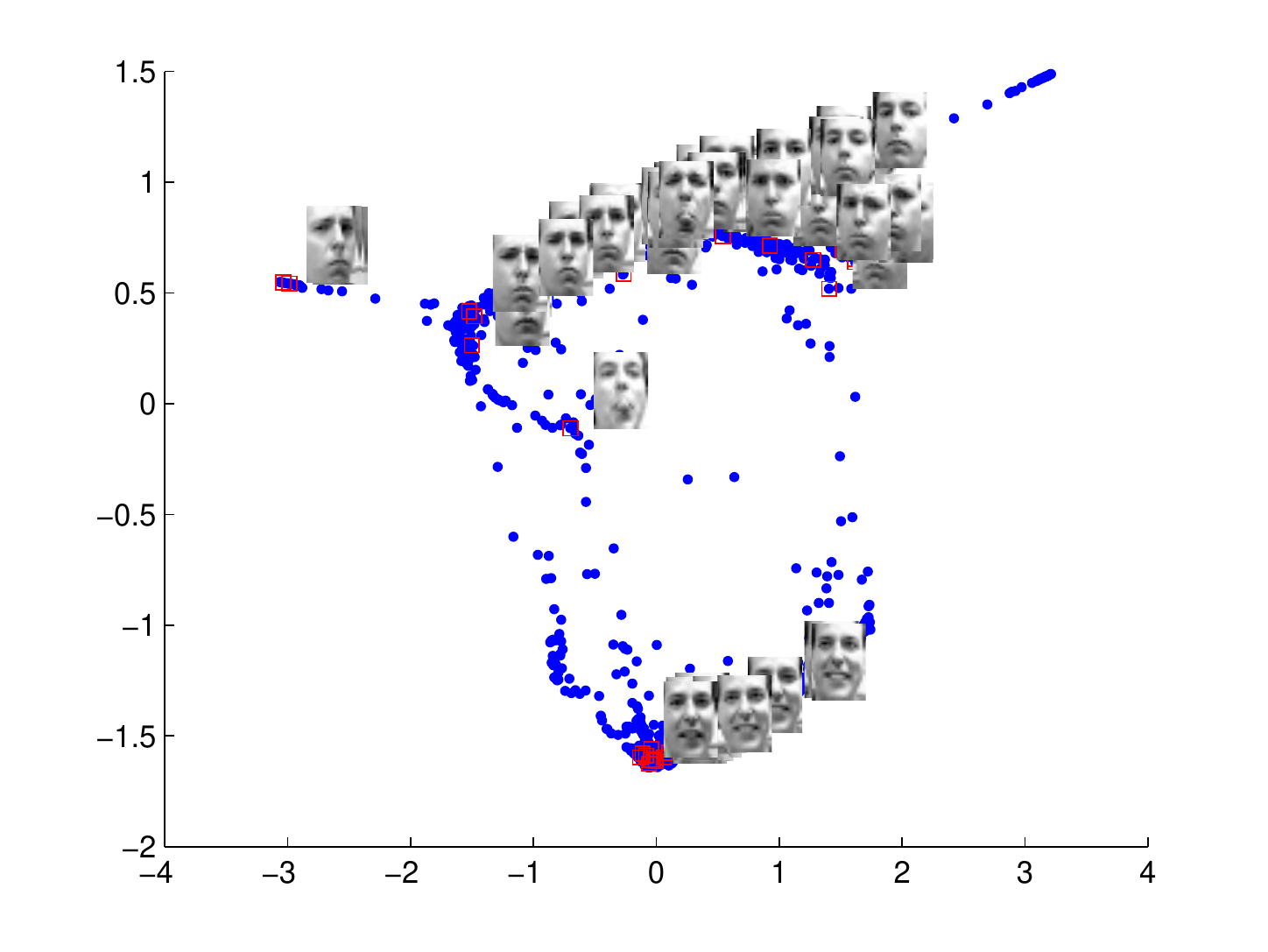}}
    \subfigure[LTSA]{\includegraphics[scale=.45]{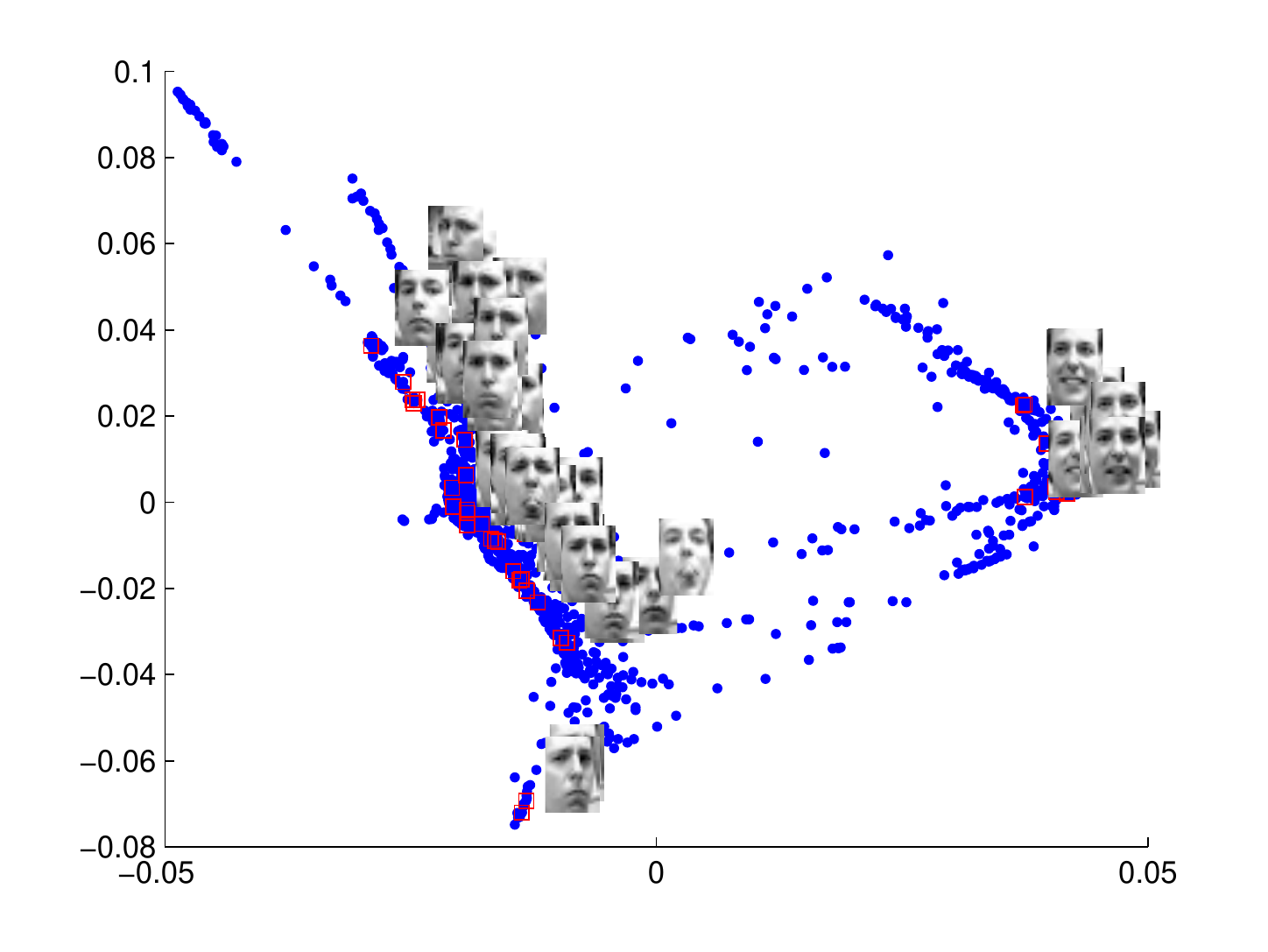}}
    \subfigure[ISOMAP]{\includegraphics[scale=.45]{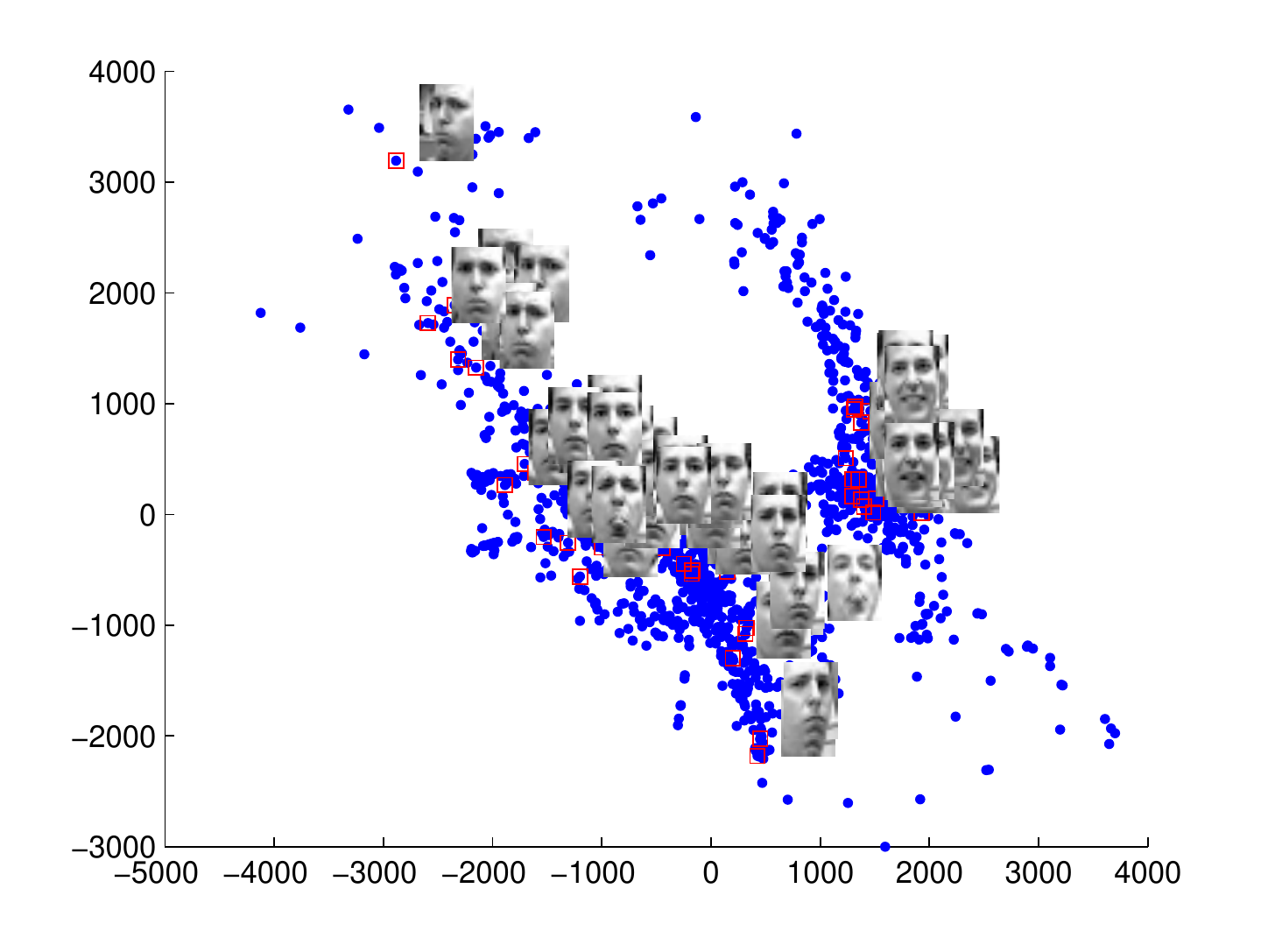}}\\
    \subfigure[$M_P$]{\includegraphics[scale=.3]{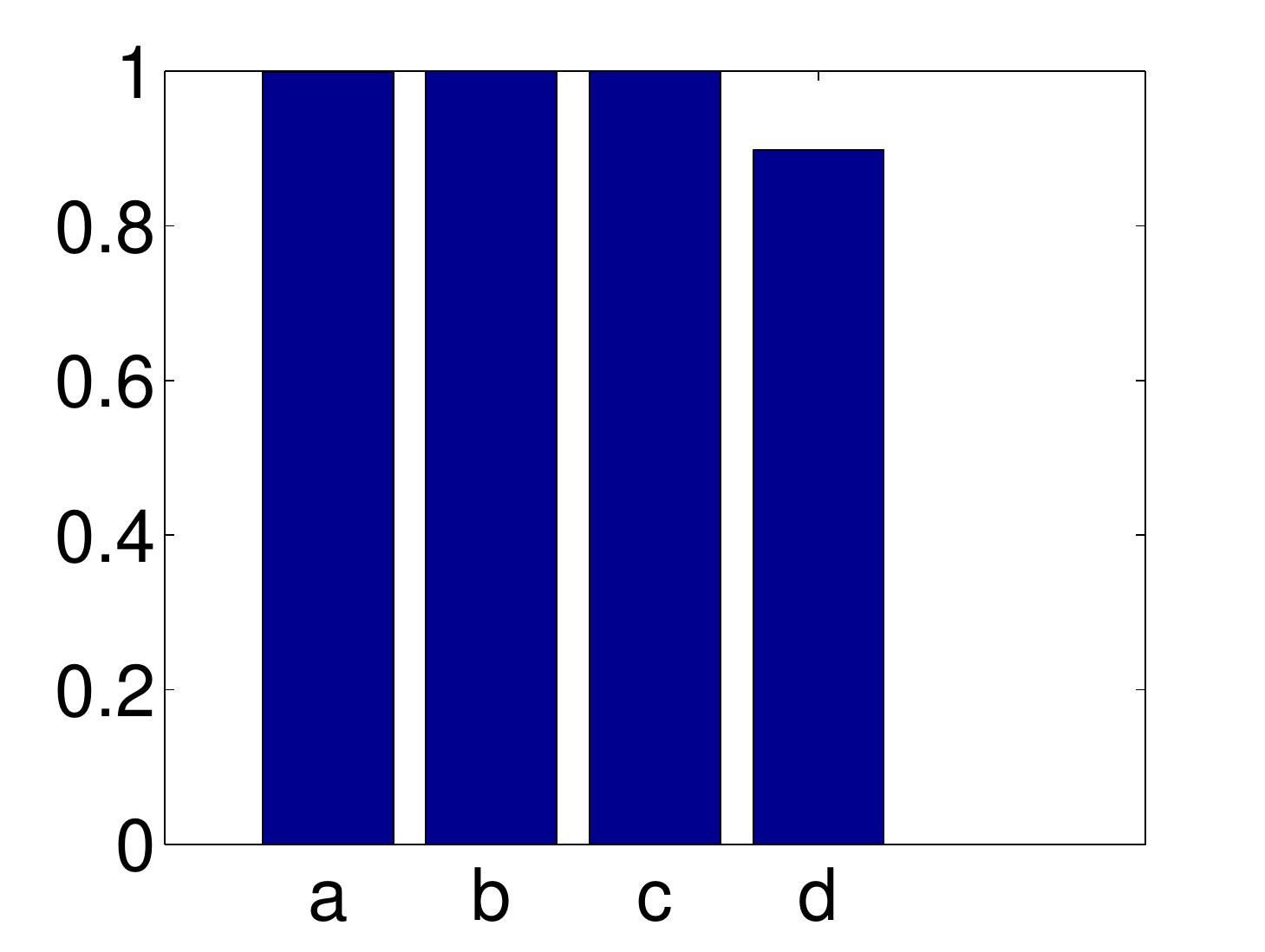}}
    \subfigure[$M_P^c$]{\includegraphics[scale=.3]{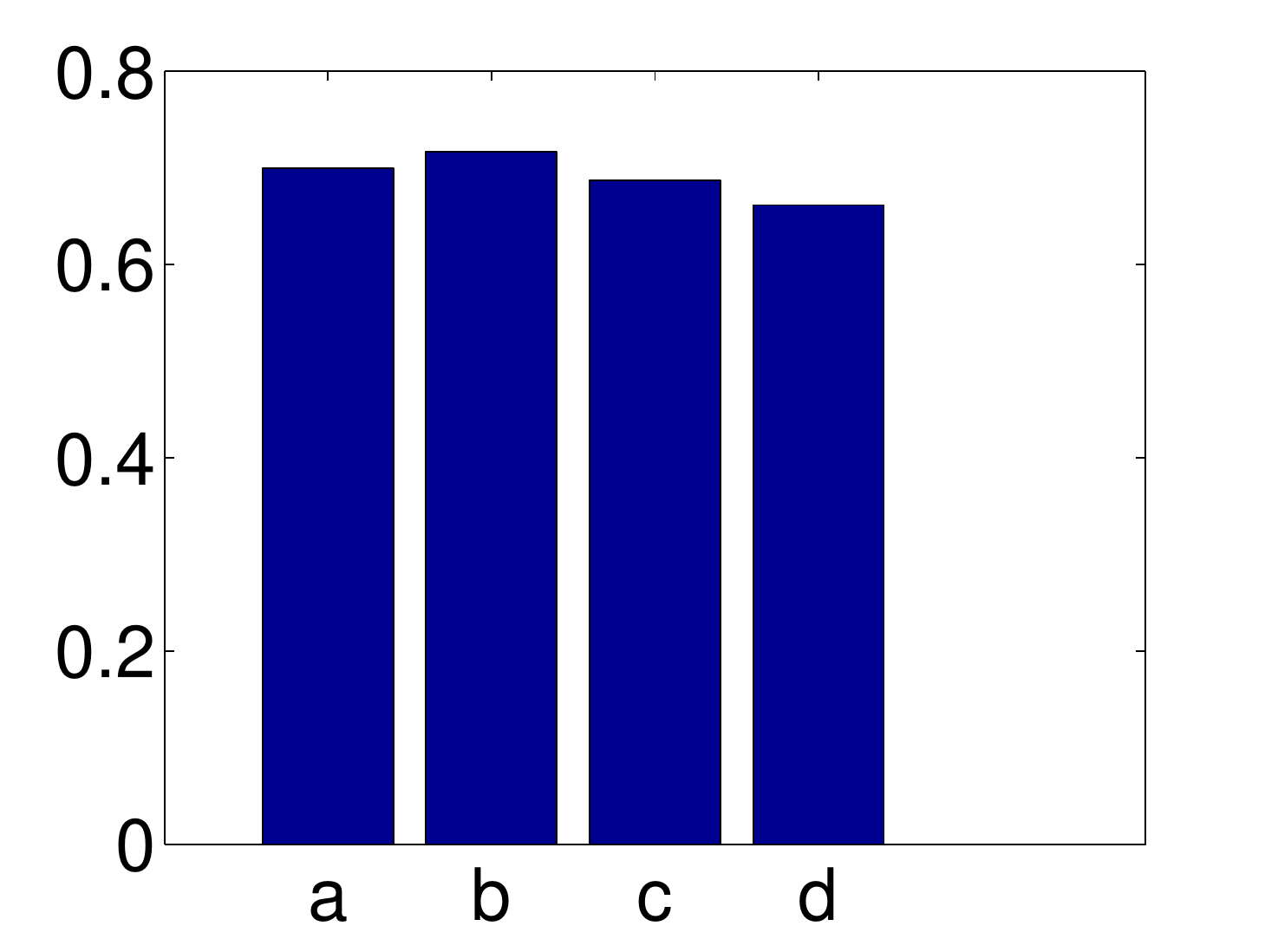}}
    \subfigure[1-$M_{LC}$]{\includegraphics[scale=.3]{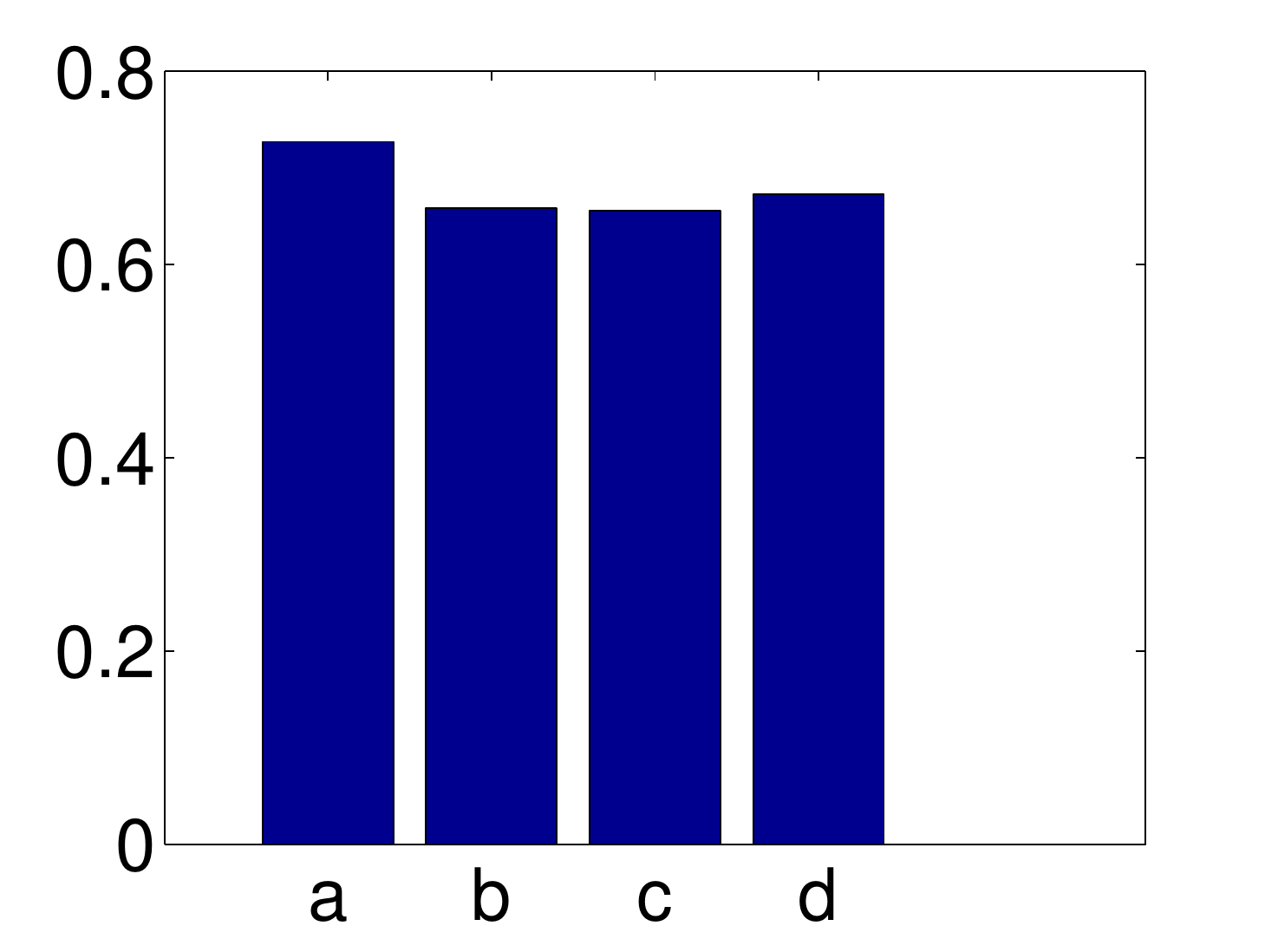}}\\
    \subfigure[$M_{RV}$]{\includegraphics[scale=.3]{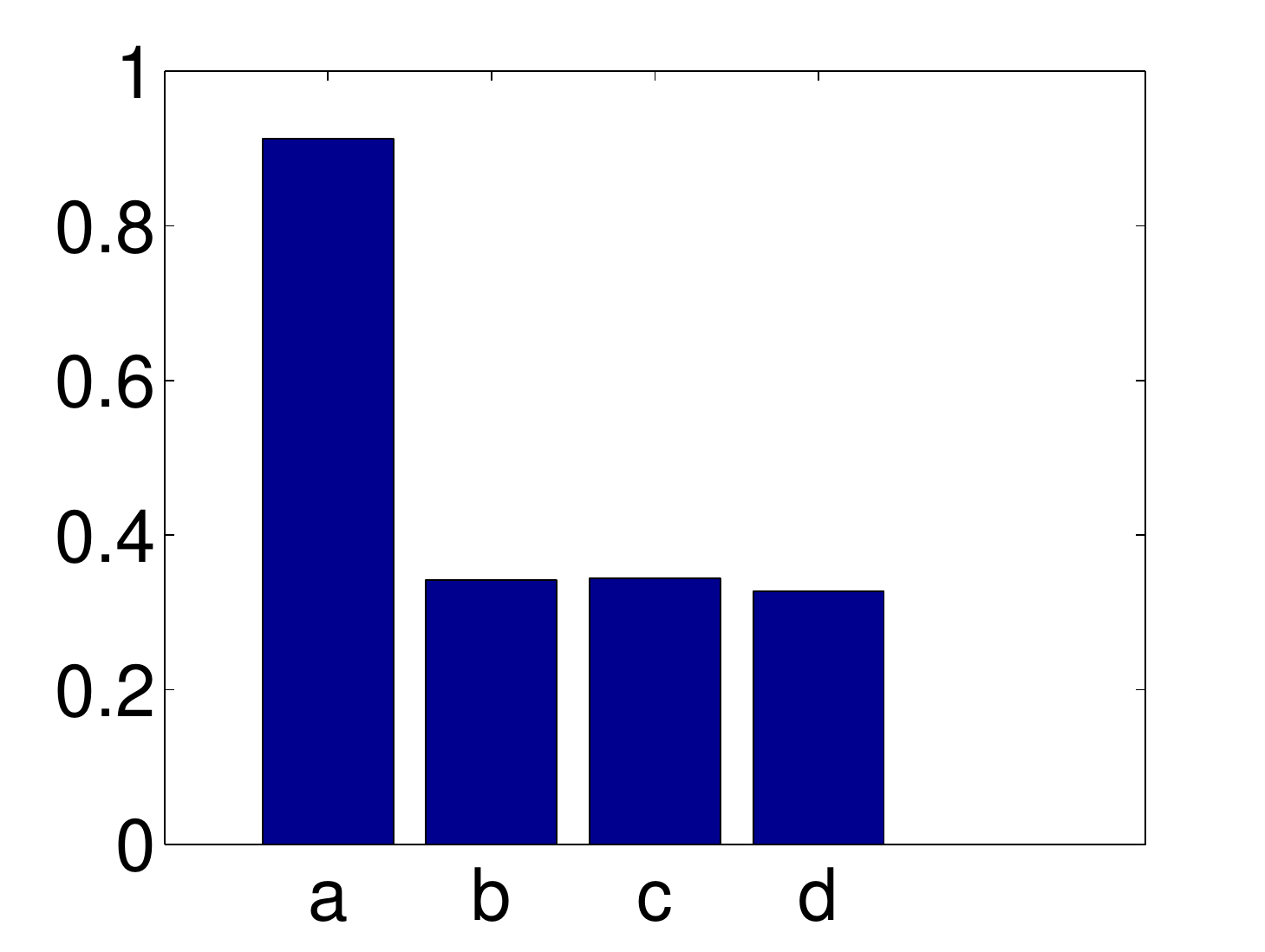}}
    \subfigure[$M_L$]{\includegraphics[scale=.3]{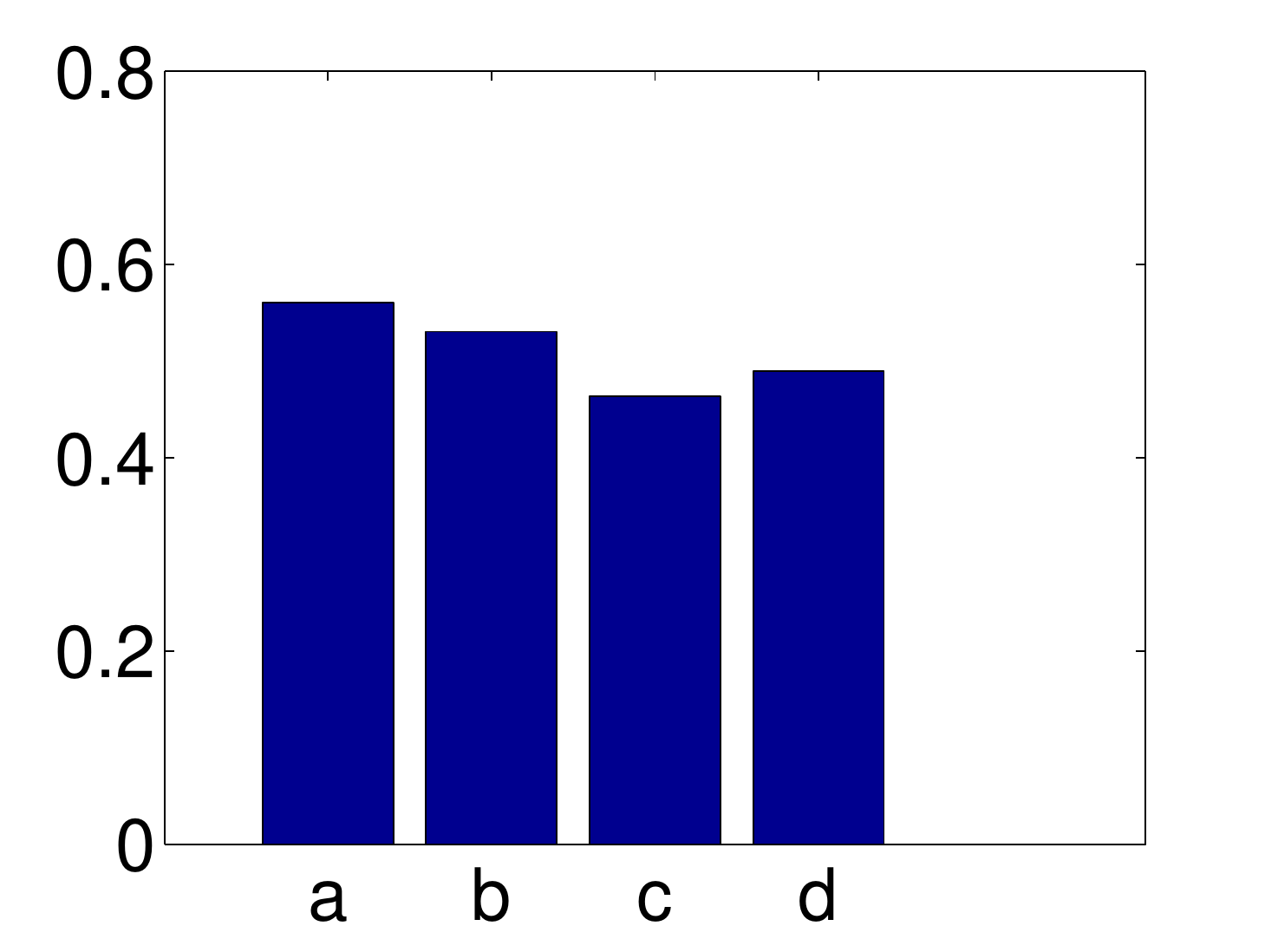}}
    \caption{Manifold learning results on \texttt{lleface}. (a)-(d) Embeddings learned by various methods. The name of each method is stated below each subfigure. (e)-(i) Bar plots of different assessments on learned embeddings. The lower-case character under each bar corresponds to
    the index of the subfigure above.}
    \label{fig:NIEQA-me-lleface}
\end{figure*}


In the next experiment, we apply NIEQA to model evaluation tasks on the
\texttt{lleface} data set, which is a high-dimensional image manifold. As the
code of RML on high-dimensional data is not available, we do not test RML on
this data set. The training data contain 1965 face images, and the intrinsic
degrees of freedom are the angle of face orientation and the variation of
facial emotion. We randomly select 1493 images as training data such that the
data graph constructed via ISOMAP is connected. We apple LLE, LE, ISOMAP and
LTSA to learn this manifold with 15 nearest neighbors. The two dimensional
embeddings learned by these methods and bar plots of the quality assessments
given by different methods are shown in Fig. \ref{fig:NIEQA-me-lleface}.

From Fig. \ref{fig:NIEQA-me-lleface} we can see that the embedding given by LLE
does not recover the change of face orientation. The other methods all
successfully extract the two intrinsic degrees of freedom despite the
difference in embedding shape. The above visual inspection is also validated by
the bar plots of quantitative assessments shown  in Figs.
\ref{fig:NIEQA-me-lleface}(e)-(i). $M_{LC}$, $M_{RV}$ and $M_L$ all suggest
that the quality of the embdding given by LLE is poor, while the others are
almost of the same quality. $M_{LC}$ and $M_L$ indicate that the embedding
given by LTSA is of the highest quality. $M_P$ and $M_P^c$ fail in this case.

\begin{remark}
In experiments on high-dimensional image manifold, we did not compute$M_G$. The
reason lies in that the computation of $M_G$ needs to estimate geodesic
distances based on shortest graph paths. However, we have no prior knowledge on
the underlying geometric structure of image manifolds, hence using $M_G$ to
assess the global topology would yield unknown bias. Also note that the values
of intrinsic degrees of freedom for image manifolds are unknown, hence we do
not compute $M_t$ either.
\end{remark}

\subsection{Model selection}
\label{sec:expt-ms}

\begin{figure*}
    \centering
    \includegraphics[scale=1]{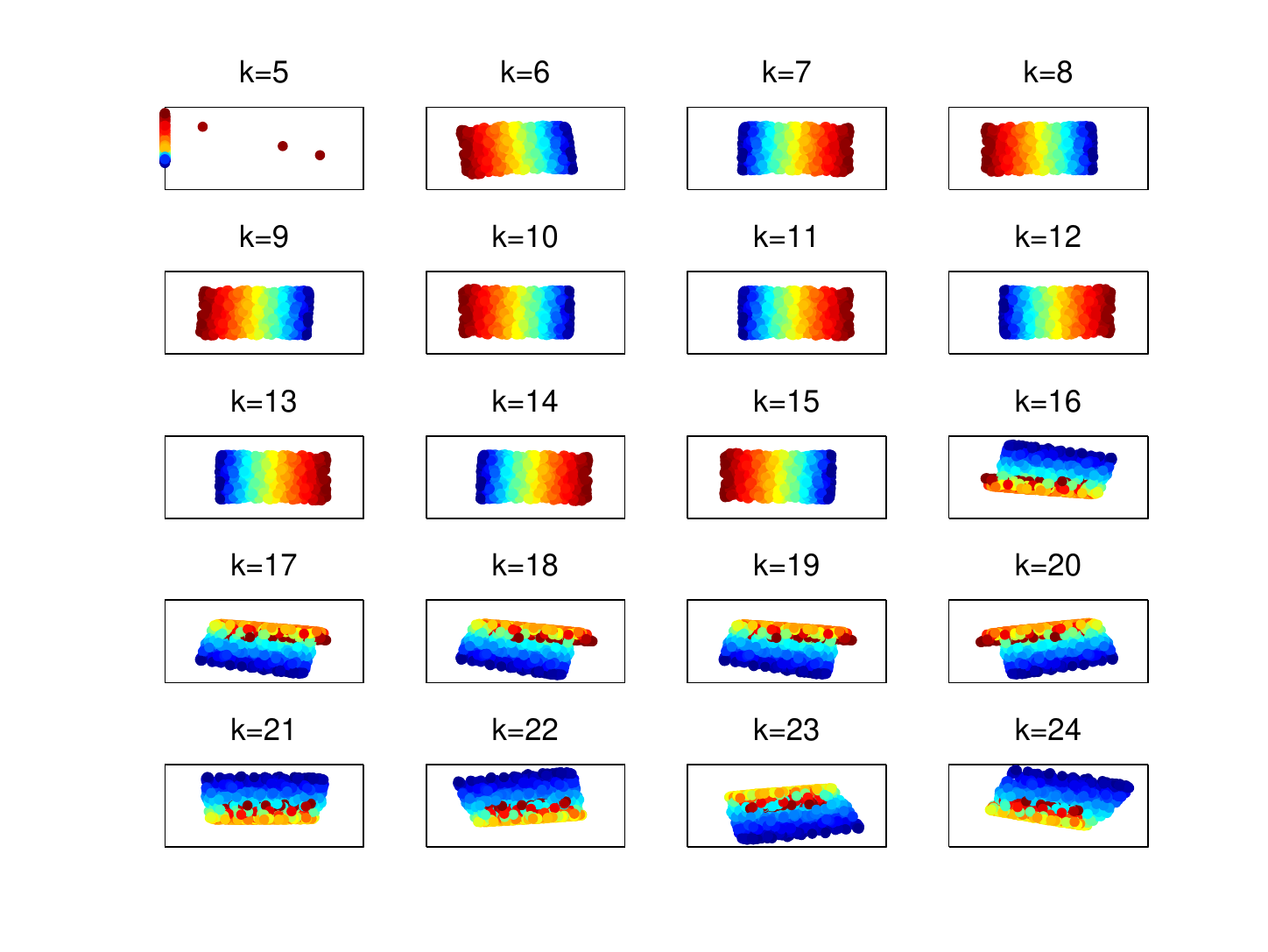}
    \caption{Embeddings given by LTSA on \texttt{Swissroll} data set with different values of $k$.}
    \label{fig:NIEQA-ms-sw}
\end{figure*}

\begin{figure*}
    \centering
    \subfigure[]{\includegraphics[scale=.45]{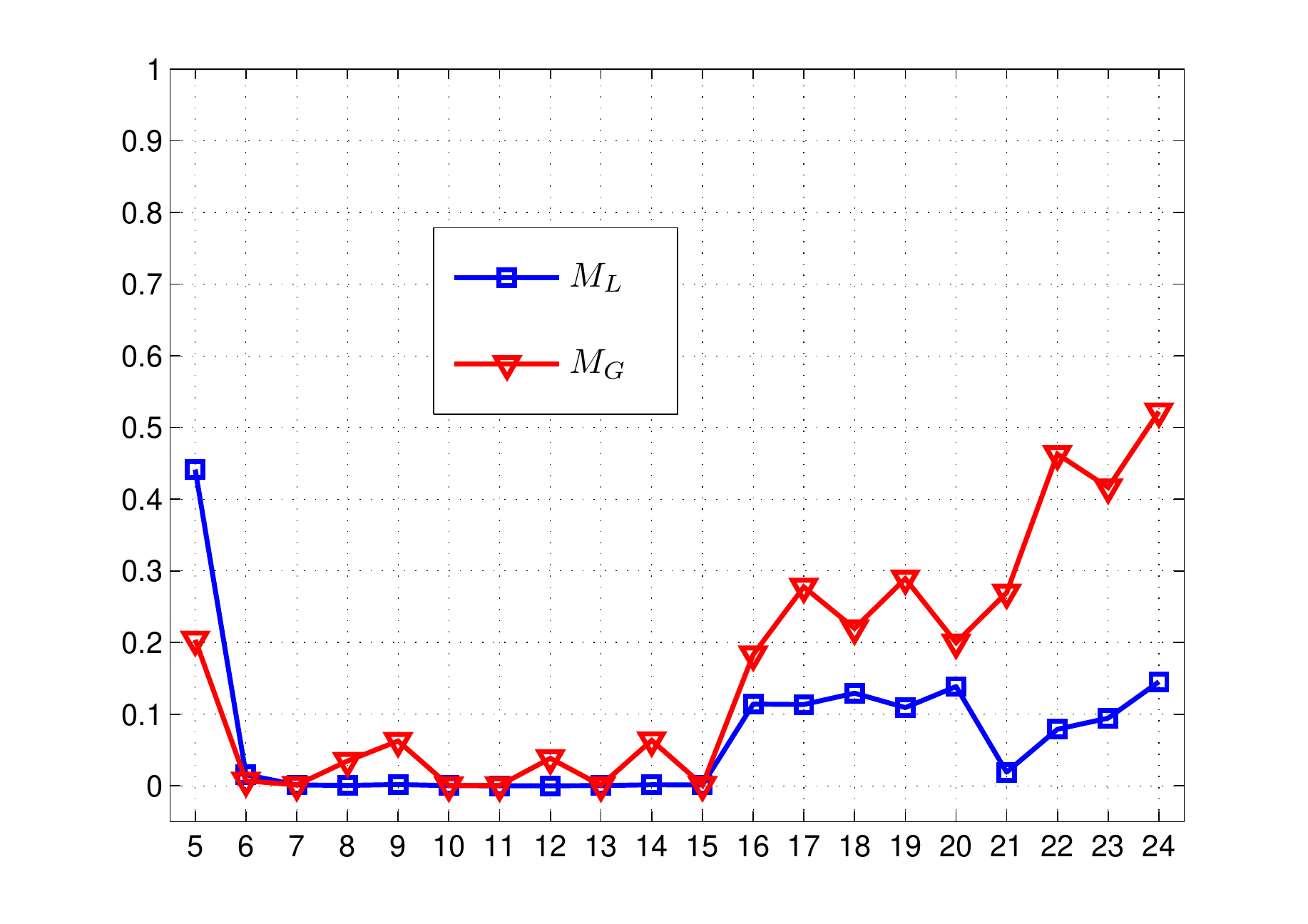}}
    \subfigure[]{\includegraphics[scale=.45]{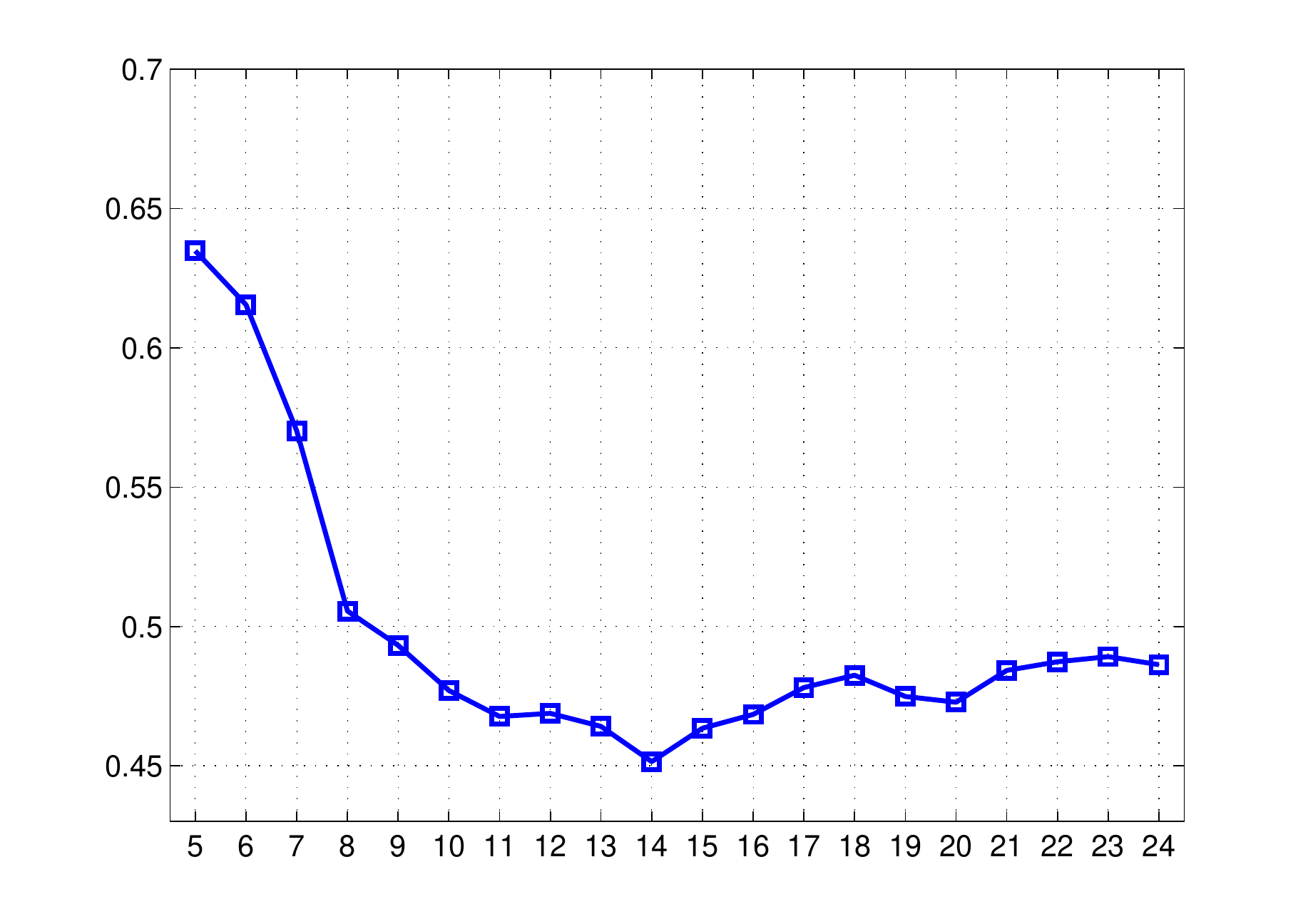}}
    \caption{Graph plot of embedding quality assessments in model selection experiment for the LTSA method. (a) Assessments on \texttt{Swissroll} data set with different valuse of $k$. (b) Assessment $M_L$ on \texttt{lleface} data set with different valuse of $k$. }
    \label{fig:NIEQA-ms-measure}
\end{figure*}

In this subsection, we take the LTSA method as an example to demonstrate the
application of NIEQA to model selection task. The most important parameter for
LTSA is the number of nearest neighbors $k$. We first apply NIEQA to selecting
$k$ for LTSA on the \texttt{Swissroll} data set. Similar to the first
experiment in Section \ref{sec:expt-me}. We randomly select 1000 samples from
the \texttt{Swissroll} manifold as training data. The values of $k$ are chosen
to be integers from 5 to 24. For each $k$, an embedding is learned with LTSA,
which are shown in Fig. \ref{fig:NIEQA-ms-sw}. The assessments given by NIEQA
corresponding to different values of $k$ are shown in Fig.
\ref{fig:NIEQA-ms-measure}(a). From the figure we can see that when $k$ is
taking values between 6 and 15, LTSA would produce embeddings with high
quality. This observation is also supported by visual inspection from Fig.
\ref{fig:NIEQA-ms-sw}, which validates the effectiveness of the NIEQA method.

\begin{figure*}
    \centering
    \subfigure[]{\includegraphics[scale=.4]{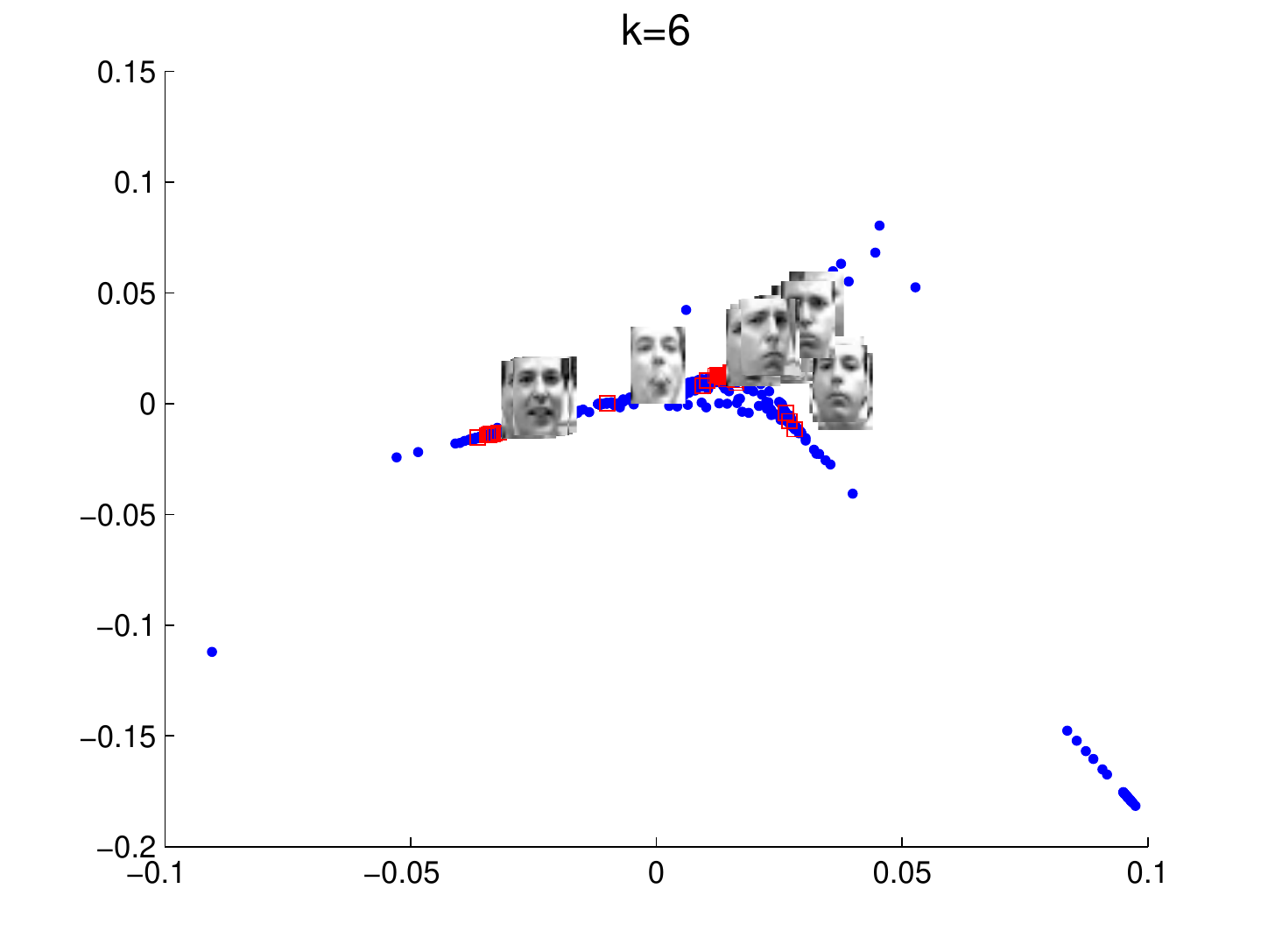}}
    \subfigure[]{\includegraphics[scale=.4]{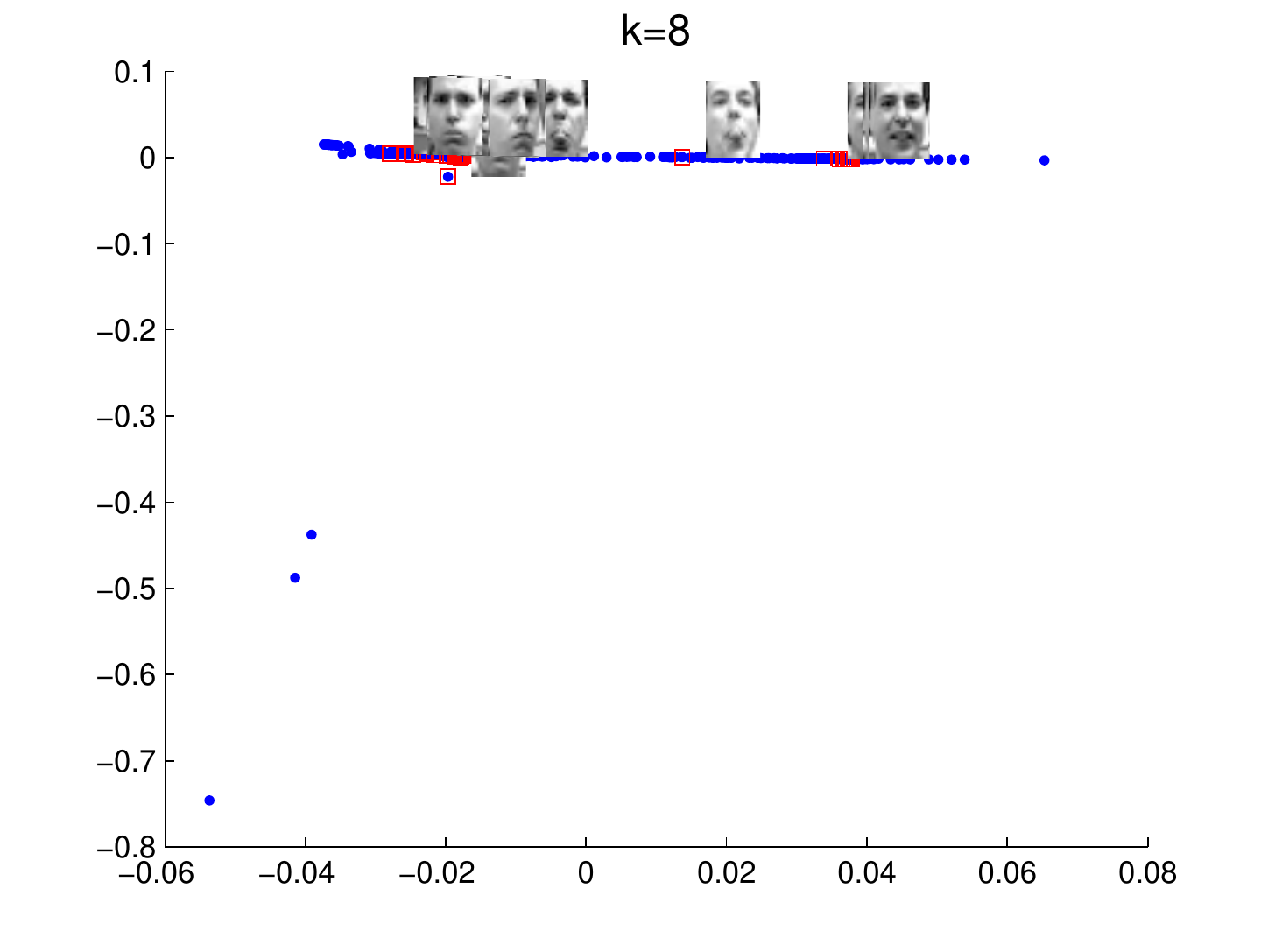}}
    \subfigure[]{\includegraphics[scale=.4]{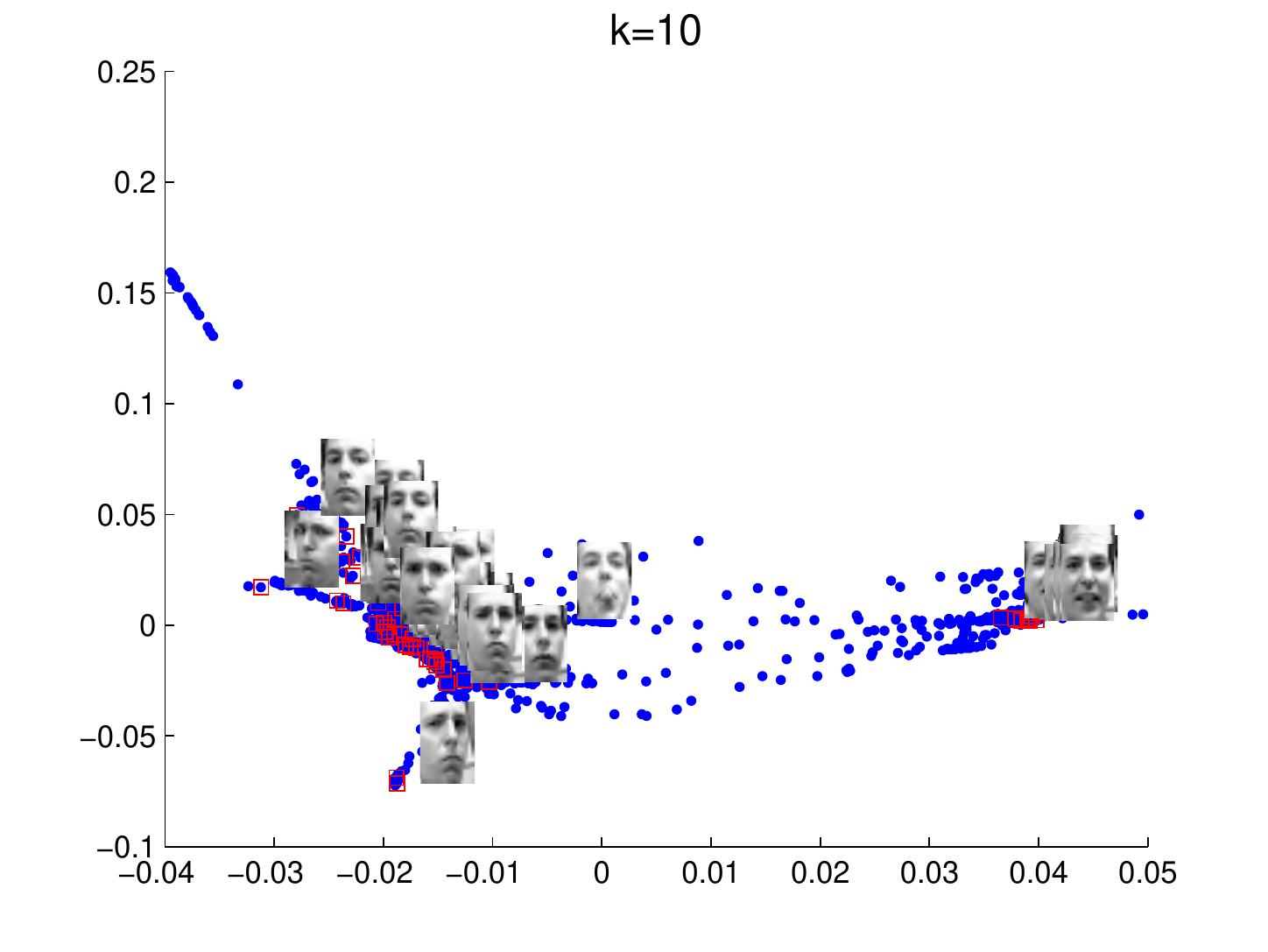}}
    \subfigure[]{\includegraphics[scale=.4]{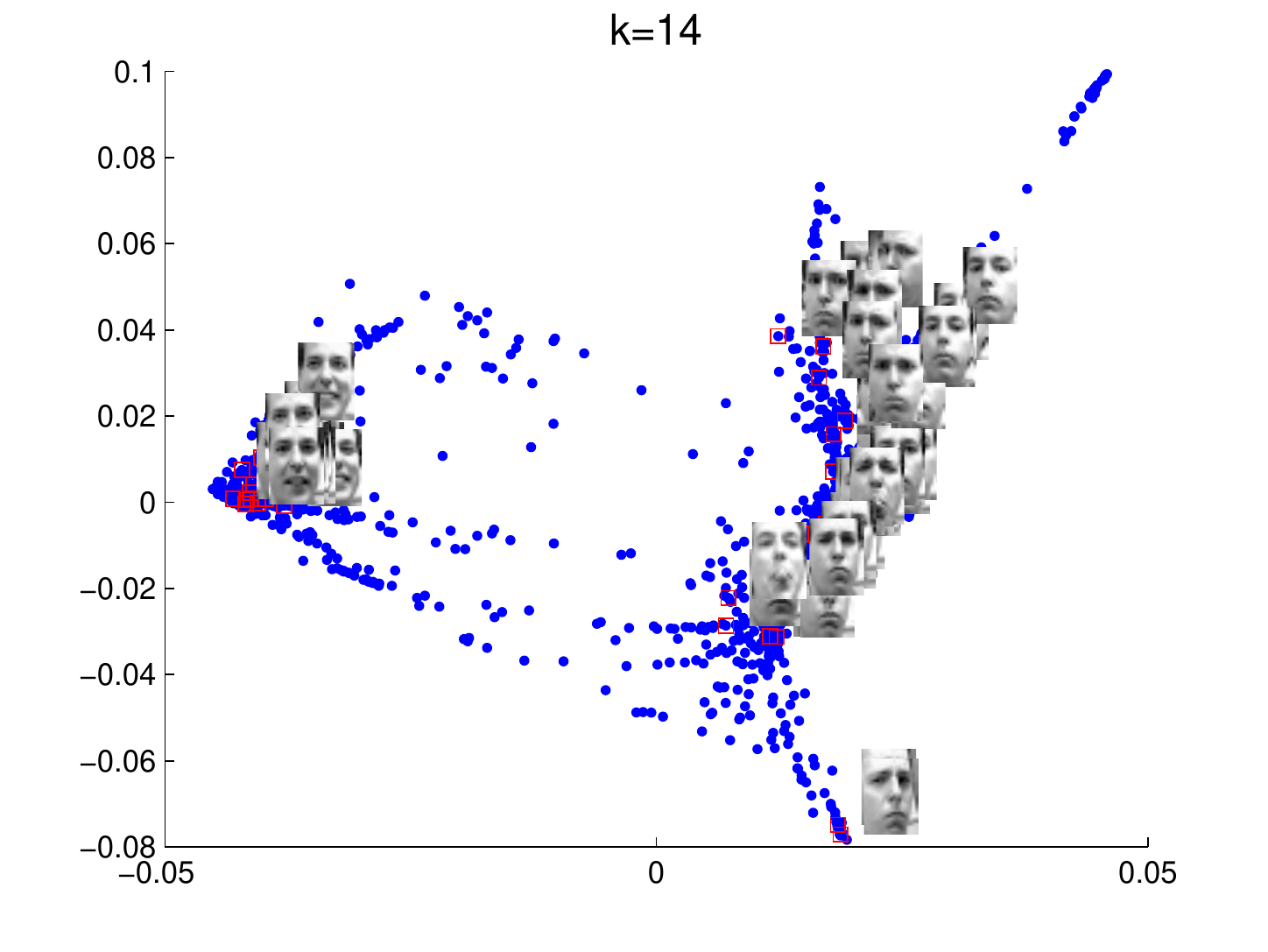}}
    \subfigure[]{\includegraphics[scale=.4]{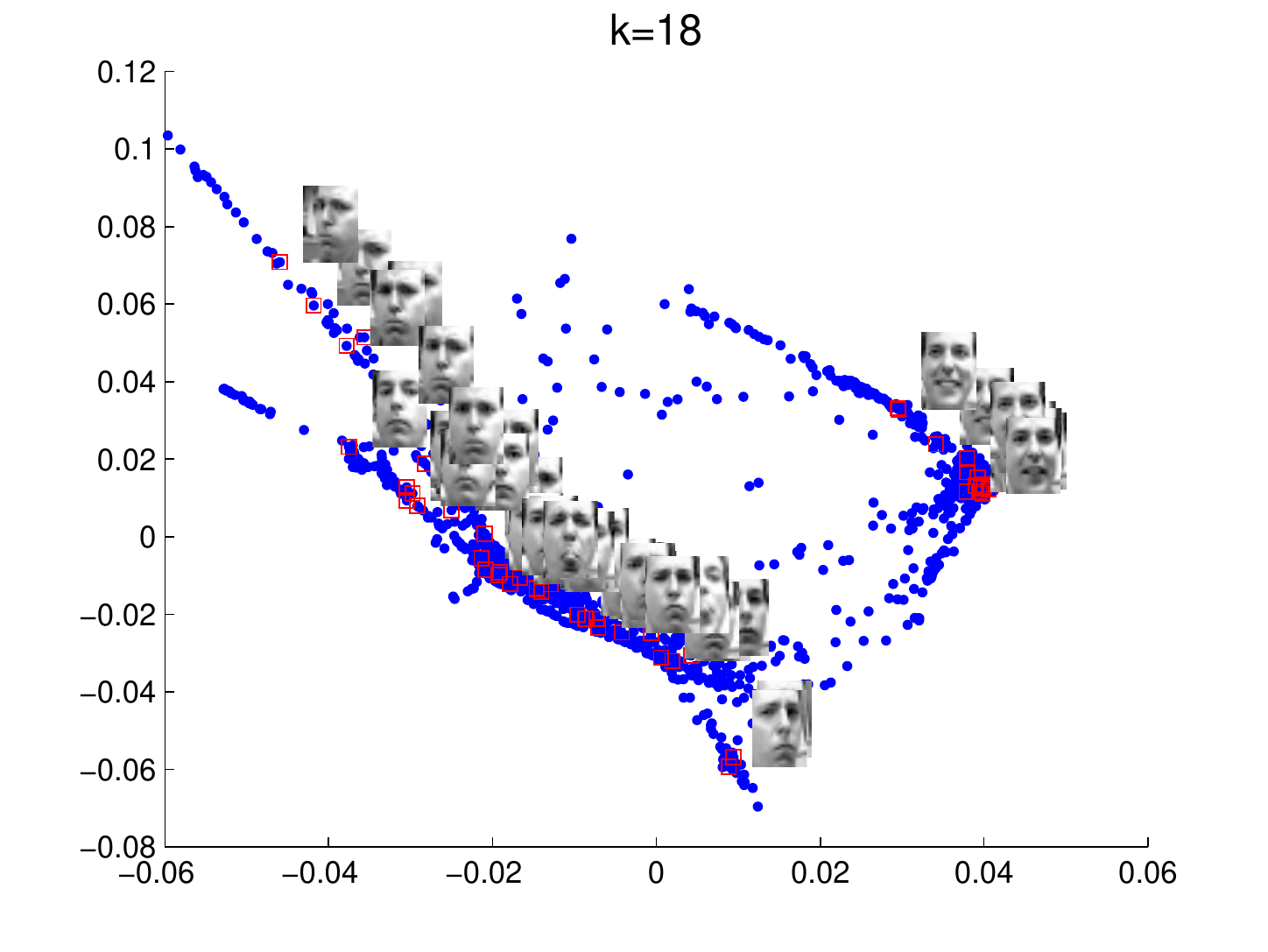}}
    \subfigure[]{\includegraphics[scale=.4]{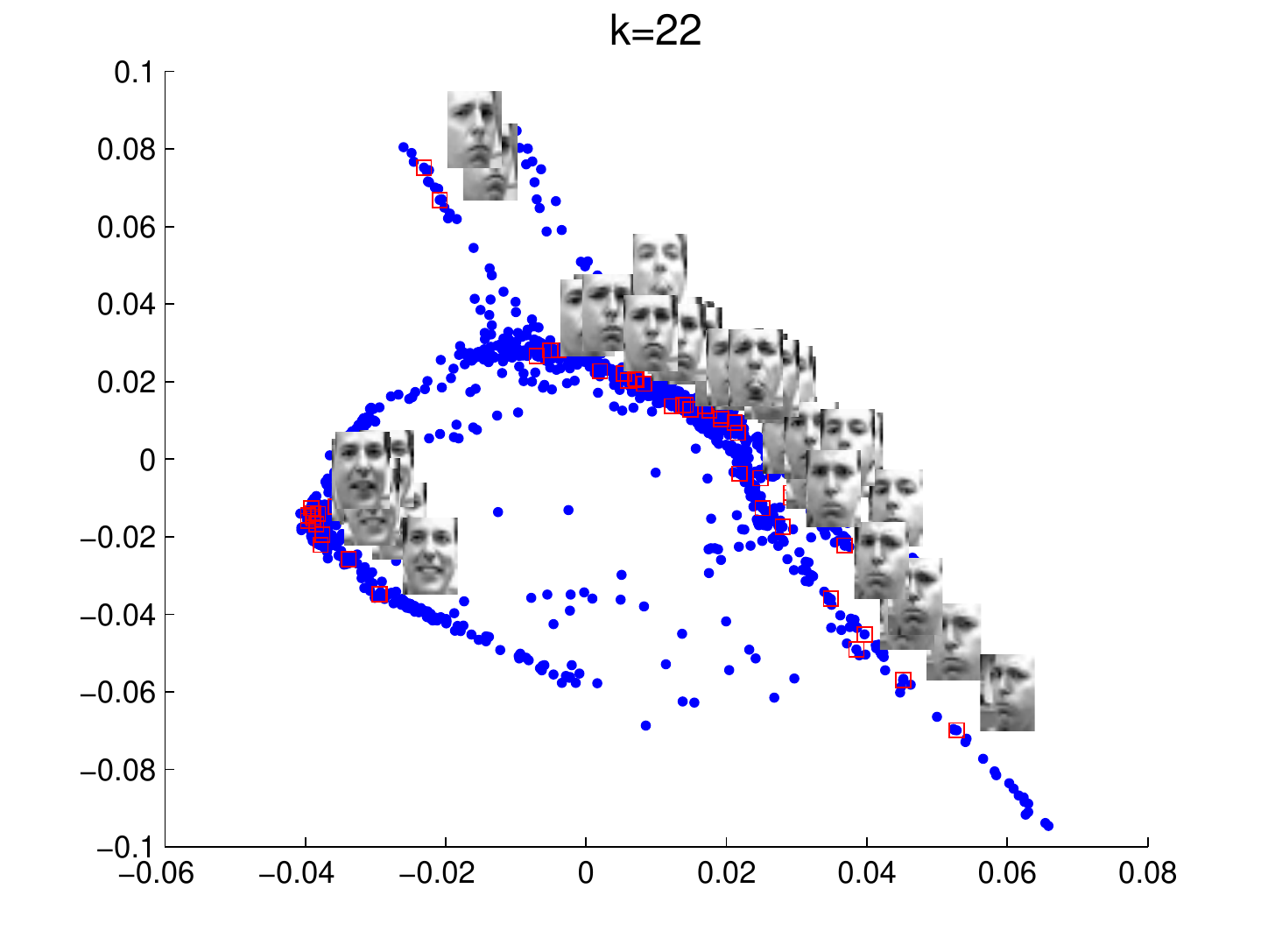}}
    \caption{Embeddings given by LTSA on \texttt{lleface} data set with different values of $k$.}
    \label{fig:NIEQA-ms-lleface}
\end{figure*}

In the second experiment, we apply NIEQA to select optimal $k$ for LTSA on the
\texttt{lleface} data set. Training data are the same to those used in the
experiment in Section \ref{sec:expt-me}. Values of $k$ are taken to be integers
from 5 to 24. For each $k$, an embedding is learned with LTSA, which is shown
in Fig. \ref{fig:NIEQA-ms-lleface}. Corresponding quality assessment given by
NIEQA are illustrated in Fig. \ref{fig:NIEQA-ms-measure}(b), from which we can
see that the embedding corresponding to $k=14$ is of the highest quality. We
can also observe that when $k>8$, the quality of embeddings improves along with
the increase of $k$, which is also validated by visual inspections from Fig.
\ref{fig:NIEQA-ms-lleface}.


\section{Conclusions and discussions}
\label{sec:conclusion}

In this paper, we proposed a novel normalization independent embedding quality
assessment (NIEQA) method for manifold learning, which has wider application
range than current approaches. We first propose a new local measure, which can
quantitatively evaluate how well local neighborhood structure is preserved
under rigid motion and anisotropic coordinate scaling. Then the NIEQA method,
which is designed based on this new measure, can effectively and quantitatively
evaluate the quality of both isometric and normalized embeddings. Furthermore,
the NIEQA method considers both local and global topology, thus it can yield an
overall assessment. Experimental tests on benchmark data sets validate the
effectiveness of the proposed method.

Some discussions and possible improvements in future works are stated below.
\begin{itemize}
    \item The measure $M_{asim}$ is computed by using gradient descent
        method on matrix manifold. Whether the solution converges to a
        global optima remains unproved and is the key part of our future
        works. Meanwhile, we will also consider how to design more
        efficient iteration method to accelerate convergence.
    \item The NIEQA method is based on a local matching methodology. Its
        basic assumption is that the manifold is densely sampled and
        training data strictly lie on the manifold. For data manifold with
        noise or outliers, the efficiency of NIEQA may be affected. A
        possible solution to this issue is to implement denoising or
        outlier removal process before training.
    \item Based on NIEQA, whether we can design a manifold learning method
        with better learning performance is also one of our future works.
\end{itemize}

\section*{Acknowledgements}


This work was partly supported by the NNSF of China grant no. 90820007, and the
973 Program of China grant no. 2007CB311002.




\end{document}